\documentclass[preprint,5p,times,twocolumn]{elsarticle}

\usepackage{amssymb}
\usepackage{amsmath}
\usepackage[dvipsnames,table]{xcolor}
\usepackage{cuted,lipsum}
\usepackage[most]{tcolorbox}
\usepackage{placeins}
\usepackage{algorithmic}
\usepackage{array}
\usepackage{textcomp}
\usepackage{stfloats}
\usepackage{url}
\usepackage{verbatim}
\usepackage{graphicx}
\usepackage{tabularray}
\usepackage{booktabs}
\usepackage{multirow}
\usepackage{rotating}
\usepackage{hyperref}
\usepackage{makecell}
\usepackage[caption=false]{subfig}
\usepackage[ruled,vlined,linesnumbered]{algorithm2e}
\usepackage{balance}
\usepackage{tabularx}
\usepackage{acronym}
\usepackage{color,soul}
\usepackage{underscore}

\newacro{IVQI}{Industrial Visual Quality Inspection}
\newacro{CAM}{Class Activation Mapping}
\newacro{CNN}{Convolutional Neural Network}
\newacro{AI}{Artificial Intelligence}
\newacro{CV}{Computer Vision}
\newacro{XAI}{Explainable Artificial Intelligence}
\newacro{DTD}{Deep Taylor Decomposition}
\newacro{LRP}{Layer-wise Relevance Propagation}
\newacro{ABN}{Attention Branch Network}
\newacro{SGD}{Stochastic Gradient Descent}
\newacro{IoU}{Intersection over Union}
\newacro{GAP}{Global Average Pooling}
\newacro{PCA}{Principal Component Analysis}
\newacro{DNN}{Deep Neural Network}
\newacro{DL}{Deep Learning}
\newacro{IoT}{Internet of Things}
\newacro{ML}{Machine Learning}
\newacro{NN}{Neural Network}
\newacro{VQI}{Visual Quality Inspection}
\newacro{MLOps}{Machine Learning Model Operationalization Management}
\newacro{LLM}{Large Language Model}
\newacro{LVLM}{Large Vision Language Model}
\newacro{COCO}{Common Objects in Context}
\newacro{RISE}{Randomized input sampling for explanation}
\newacro{ASPP}{Atrous Spatial Pyramid Pooling}
\newacro{SHAP}{SHapley Additive exPlanations}
\newacro{XIL}{eXplanatory Interactive Learning}
\newacro{RRR}{Right for the Right Reasons}
\newacro{TPU}{Tensor Processing Unit}
\newacro{FPGA}{Field Programmable Gate Array}
\newacro{FCN}{Fully Convolutional Network}
\newacro{HLS}{High-Level Synthesis}
\newacro{RNN}{Recurrent Neural Networks}

\journal{Information Fusion}

\begin{document}

\begin{frontmatter}



\title{XEdgeAI: A Human-centered Industrial Inspection Framework \\with Data-centric Explainable Edge AI Approach} 

\author[label1]{Hung Truong Thanh Nguyen\corref{cor1}}
\ead{hung.ntt@unb.ca}
\author[label2]{Loc Phuc Truong Nguyen\corref{cor2}}
\ead{loc.pt.nguyen@fau.de}
\author[label1]{Hung Cao}
\ead{hcao3@unb.ca}

\cortext[cor1]{Corresponding author.}
\affiliation[label1]{organization={Analytics Everywhere Lab, University of New Brunswick},            city={Fredericton},
postcode={E3B 5A3}, 
state={New Brunswick},
country={Canada}}
\affiliation[label2]{organization={Friedrich-Alexander-Universität Erlangen-Nürnberg},
            city={Erlangen},
            postcode={91054},
            state={Bavaria},
            country={Germany}}

\begin{abstract}
Recent advancements in deep learning have significantly improved visual quality inspection and predictive maintenance within industrial settings. 
However, deploying these technologies on low-resource edge devices poses substantial challenges due to their high computational demands and the inherent complexity of Explainable AI (XAI) methods.
This paper addresses these challenges by introducing a novel XAI-integrated Visual Quality Inspection framework that optimizes the deployment of semantic segmentation models on low-resource edge devices.
Our framework incorporates XAI and the Large Vision Language Model to deliver human-centered interpretability through visual and textual explanations to end-users. This is crucial for end-user trust and model interpretability.
We outline a comprehensive methodology consisting of six fundamental modules: base model fine-tuning, XAI-based explanation generation, evaluation of XAI approaches, XAI-guided data augmentation, development of an edge-compatible model, and the generation of understandable visual and textual explanations. 
Through XAI-guided data augmentation, the enhanced model incorporating domain expert knowledge with visual and textual explanations is successfully deployed on mobile devices to support end-users in real-world scenarios. 
Experimental results showcase the effectiveness of the proposed framework, with the mobile model achieving competitive accuracy while significantly reducing model size. 
This approach paves the way for the broader adoption of reliable and interpretable AI tools in critical industrial applications, where decisions must be both rapid and justifiable.
Our code for this work can be found at \url{https://github.com/Analytics-Everywhere-Lab/vqixai}.
\end{abstract}

\begin{keyword}
Explainable Edge AI \sep Industrial Visual Quality Inspection \sep Large Vision Language Model
\end{keyword}

\end{frontmatter}




\section{Introduction}
%
Industrial Visual Quality Inspection systems are automated mechanisms, typically engineered to examine and continuously monitor the status of industrial hardware assets. 
In the rapidly evolving landscape of \ac{AI}, especially \ac{CV}, the integration of vision techniques has revolutionized how manufacturers automatically maintain and ensure product quality and compliance, thereby mitigating human error and augmenting efficiency.

\ac{DL}-based models, such as \acp{DNN}, have markedly improved the precision of numerous visual quality inspection systems has been significantly enhanced in terms of accuracy, efficiency, and running time. 
However, these advancements have significant trade-offs between accuracy, computational complexity, and interpretability. 
While achieving higher accuracy, the increased computational complexity of these models often results in a lack of interpretability. 
These opaque models, often perceived as ``black boxes,'' pose challenges for domain experts and end-users in comprehending their internal decision-making processes. 
This opacity becomes a significant concern in sensitive domains, where decisions have profound implications \cite{garouani2022towards,nguyen2023towards,wu2021locally,xu2020explainable}.

The advent of \ac{XAI} has steered a new era of model transparency and interpretability, fundamentally transforming how we understand and interact with AI systems.
\ac{XAI} methods enable domain experts to validate the model's reasoning process and identify potential biases or errors in the data \cite{lr7,lr9,lr11} or model \cite{weber2022beyond,clement2023coping,lr22,lr23,lr24}.
Despite the growing trend on \ac{XAI}, there remains a significant gap in the practical application of these \ac{XAI} techniques to enhance the performance and interpretability of visual quality inspection models, particularly in resource-constrained environments like edge devices.
This gap highlights a critical need for innovative solutions that adapt XAI methodologies for use in environments where computational resources are at a premium.

Our research introduces a cutting-edge framework for industrial visual quality inspection systems that integrates XAI and \ac{LVLM} to 1) \textit{improve the accuracy of semantic segmentation models for industrial assets} and 2) \textit{provide visual and textual human-centered explanations to end-users}.
To this end, we address the following research questions:

\begin{enumerate}
\item How can we effectively integrate XAI methods into Industrial Visual Quality Inspection systems to provide meaningful explanations for model predictions?
\item How can domain expert insights, informed by XAI explanations, be utilized for data augmentation strategies to improve the performance of semantic segmentation models?
\item What are the optimal techniques for adapting and optimizing the enhanced model for deployment on edge devices, ensuring that it remains both effective and efficient in resource-constrained environments?
\item How can we design the delivery mechanisms for human-centered explanations on edge devices, making them accessible and comprehensible to end-users without prior expertise in AI or XAI?
\end{enumerate}

To address these research questions, we propose a framework that is organized into six modules: base model finetuning, XAI-based explanation generation, XAI evaluation, XAI-guided data augmentation, edge model development, and the generation of visual and textual explanations for the edge model. 
Our contributions are as follows:

\begin{enumerate}
\item We develop a comprehensive XAI-integrated Visual Quality Inspection framework that incorporates multiple XAI techniques directly into the model development process, enhancing both interpretability and performance.
\item We validate the effectiveness of XAI-guided data augmentation, demonstrating substantial improvements in the performance of semantic segmentation models for industrial asset inspection.
\item We introduce a tailored mobile optimization algorithm that employs pruning and quantization techniques, enabling efficient deployment of our advanced models on resource-constrained edge devices.
\item We extend the functionality of our framework by integrating \ac{LVLM} to provide clear, contextual textual explanations to end-users on edge devices, making the system's decisions transparent and understandable.
\end{enumerate}

The remainder of the paper is structured as follows:
Section \ref{s:background} presents the \ac{XAI} landscape and recent development of XAI-based model improvement methods.
Section~\ref{s:related-work} provides an overview of related work in semantic segmentation, \ac{XAI} techniques, the industrial visual quality inspection, the application of \ac{XAI} in industrial settings and the \ac{LVLM} utilization in the human-centered \ac{XAI} context.
Section~\ref{s:methodology} presents our methodology, detailing each module of the XAI-integrated Visual Quality Inspection framework.
Section~\ref{s:implementation} describes the implementation details, including model architecture, the procedure of \ac{XAI} and \ac{LVLM} integration, and the mobile optimization algorithm.
Section~\ref{s:exp} presents our first experiment with experimental results on an industrial assets dataset, evaluating the performance of the models in different stages, the effectiveness of XAI-guided data augmentation, and the demonstration of visual and textual explanations.
In Section~\ref{s:datacentric}, we further extend our framework with the data-centric approach, where we evaluate the framework ability on a new industrial dataset while inheriting findings and models from the previous experiment.
Finally, in Section~\ref{s:discussion} and \ref{sec:conclusion}, we discuss and conclude the key findings and limitations from the experimental results and future mitigation research directions.
\section{Background}\label{s:background}
This section offers a comprehensive overview of the XAI landscape and the key factors driving emerging research trends in XAI. Additionally, we summarize common XAI-based approaches for enhancing AI model performance.

\begin{figure*}[!ht]
    \centering
    \includegraphics[width=0.7\linewidth]{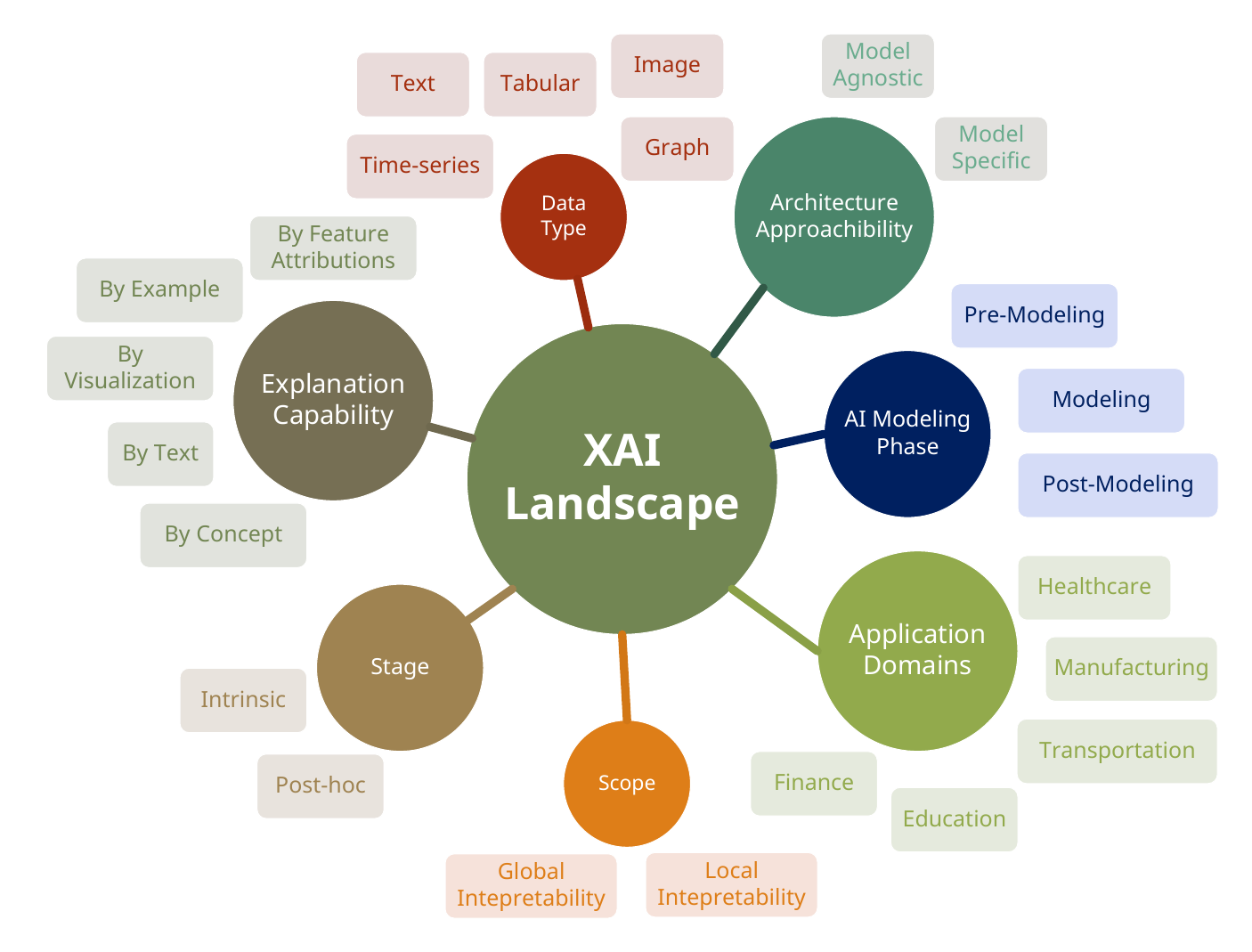}
    \caption{The XAI landscape, categorizing XAI techniques based on explanation scope, data type, modeling phase, architecture, and application domains.}
    \label{fig:xai-classification}
\end{figure*}

\subsection{\acf{XAI} Landscape} 
In this section, we delve into the comprehensive landscape of XAI, focusing on the critical elements that shape its development and application. Specifically, we examine the Modeling Phase, Scope of Interpretability, Explanation Stage, Explanation Capability, and Architecture Approachability, as well as the significance of data-centric approaches in XAI. Additionally, we discuss various Application Domains to illustrate the practical impact of XAI across different fields. This detailed overview aims to provide a clear understanding of how XAI techniques are categorized and applied, shedding light on the diverse range of interpretability methods.

\subsubsection{Modeling Phase}
The explainability of an AI model can be divided into three main phases: Pre-Modeling, Modeling, and Post-Modeling~\cite{10.1145/3580305.3599578}.
\begin{itemize}
    \item \textbf{Pre-Modeling:} This phase focuses on analyzing, characterizing, and exploring the input data to understand the underlying properties of its features or to interpret the overall dataset summary. Typical techniques for data explanation include dataset descriptions and standardization methods (e.g., metadata descriptions, provenance, variable relationships, statistics, ground truth correlation, data cleaning, normalization, standardization), exploratory data analysis, explainable feature engineering, dataset summarization and enhancement (e.g., identifying meaningful abnormalities, summarizing interpretable outliers, selecting prototypes, and enhancing data interpretability). These techniques are generally applied before model creation.   
    \item \textbf{Modeling:} This phase is dedicated to designing, developing, and implementing explainable model architectures and algorithms. The aim is to create new, understandable architectures and algorithms or to transform existing ``black box'' models into transparent ones. Approaches in the Modeling Phase include mixed explanation and prediction (e.g., multimodal explanations, rationalizing neural predictions), architectural modifications for explainability, hybrid explainable models, and regularization for explainability. These methods are typically employed during the model creation process.
    \item \textbf{Post-Modeling:} This phase aims to extract explanations and enhance the interpretability of output results. Techniques used for explaining outputs include macro-explanations, output visualizations, and interpreting targets at different levels of complexity (e.g., comparing functional (external, interpretative) explanations with mechanistic (internal, algorithmic) explanations). These methods are typically applied after the model has been created.    
\end{itemize}

\subsubsection{Scope of Interpretability}
Model interpretability often involves understanding the rationale behind a model's decision-making process. There are two levels of interpretability: global explanations and local explanations \cite{molnar2022}.
\begin{itemize}
    \item \textbf{Global explanations:} Global explanations provide an overall understanding of the key features in an AI model. This is achieved by analyzing how these features impact model accuracy or their influence on the model’s predictions. Such explanations are valuable for guiding policy decisions and testing hypotheses about feature importance.
    \item \textbf{Local explanations:} A local explanation provides insight into how an AI model arrived at a specific prediction or decision. They can answer questions  such as \textit{``Why did the model produce this output for this input?''} or \textit{``What if this feature had a different value?''} Local explanations are essential for validating and refining the model's decisions on a case-by-case basis, particularly when unexpected results occur.
\end{itemize}

\subsubsection{Explanation Stage}
Model explanations can be achieved in two primary ways: by adopting a transparent model (intrinsic) or by interpreting the model after it has been trained (post-hoc) \cite{molnar2022}. 
\begin{itemize}
    \item \textbf{Intrinsic:} Intrinsically explainable methods are designed for transparency, providing direct access to their decision-making logic. Examples include linear regression models, decision trees, and models that generate explanations during the learning process.
    \item \textbf{Post-hoc:} Post-hoc explanation methods provide insights into ``black-box" models, such as DNNs, which are not directly interpretable.
\end{itemize}

\subsubsection{Explanation Capability}
Model interpretability is achieved through various explanation techniques, each tailored to specific needs and contexts. Typically, these explanations assist users in five forms: text, visualizations, examples, concepts, and feature attributions \cite{arrieta2020explainable}.
\begin{itemize}
    \item \textbf{By Text:} Text-based explanations aim to provide insights into model decisions by generating relevant words, phrases, or sentences that describe the model's reasoning. These explanations can also include symbols that represent the model's functioning, effectively mapping the model's rationale into a comprehensible form.
    \item \textbf{By Visualization:} Visual explanations utilize images or visualizations to interpret a model's decisions and behavior. Techniques such as saliency maps, feature highlighting, and dimensionality-reduced visualizations highlight the features and parts of a data instance that are most relevant to the model's decision.
    \item \textbf{By Example:} Example-based explanations clarify model decisions by providing representative instances or examples similar to the query instance. These methods help users understand model behavior by comparing decisions to known examples. The types of examples can include counterfactuals (instances that significantly change the model's prediction), adversarial examples (instances that can deceive models), prototypes (representative instances of the entire dataset), and influential instances (training instances that most impact the model's parameters and predictions).
    \item \textbf{By Concept:} 
    Concept-based explanations enhance our understanding of a model's decision-making process by conveying the driving factors behind a prediction in terms of human-understandable concepts.
    \item \textbf{By Feature Attributions:} Feature attribution methods calculate relevance scores to identify and emphasize the specific features of the input data that have a significant impact on the model's output. This approach is commonly used in text and image data, where it assigns importance to specific words or regions within the images, thereby clarifying their influence on the model’s predictions.
\end{itemize}

\subsubsection{Architecture Approachability}
Post-hoc XAI methods can be categorized as model-specific or model-agnostic, depending on their applicability \cite{molnar2022}.
\begin{itemize}
    \item \textbf{Model-specific:} Model-specific methods are limited to specific types of models. For example, interpreting regression weights in a linear model is a model-specific approach, as the interpretation is unique to that model type.
    \item \textbf{Model-agnostic:} Model-agnostic methods
    can be applied to any AI model and are used after the model has been trained (post hoc). These methods typically analyze the relationships between input and output pairs. By definition, they do not access the model's internals, such as weights or structural information.
\end{itemize}

\subsubsection{Data-centric in XAI}
Data is fundamental in developing AI systems. For many years, efforts have been concentrated on model-centric AI, primarily using fixed benchmark datasets. This focus has led developers to prioritize refining their models to enhance performance, placing a heavy reliance on data quality. However, this approach often overlooks the fact that data can be flawed and unreliable. As a result, there has been a recent shift towards enhancing the reliability, integrity, and quality of data, rather than focusing solely on model improvements. 

In the data-centric AI approach, models are kept locked and unchanged during the experimentation and fine-tuning processes, while datasets are continuously modified, improved, and augmented. This method ensures that models perform accurately and remain unbiased across various data distributions and conditions. Enhancing data explainability helps the AI community build more robust systems and tackle complex real-world problems.

Zha \textit{et al.} \cite{zha2023data} summarize the tasks involved in developing data to simplify input and improve diversity, fairness, and understandability. The authors claim that this procedure potentially leads to better model performance, generalization, and robustness. In the field of XAI, most methods focus on three main data types: tabular data, images, and text. On the other hand, there are limited XAI methods for tackling real-world challenges with time-series or graph data~\cite{bodria2023benchmarking}. Several examples of data-centric approaches in XAI can be found in Section \ref{subsec:app-domain}.

\subsubsection{Application Domains}
\label{subsec:app-domain}
\ac{XAI} has gradually been integrated into multiple sectors, such as automated transport, healthcare, finance, and education. These real-world implementations have demonstrated XAI's significant potential in enhancing model transparency, trust, reliability, and user acceptance. Table \ref{tab:app-sum} provides a summary of the state-of-the-art XAI applications across these sectors.

\def\thickhline{\noalign{\hrule height1pt}}
\begin{table*}[!ht]
    \centering
    \small
    \setlength{\tabcolsep}{3pt}
    \renewcommand{\arraystretch}{1.2} %
    \renewcommand\tabularxcolumn[1]{m{#1}}
    \begin{tabularx}{\textwidth}{ccX|ccc|cc|cc|ccccc|ccc|ccccc}
    \thickhline
          &    &   & \multicolumn{3}{c|}{\textbf{Phase}} & \multicolumn{2}{c|}{\textbf{Scope}} & \multicolumn{2}{c|}{\textbf{Stage}} & \multicolumn{5}{c|}{\textbf{Capability}} & \multicolumn{3}{c|}{\textbf{Audience}} & \multicolumn{5}{c}{\textbf{Augmentation}}
          \\ 
          \textbf{Domain} & \textbf{Ref.} & \textbf{XAI Contributions} & \rotatebox{90}{Pre-Modeling} & \rotatebox{90}{Modeling} & \rotatebox{90}{Post-Modeling} & \rotatebox{90}{Global} & \rotatebox{90}{Local} & \rotatebox{90}{Intrinsic} &  \rotatebox{90}{Post-hoc} & \rotatebox{90}{Text} & \rotatebox{90}{Visualization} & \rotatebox{90}{Example} & \rotatebox{90}{Concept} & \rotatebox{90}{Feature Attributions} & \rotatebox{90}{AI Experts} & \rotatebox{90}{Domain Experts} & \rotatebox{90}{End-users} & \rotatebox{90}{Data} & \rotatebox{90}{Feature} & \rotatebox{90}{Loss} & \rotatebox{90}{Gradient} & \rotatebox{90}{Model} \\
    \hline
    \hline
        \multirow{6}{*}{\rotatebox{90}{Manufacturing}} & \cite{clement2023coping}  & Use 2D SHAP-clustered explanations in automatic tuning model's hyperparameters  to predict power consumption & & $\bullet$ & & $\bullet$ & & & $\bullet$ & & & &  & $\bullet$ &$\bullet$ &  & & & & & & $\bullet$\\
    \cline{2-23}
        & \cite{garouani2022towards} & Present a self-explained AutoML with an interactive visualization module in the field of predictive maintenance & $\bullet$ & & & $\bullet$ & $\bullet$ & $\bullet$ & & & & & $\bullet$ & $\bullet$ & $\bullet$ & & & & & & & $\bullet$ \\ 
    \cline{2-23}
        & \cite{rlb4} & Employ LRP to interpret and improve a DL model for classifying metal surface defects & & $\bullet$ & & & $\bullet$ & & $\bullet$ & & $\bullet$ & & & $\bullet$ & $\bullet$ & $\bullet$ & & $\bullet$ & & & &\\
    \hline
        \multirow{7}{*}{\rotatebox{90}{Transportation}}  & \cite{lorente2021explaining}  & Enhance the trust in driver emotion and distraction detection models & & & $\bullet$ & & $\bullet$ & & $\bullet$ & & $\bullet$ & & & & $\bullet$ & & & & & & &  \\ 
    \cline{2-23}
         & \cite{xu2020explainable} & Produce explanations for actions in response to the state of action-inducing objects that could create a hazard & & $\bullet$ & & & $\bullet$ & $\bullet$ & & $\bullet$ & $\bullet$ & & & & $\bullet$ & $\bullet$ & $\bullet$ & & & $\bullet$ & & \\
    \cline{2-23}
         & \cite{rlb5} & Use SHAP to interpret XGBoost, thereby analyzing factors that influence users' willingness to engage in ridesharing & & & $\bullet$ & & $\bullet$ & & $\bullet$  & & $\bullet$ & & & $\bullet$ & $\bullet$ & $\bullet$ & & & & &\\
    \hline
        \multirow{7}{*}{\rotatebox{90}{Finance}} & \cite{dikmen2022effects} & Enhance the trust and performance of model in a complex financial decision-making context & & & $\bullet$ & $\bullet$ & $\bullet$ & & $\bullet$ & & & & $\bullet$ & & & & $\bullet$ & & & & &  \\ 
    \cline{2-23}
        & \cite{wu2021locally} & Use LIME-based explainers for fraud detection model on how each instance contributes to the final model output & & & $\bullet$ & & $\bullet$ & & $\bullet$ & & & & & $\bullet$ & $\bullet$ & $\bullet$  & &&&&&\\
    \cline{2-23}
        & \cite{rlb6} & Comparatively apply SHAP, LIME, GradCAM, Saliency Maps to explain credit scoring model predictions & & & $\bullet$ && $\bullet$ && $\bullet$ & $\bullet$ & $\bullet$ &&&&& $\bullet$ &&&&&&\\
    \hline
        \multirow{5}{*}{\rotatebox{90}{Healthcare}} & \cite{nguyen2023towards} & Enhance the trust for the two-stage object detectors in the thyroid nodule diagnosis system & & $\bullet$ & & $\bullet$ & & $\bullet$ & & $\bullet$ & & & $\bullet$ & $\bullet$ & $\bullet$ & & & & & & & \\ 
    \cline{2-23}
        & \cite{vzlahtivc2023agile} & Propose an intrinsic explainable algorithm to enable the data validation by the medical experts & & $\bullet$ & & $\bullet$ & $\bullet$ & $\bullet$ & & & $\bullet$ & & $\bullet$ & $\bullet$ & $\bullet$ & $\bullet$ & & $\bullet$ & & & & \\
     \cline{2-23}
        & \cite{rlb1} & Implement SHAP and LIME for enhancing interpretability in cardiovascular disease prediction & & & $\bullet$ & & $\bullet$ & & $\bullet$ & & $\bullet$ & & & $\bullet$ & & $\bullet$ & & & $\bullet$ & & &\\ 
    \hline
        \multirow{5}{*}{\rotatebox{90}{Education}} & \cite{melo2022use}  & Provide insights into factors influencing school dropout predictions & & & $\bullet$ & $\bullet$ & $\bullet$ & & $\bullet$ & & & & &$\bullet$ & & $\bullet$ & & & & & &\\ 
    \cline{2-23}
        & \cite{nur2022explainable} &  Identify features important to the development of an interactive knowledge discovery tool & & & $\bullet$ & $\bullet$ & $\bullet$ & & & & & $\bullet$ & & $\bullet$ & & $\bullet$ & & & & & &\\ 
    \cline{2-23}
        & \cite{tsiakmaki2021case} & Study counterfactual explanations focusing on student performance prediction & & & $\bullet$ & & $\bullet$ & & & & & $\bullet$ & & & & $\bullet$ & & & & & & 
        \\ 
    \thickhline
    \end{tabularx}
    \caption{An overview of XAI applications across various domains categorized by their modeling phase, the scope of interpretability, explanation stage, explanation capability, target audiences, and model improvement augmentation types.}
    \label{tab:app-sum}
\end{table*}

\begin{itemize}
    \item \textbf{Manufacturing:}
    \ac{XAI} techniques have been extensively applied in manufacturing to support the adoption of \ac{DL} models in safety-critical aspects. These include predictive maintenance~\cite{garouani2022towards}, energy consumption optimization~\cite{clement2023coping}, visual inspection~\cite{kardovskyi2021artificial,rlb4}, and qualification~\cite{diaz2021guided}. By integrating \ac{XAI}, the reliability and efficiency of manufacturing processes are significantly enhanced.

    \item \textbf{Transportation:}
    In the transportation sector, \ac{XAI} is essential for ensuring safety, enhancing system transparency, and fostering user trust in autonomous vehicles. By providing clear explanations of autonomous decisions, \ac{XAI} helps diagnose and rectify potential issues, contributing to safer and more reliable automated transport systems~\cite{xu2020explainable}. Studies have introduced various XAI models to improve human-vehicle interaction, explain decisions made by advanced driver-assistance systems, and enhance the reliability of autonomous driving systems by mimicking human decision-making processes~\cite{lorente2021explaining,rlb5}.

    \item \textbf{Finance:}
    Financial institutions face increasing pressure to ensure their AI models comply with evolving regulations aimed at safeguarding consumer rights and maintaining market stability. AI models equipped with \ac{XAI} capabilities can provide personalized advice, product recommendations, and risk assessments, thereby enhancing customer experience and building trust. Achieving a balance between interpretability and model performance is crucial for transparency in decision-making processes related to investments, credit scoring, and risk management~\cite{dikmen2022effects,rlb1}. Moreover, \ac{XAI} improves fraud detection mechanisms by offering clear explanations of fraud detection decisions, enabling financial institutions to respond swiftly and accurately~\cite{wu2021locally}.

    \item \textbf{Healthcare:}
    \ac{XAI} significantly benefits healthcare, pharmacy, and bioinformatics by enhancing diagnostic accuracy, building trust, and ensuring the ethical use of AI technologies. Specifically, \ac{XAI} applications not only improve diagnostic and therapeutic outcomes but also unravel complex biological data, revolutionizing patient care and medical research. Recently, \ac{XAI} has further increased diagnostic accuracy by making AI-driven diagnoses more understandable and trustworthy for both clinicians and patients. An increasing number of advanced \ac{XAI} methods are being developed and implemented to interpret AI decision-making processes, yielding promising results~\cite{nguyen2023towards,vzlahtivc2023agile,rlb1}.
    
    \item \textbf{Education:}
    Integrating \ac{XAI} into learning systems can significantly enhance the interpretability of student data analysis, offering educators and students valuable insights into personalized learning patterns, content suggestions, performance metrics, and customized feedback mechanisms~\cite{melo2022use,nur2022explainable,tsiakmaki2021case}. By providing clear explanations of these personalized details, XAI empowers learners to take control of their learning process, fostering a more engaging and effective educational experience.
\end{itemize}

\subsection{XAI-based Model Improvement Methods}
In the rapidly expanding industrial sector, developers require comprehensive information on how AI models process data to maintain accuracy and reliability \cite{lr1,lr2,lr3}. This is not just about understanding model decisions; it involves leveraging that knowledge to improve different aspects of model performance. To meet this demand, \ac{XAI} applications have extended beyond theoretical explanations and visual aids to provide practical insights that actively refine models. However, a significant gap remains despite the wealth of \ac{XAI} research. Very few studies offer a systematic approach to closely examine and classify how \ac{XAI} can enhance both model interpretability and functionality \cite{lr4}. This can create significant challenges for developers who may not fully comprehend the working mechanisms of emerging model improvement techniques, potentially resulting in their misuse or misinterpretation \cite{lr5}. To address this issue, Weber \textit{et al.} \cite{lr6} have proposed a framework that categorizes \ac{XAI}-based techniques according to the component of the training loop they optimize, such as data, intermediate features, loss function, gradients, or model architecture. Below are detailed descriptions of the five main augmentation types and some corresponding notable methods.

\subsubsection{Data augmentation} \ac{XAI}-based data augmentation leverages explanations to reshape data structure. It takes the original dataset and its attributions as inputs, generating augmented data by creating new samples or adjusting existing data distribution. The main objective is to minimize bias and errors that result from using the original data structure. This type of augmentation is often implemented early in the model training process, particularly in the initial forward-backward phase, which influences all subsequent components \cite{lr6}. Approaches that utilize data augmentation techniques for improving model performance include \cite{lr7,lr9,lr11}.
\subsubsection{Feature augmentation} \ac{XAI}-based feature augmentation utilizes explanations to adjust the model's feature representations. First, intermediate features from a specific layer and their attributions are taken as input. Then, \ac{XAI} is employed to evaluate each feature's importance, allowing key features to be selectively scaled, masked, or transformed. Thereby, feature augmentation indirectly influences all higher-level feature representations and subsequent training components, including gradient updates and the final model. This holistic approach helps reduce potential biases and enhances the model's generalization, resulting in more accurate and reliable predictions \cite{lr6}. Techniques that improve model performance through this type of augmentation are proposed in \cite{lr13,lr14,rlb1}.
\subsubsection{Loss augmentation} \ac{XAI}-based loss augmentation modifies the loss function using insights derived from local explanations to guide the training process. This type of augmentation can take different forms, such as adding regularization terms that adjust the model based on feature relevance or modifying the loss function to focus on challenging or underrepresented data points. These adjustments ensure that the model not only focuses on reducing the overall error but also addresses specific inaccuracies identified through explanations, leading to a more balanced and equitable performance across various conditions. Due to its working mechanism, loss augmentation indirectly impacts all components of the backward pass as well as the final model \cite{lr6}. Notable \ac{XAI}-based model improvement methods employing this technique include \cite{lr16,lr17,lr18}.

\subsubsection{Gradient augmentation} \ac{XAI}-based gradient augmentation involves two specific types: feature gradient augmentation and parameter gradient augmentation. Feature gradient augmentation adjusts gradients of intermediate features using scaling, masking, or transformation techniques based on their importance determined by \ac{XAI}. On the other hand, parameter gradient augmentation targets gradients of specific model parameters and relies on \ac{XAI} insights to fine-tune them precisely. Both methods occur during the training stage, with feature gradient augmentation impacting a more significant part of the network by modifying gradient flow across all layers below the target. In contrast, parameter gradient augmentation focuses its effects on specific layers, minimizing its broader network effects. By prioritizing critical gradients during backpropagation, \ac{XAI}-based gradient augmentation effectively guides the learning process, leading to improved performance, faster convergence, and higher data efficiency \cite{lr6}. Several methods utilize this augmentation type to enhance model performance, including \cite{lr19,lr20,lr21}.
\subsubsection{Model augmentation} \ac{XAI}-based model augmentation incorporates two main techniques: pruning and quantization. Pruning focuses on the model’s architecture, using explanations to assess the importance of its parameters. It systematically removes less critical parameters to reduce the model's complexity and storage requirements while maintaining the overall performance. Quantization, on the other hand, aims to optimize the precision of the model's parameters. It leverages \ac{XAI} to find and modify parameters that can tolerate lower precision without causing significant performance degradation. It is worth noticing that both methods are implemented post-training and generally have minimal influence on aspects like accuracy or robustness. However, they are essential for increasing the model’s computational efficiency and speeding up its inference, making it more suitable for deployment in resource-constrained environments \cite{lr6}. A wide range of model improvement methods employing this augmentation approach can be found in \cite{garouani2022towards,clement2023coping,lr22,lr23,lr24}.

\section{Related Work}\label{s:related-work}
In this section, we examine the latest advancements in semantic segmentation and the application of \ac{XAI} techniques to enhance model interpretability and transparency, particularly in the context of industrial visual quality inspection. We also explore the emerging paradigm of Explainable Edge AI and current developments in human-centered XAI, which uses \ac{LVLM} to generate explanations that are accessible to users without a background in AI or \ac{XAI}.

\subsection{Semantic Segmentation}
Semantic segmentation is a process in \ac{CV} where each pixel in an image is classified into one of several predefined categories, thereby segmenting the image into regions with distinct object identities. Unlike image classification, which assigns a single label to an entire image, semantic segmentation provides a detailed, pixel-level understanding of the scene. This section covers recent advancements in semantic segmentation models and local post-hoc XAI methods tailored for semantic segmentation.

\subsubsection{Semantic Segmentation Models}
\ac{DL} backbones such as VGG~\cite{simonyan2014very}, YOLO~\cite{redmon2016you}, ResNet~\cite{he2016deep}, and MobileNet~\cite{howard2017mobilenets} have revolutionized visual quality inspection by offering robust feature extraction capabilities that are essential for segmentation tasks. These backbones form the foundation of semantic segmentation models, enabling precise identification and classification of objects within an image.

Semantic segmentation is a crucial tool for visual quality inspection systems, as it enables these systems to focus on critical parts of an image while ignoring irrelevant regions. Notable examples of semantic segmentation models include FCN~\cite{long2015fully}, LRASPP~\cite{howard2019searching}, and DeepLabV3~\cite{deeplabv32018}. These models represent substantial advancements in semantic segmentation due to their high performance and applicability on mobile devices. 

DeepLabV3~\cite{deeplabv32018}, with its innovative \ac{ASPP}~\cite{chen2016deeplab} module, significantly enhances semantic segmentation models by capturing objects at multiple scales and improving boundary delineation. The latest iteration, DeepLabV3Plus~\cite{yang2023semantic}, incorporates an encoder-decoder structure to further refine object boundaries and details, demonstrating superior performance in various segmentation benchmarks. These continuous improvements in model architecture, efficiency, and accuracy have greatly advanced visual quality inspection systems. By enhancing the segmentation of small objects and intricate details, these developments enable more precise and reliable inspection processes across multiple industries.

\subsubsection{Local Post-hoc XAI Methods for Semantic Segmentation}
This section introduces local post-hoc XAI methods specifically designed for semantic segmentation tasks. It is worth noticing that methods initially tailored for the classification task can also be adapted to work with the outputs of semantic segmentation models \cite{vinogradova2020towards}.
These methods can be categorized based on their explanation generation mechanisms, including Backpropagation-based, \ac{CAM}-based, Perturbation-based \cite{rebuffi2020there}, and Example-based methods:
\begin{itemize}
    \item \textbf{Backpropagation-based}: Backpropagation-based XAI methods explain neural network predictions by using the backpropagation algorithm to compute the gradients of the output with respect to the input features. These gradients reveal how changes in each input feature impact the output, thereby highlighting the most influential features~\cite{bach2015pixel,shrikumar2017learning}.
    
    \item \textbf{\acf{CAM}-based}: 
    CAM-based XAI methods explain neural network predictions by producing heatmaps highlighting important regions in the input data. They work by extracting feature maps from the final convolutional layer, weighting them according to their relevance to the predicted class, and summing the weighted feature maps to generate the heatmaps. This process identifies which parts of the input are most influential in the model's decision~\cite{zhou2016learning,chattopadhay2018grad,selvaraju2017grad,wang2020score,ramaswamy2020ablation,muhammad2020eigen,fu2020axiom,nguyen2022secam,hasany2023seg,nguyen2024efficient}. 
    
    \item \textbf{Perturbation-based}: Perturbation-based XAI methods explain neural network predictions by systematically modifying parts of the input data and observing changes in the model's output. This involves altering specific input features, such as occluding image regions, adding noise, or changing values in structured data. By comparing the original and perturbed outputs, these methods identify which changes significantly impact predictions, indicating the importance of the corresponding input parts. The results are used to create heatmaps highlighting the most influential regions or features \cite{zeiler2014visualizing,lr10,petsiuk2018rise, petsiuk2021black,yang2021mfpp,truong2023towards}.
    
    \item \textbf{Example-based}: Example-based XAI methods explain neural network predictions by identifying and presenting similar instances from the training data. They extract features from the input, measure similarity with training examples, and find the nearest neighbors. The most similar examples and their predictions are then shown to the user, helping them understand how the model makes decisions based on these similarities~\cite{sacha2023protoseg,heide2021x}.
\end{itemize}

\begin{figure}[!ht]
    \centering
    \includegraphics[width=\linewidth]{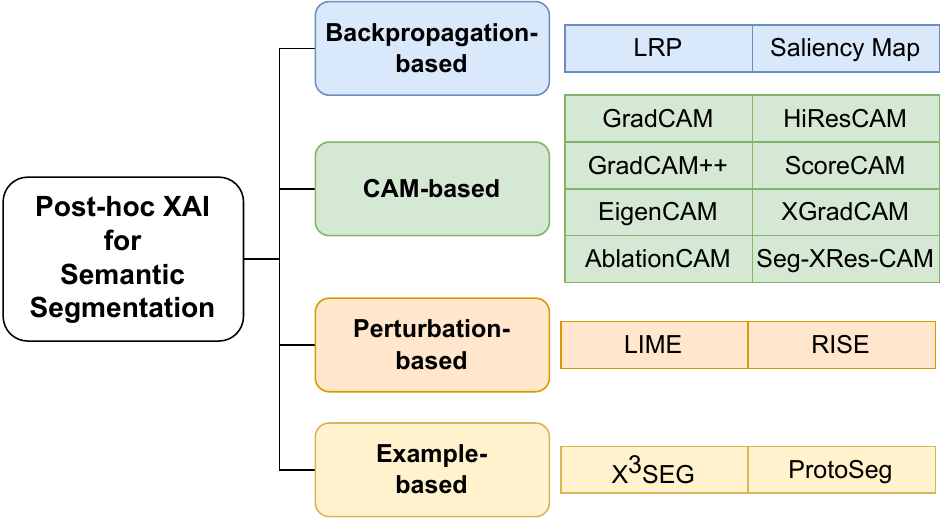}
    \caption{Local Post-hoc XAI Methods for the Semantic Segmentation.}
    \label{fig:xai-classification}
\end{figure}

In this paper, we employ XAI methods that offer visual explanations, with a focus on CAM-based techniques such as GradCAM~\cite{selvaraju2017grad}, GradCAM++~\cite{chattopadhay2018grad}, XGradCAM~\cite{fu2020axiom}, HiResCAM~\cite{draelos2020hirescam}, ScoreCAM~\cite{wang2020score}, AblationCAM~\cite{ramaswamy2020ablation}, GradCAMElementWise~\cite{jacobgilpytorchcam}, EigenCAM~\cite{muhammad2020eigen}, and EigenGradCAM~\cite{muhammad2020eigen}.
These techniques are chosen for their applicability and effectiveness in semantic segmentation tasks. Additionally, we incorporate a perturbation-based XAI method, namely RISE~\cite{petsiuk2018rise}, due to its model-agnostic working mechanism.

\subsection{XAI in Industrial Visual Quality Inspection}
Visual quality inspection is crucial in manufacturing and related industries for ensuring product quality and reducing costs. Traditional methods, however, are often time-consuming and expensive~\cite{tang2017manufacturing}. Recent advancements in deep learning have revolutionized visual quality inspection across sectors such as automotive, electronics, and construction~\cite{sun2018research, md2022review, yasuda2022aircraft}. These AI-driven systems have demonstrated exceptional performance in automating complex inspection tasks, including steel bar quality assessment in construction~\cite{kardovskyi2021artificial}, pear quality inspection in the food industry~\cite{Ilchuk2023COMPUTER}, and guided visual inspection for asset maintenance~\cite{diaz2021guided}.

As AI becomes increasingly prevalent in industrial visual quality inspection, the importance of XAI has become more prominent. XAI aims to make AI systems more transparent and interpretable, fostering better collaboration between human experts and AI~\cite{arrieta2020explainable}. In the industrial quality control and inspection sector, several studies have proposed leveraging XAI to enhance model quality and transparency in visual inspection tasks. For example, Rovzanec \textit{et al.}~\cite{rovzanec2024adaptive} introduced a framework where explanations provide feedback to inspectors, helping improve the underlying classification model for visual defect inspection. Lupi \textit{et al.}~\cite{lupi2023framework} developed a framework for reconfigurable vision inspection systems that use XAI to adapt to varying product types and manufacturing conditions, making XAI more accessible to non-specialist users. Gunraj \textit{et al.}~\cite{gunraj2023soldernet} presented SolderNet, a deep learning-driven system for inspecting solder joints in electronics manufacturing, which incorporates XAI to explain its predictions and enhance trust.

Overall, integrating XAI into inspection systems greatly enhances transparency, adaptability, and user-friendliness, resulting in improved performance and collaboration between human experts and AI. As the field evolves, adopting XAI in industrial visual quality inspection is anticipated to drive more efficient and reliable quality control processes across various industries. \cite{hoffmann2023systematic}. 

\subsection{Edge Explainable AI (XEdgeAI)}
The fusion of \ac{XAI} with edge computing, known as Edge Explainable AI (XEdgeAI), marks a significant paradigm shift with the potential to transform a wide range of industries. This innovation aims to make automated systems more transparent, understandable, and trustworthy.

Recent studies have illustrated the potential of XEdgeAI in various fields. For instance, Kok \textit{et al.}~\cite{kok2023explainablegreen} developed an XAI-powered edge computing solution for optimizing energy management in smart buildings. This system provides explanations of energy usage patterns and decision-making processes, enabling building managers to implement informed energy-saving changes. In another study, Garg \textit{et al.}~\cite{garg2023trusted} addressed the challenges of building trust in AI systems within a 6G edge cloud environment, highlighting the critical role of transparency and interpretability in edge computing. In the healthcare domain, Dutta \textit{et al.}~\cite{dutta2023human} designed a human-centered XAI application for edge computing to ensure that healthcare professionals can understand and act on AI-driven decisions. This approach has great potential to enhance the quality and reliability of healthcare services.

While XEdgeAI has made significant progress, it still faces several challenges such as limited resources and security concerns. Therefore, it is essential to integrate XAI into both edge and cloud environments, as each plays a critical role in developing intelligent systems. As researchers continue exploring the potential of XEdgeAI, this field will likely shape the future of AI across various industries. Applications such as industrial predictive maintenance and quality inspection in edge computing environments are expected to benefit greatly from these advancements.

\subsection{Human-centered XAI with \acp{LVLM}}
Researchers have employed various innovative strategies to make AI systems more transparent and comprehensible to users without a background in AI or XAI. These strategies include aligning AI explanations with human psychology~\cite{wang2019designing}, simplifying algorithms~\cite{yu2023towards}, providing interactive explanations~\cite{bertrand2023selective}, and offering textual explanations~\cite{poli2021generation,park2018multimodal,hendricks2018grounding,xu2015show,kim2018textual}.

Recent advancements in \ac{LLM} have led to the development of \acp{LVLM}, which blends language understanding with vision encoding and reasoning. These models excel in tasks such as image captioning, document understanding, visual question answering, and multi-modal in-context learning~\cite{dai2024instructblip,brown2020language,chowdhery2023palm,peng2023kosmos,awadalla2023openflamingo,fuyu-8b,chen2022pali,gpt4vision,dong2024internlm,zhu2023minigpt}. This progress introduces new opportunities for integrating LVLMs to generate textual explanations for visual perception tasks, thereby enhancing explainability~\cite{nguyen2024langxai}. As a good example~\cite{chen2023x}, X-IQE evaluates text-to-image generation methods by generating textual explanations using a hierarchical Chain of Thought (CoT) within MiniGPT-4 as the base LVLM. On the other hand, the LangXAI framework~\cite{nguyen2024langxai} demonstrates how integrating LVLMs can improve the understandability of visual AI systems. By generating textual explanations for saliency maps, LangXAI also emphasizes the importance of user-centric design. 

\section{Methodology}\label{s:methodology}

\begin{figure*}[htbp]
    \centering
    \includegraphics[width=\linewidth]{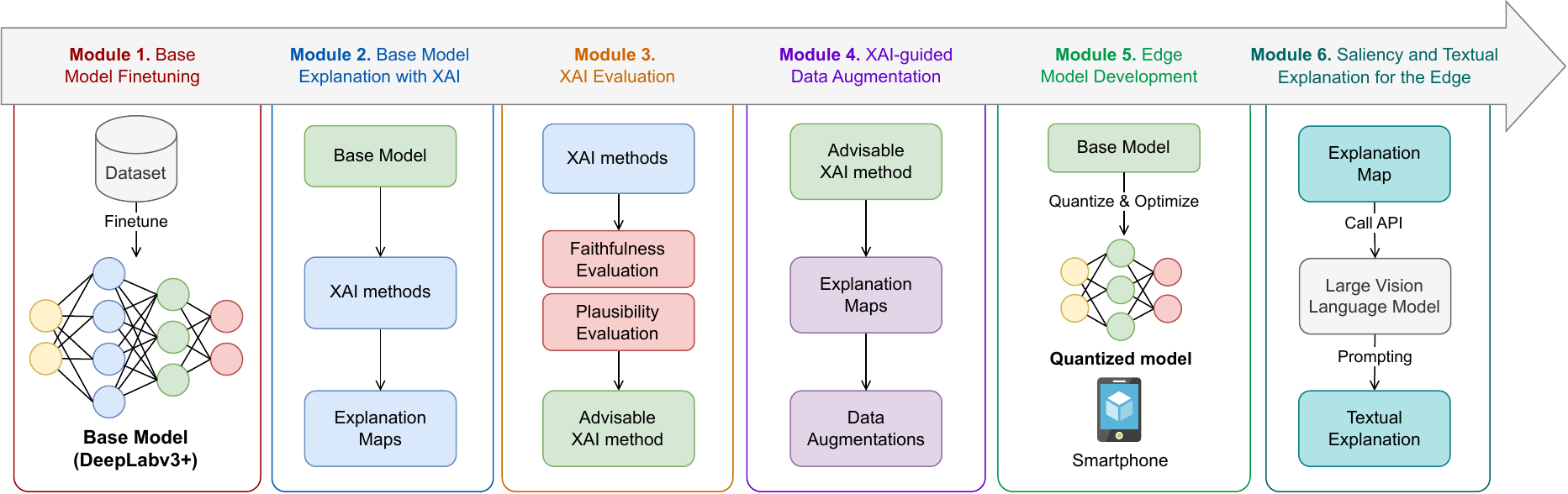}
    \caption{The methodology of the XAI-integrated Visual Quality Inspection framework integrated with XAI methods with 6 building modules: (1) Base Model Finetuning with a provided visual quality dataset, (2) Base Model Explanation with XAI, (3) XAI Evaluation, (4) XAI-guided Data Augmentation to improve the base model performance, (5) Edge Model Development on mobile devices and (6) Saliency and textual Explanation for the Edge. The end-users interact with the framework via a mobile application, while the domain experts can interact via a web application.}
    \label{fig:evqi}
\end{figure*}

Building upon the advancements of segmentation for visual quality inspection, XAI-based model improvement, and human-centered explanations, we propose a novel XAI-integrated Visual Quality Inspection framework that enhances the interpretability and efficiency of visual quality inspection models on edge devices.
Our framework integrates state-of-the-art semantic segmentation models, such as DeepLabv3Plus~\cite{yang2023semantic}, with advanced XAI methods to generate highly plausible and faithful explanations for the model's decisions. 
Furthermore, we introduce an XAI-guided data augmentation module that leverages expert knowledge to improve the base model's performance, addressing the need for continuous model refinement in industrial settings.
To ensure the framework's applicability in real-world scenarios, we focus on developing an efficient edge model that can be deployed on mobile devices, enabling on-site visual quality inspections. 
Additionally, we incorporate a human-centric explanation module that generates both saliency maps and textual explanations using \acp{LVLM}, such as GPT-4 Vision \cite{gpt4vision}, to make the inspection results more accessible and understandable to end-users.

Our proposed XAI-integrated Visual Quality Inspection framework consists of six main modules as illustrated in Figure \ref{fig:evqi}: base model finetuning, base model explanation with XAI, XAI evaluation, XAI-guided data augmentation, edge model development, and saliency and textual explanation for the edge. 

\begin{enumerate}
    \item \textbf{Base Model Finetuning:} In the first module, we prepare the visual quality dataset and finetune a semantic segmentation model, DeepLabv3Plus, to serve as our base model. The dataset is preprocessed and split into training and validation sets. The model is trained using the Dice loss function, which is well-suited for imbalanced classes in image segmentation tasks.
    \item \textbf{Base Model Explanation with XAI:} The second module focuses on explaining the base model using various XAI methods, including GradCAM, GradCAM++, XGradCAM, HiResCAM, ScoreCAM, AblationCAM, GradCAMElementWise, EigenCAM, EigenGradCAM, and RISE. These methods generate saliency maps highlighting the regions in the input image that have the highest influence on the model's segmentation decision.
    \item \textbf{XAI Evaluation:} In the third module, we evaluate the XAI methods using plausibility and faithfulness metrics. Plausibility is assessed using the Energy-Based Pointing Game (EBPG), Intersection over Union (IoU), and Bounding Box (Bbox), which measure how well the explanations align with human intuition. Faithfulness is evaluated using Deletion and Insertion metrics, which quantify the alignment between the explanations and the model's predictive behavior. Based on these evaluations, the most suitable XAI method is selected.
    \item \textbf{XAI-guided Data Augmentation:} The fourth module involves XAI-guided data augmentation. The chosen XAI method is used to guide the annotation augmentation process, where the training set annotations are relabeled based on expert recommendations. The model is then retrained on the enhanced training dataset to demonstrate the potential of annotation augmentation in improving semantic segmentation models.
    \item \textbf{Edge Model Development:} In the fifth module, we develop the edge model by quantizing and optimizing the improved base model for deployment on mobile devices. Dynamic quantization is applied to specific layers to reduce model size, and the model is converted to TorchScript format for efficient execution. Additional optimizations are performed to enhance the model's performance on mobile devices.
    \item \textbf{Saliency and Textual Explanation for the Edge:} Finally, in the sixth module, we generate visual explanations as saliency maps and textual explanations for the segmentation results obtained by the mobile model on edge devices. The chosen XAI method is used to generate the saliency maps. At the same time, a \ac{LVLM}, such as GPT-4 Vision, is utilized to generate human-readable textual explanations based on the segmented image and explanation map.
\end{enumerate}

By following this methodology, our framework aims to provide a high-plausible, interpretable, and efficient visual quality inspection solution for industrial assets that can be effectively deployed on edge devices.

\section{Implementation}\label{s:implementation}
This section presents the implementation details of our proposed XAI-integrated Visual Quality Inspection framework, comprising six main modules. 
We describe the data preparation, base model finetuning, XAI methods, evaluation metrics, XAI-guided data augmentation, edge model development, and the generation of visual and textual explanations for the mobile model on edge devices. 
The implementation leverages and evaluates state-of-the-art techniques, such as the DeepLabv3Plus model, various XAI methods, and the \ac{LVLM}, to create an interpretable and efficient visual quality inspection system.

\begin{figure*}[h]
    \centering
    \includegraphics[width=\linewidth]{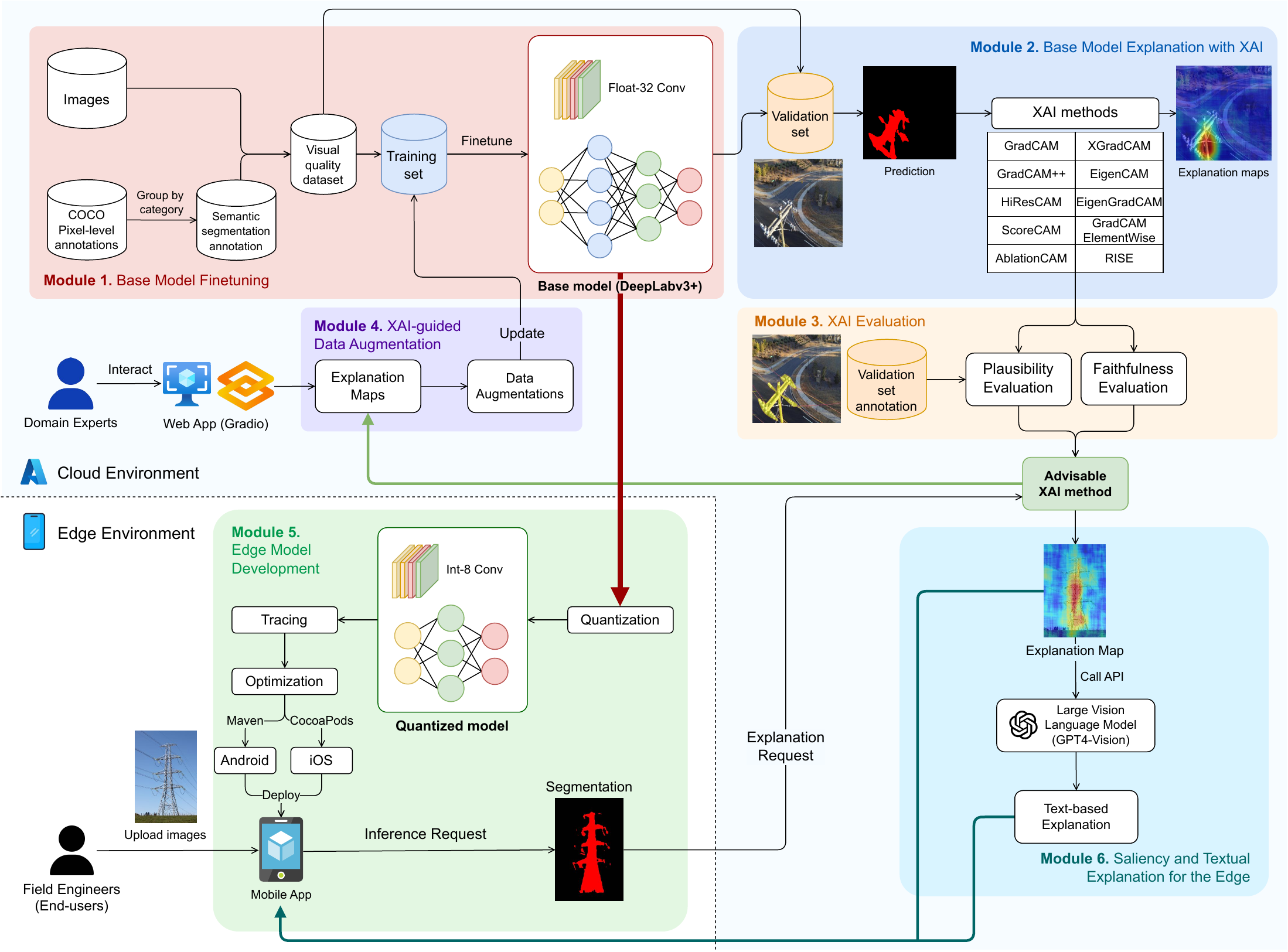}
    \caption{The implementation details of our proposed XAI-integrated Visual Quality Inspection framework.}
    \label{fig:imp}
\end{figure*}

\subsection{Module 1 – Base Model Finetuning}
In this module, we present the process of preparing the visual quality dataset to build a base model by finetuning the semantic segmentation model.

\subsubsection{Data Preparation and Preprocessing}
The acquired dataset \(\mathcal{D}\) contains the image set \(\mathcal{I}\) and the corresponding annotation set \(\mathcal{A}\). 

The varying sizes of images are handled by being dynamically adjusted, where image \(I \in \mathcal{I}\) is processed through two primary transformations: conversion to a tensor and normalization.
Initially, we convert an image from its native representation, where pixel values are in \([0, 255]\), to a tensor format with values normalized to \([0, 1]\), using \(I'_{cij} = \frac{I_{cij}}{255}\) for each pixel \(I_{cij}\) in channel \(c\).
Subsequently, we apply channel-wise normalization to this tensor, adjusting each pixel value to zero mean and unit variance by the formula:
\begin{equation}
    I''_{cij} = \frac{\left(\frac{I_{cij}}{255}\right) - \mu_c}{\sigma_c}
\end{equation}
where: \( \mu = [0.485, 0.456, 0.406] \) are the mean values for the RGB channels,
\( \sigma = [0.229, 0.224, 0.225] \) are the standard deviation values for the RGB channels.

The corresponding \ac{COCO} annotations are stored in JSON format, where the annotated masks are supported for the semantic segmentation task.
These masks are grouped by object categories and generated by drawing polygons around the specified objects, and subsequently used as ground truth for model training. 
In detail, the mask \( M_{ij}: \mathbb{R}^2 \rightarrow \{0,1\} \) indicates the presence (\(1\)) or absence (\(0\)) of an object \( j \) in image \( i \), based on the polygon coordinates provided in \( \mathcal{A}_i \).

We divide the original dataset \( \mathcal{D} \)  into an 80\%-20\% training \( \mathcal{D}_{\text{train}} \) and validation \( \mathcal{D}_{\text{val}} \) sets, with all images  \( I \in \mathcal{I} \) resized to $700 \times 700$ pixels, with the resizing operation \( R(I) \), such that \( R: \mathbb{R}^{H \times W \times C} \rightarrow \mathbb{R}^{700 \times 700 \times C} \). 
The training set \( \mathcal{D}_{\text{train}} \) is used to finetune the model. In contrast, the validation set \( \mathcal{D}_{\text{val}} \) is leveraged to extract the explanation for the model's prediction, evaluate the XAI methods' performance, and support the domain expert to identify any data error.

\subsubsection{Model Finetuning}
In this section, we describe the finetuning process for the base visual quality inspection model $\Theta$. 
The fine-tuning of the DeepLabv3Plus model is conducted using PyTorch.

\begin{figure}[h]
    \centering
    \includegraphics[width=\linewidth]{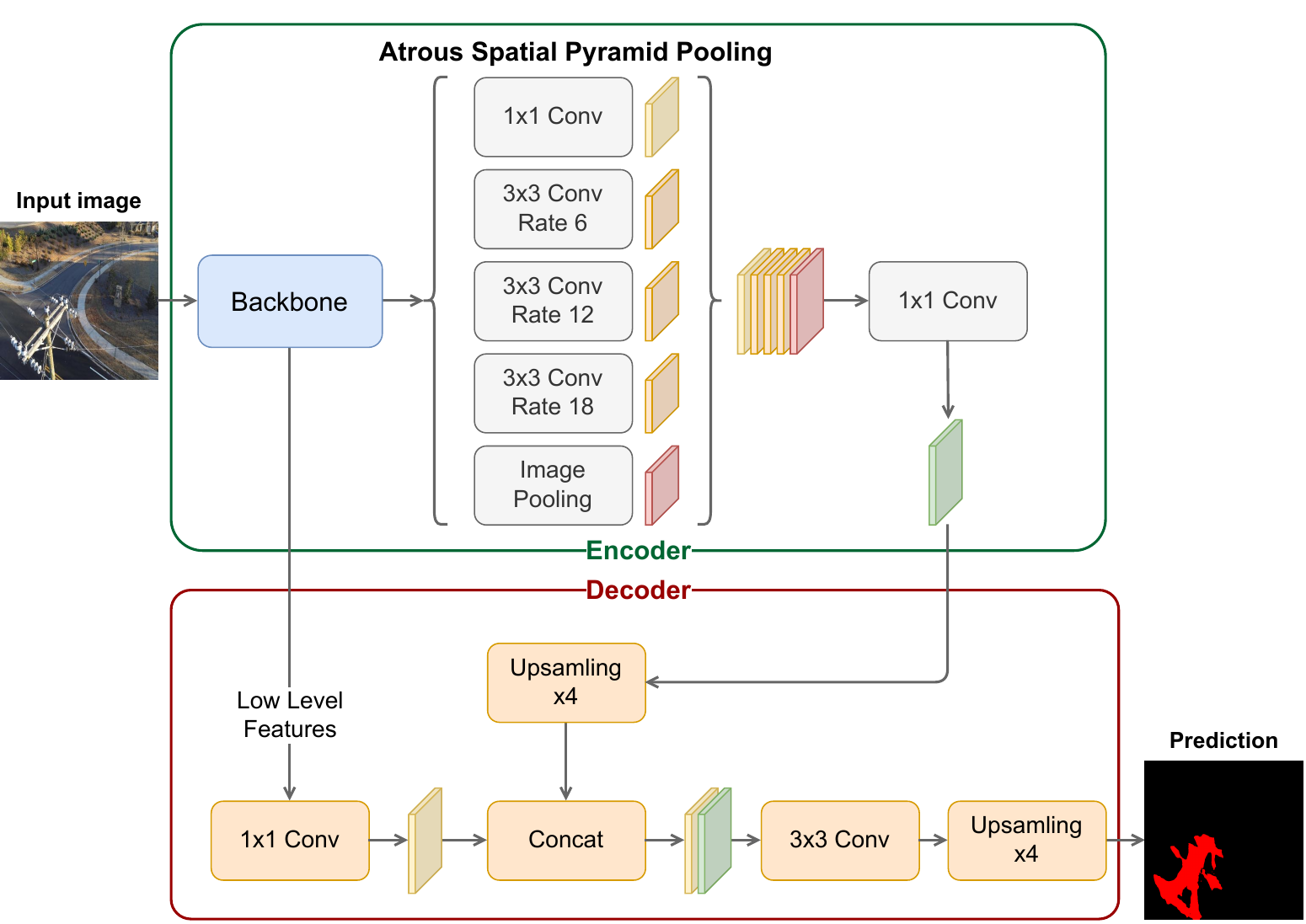}
    \caption{The architecture of the DeepLabv3Plus model, featuring an encoder with Atrous Spatial Pyramid Pooling and a decoder for upsampling and refining segmentation outputs.}
    \label{fig:deeplabv3p}
\end{figure}

\paragraph{Model Preparation}
The DeepLabv3Plus architecture is set as the base model $\Theta$, where its architecture is illustrated in Figure~\ref{fig:deeplabv3p}.
This model combines the strengths of encoder-decoder architecture with atrous separable convolution to enhance segmentation precision, particularly along object boundaries.

The encoder part of DeepLabv3Plus adopts a deep \ac{CNN} model as the backbone, such as ResNet \cite{he2016deep}, or MobileNet~\cite{howard2017mobilenets}, which is modified for segmentation tasks by incorporating depthwise separable convolutions that reduce the number of parameters and computational costs while maintaining effective feature extraction.
The ImageNet pre-trained weights~\cite{deng2009imagenet} are used for encoder initialization.
The atrous convolution allows the network to grasp contextual information at multiple scales through the \ac{ASPP} module without increasing computational demand.

The decoder module focuses on refining the segmentation outputs, which is critical for achieving high-resolution and accurate boundary representation. By gradually restoring spatial information and enhancing feature resolution, the decoder aids in producing more precise segmentation maps.
The weights and activations of the convolution layers are in the Single-precision floating-point format (float32). 

\paragraph{Loss Function and Optimization}
The Dice loss function is used for training the model, which is particularly useful for imbalanced classes in the image segmentation task, as it considers the overlap between the predicted and ground truth masks~\cite{sudre2017generalised}.
Let $\text{P}$ and $\text{GT}$ represent predicted and ground truth masks, respectively. The Dice loss \( \mathcal{L}_{\text{Dice}} \) is defined as:

\begin{equation}
    \mathcal{L}_{\text{Dice}}(\text{P}, \text{GT}) = 1 - \frac{2 \times |\text{P} \cap \text{GT}|}{|\text{P}| + |\text{GT}|}
\end{equation}

The accuracy of segmentation models is assessed using the IoU metric, also known as the Jaccard coefficient~\cite{murphy1996finley}, one of the most commonly used metrics in semantic segmentation. 
The IoU metric is a prevalent evaluation measure in semantic segmentation. It quantifies the overlap between the predicted segmentation and the ground truth for each class. Mathematically, for the predicted mask $\text{P}_c$ and the ground truth mask $\text{GT}_c$ of a category, the IoU is calculated as follows:

\begin{equation}
    \text{IoU}_c = \frac{\text{P}_c \cap \text{GT}_c}{\text{P}_c \cup \text{GT}_c}
\end{equation}

The model is optimized using the Adam optimizer \cite{kingma2014adam} with an initial learning rate $\alpha=0.0001$. $\alpha$ is halved to $1\mathrm{e}{-5}$ after 25 epochs to facilitate convergence as the model approaches optimal performance. The batch size is set to 8, balancing computational efficiency and memory constraints during training.

\subsection{Module 2 – Base Model Explanation with XAI}\label{ss:imp-base-model-xai}
We implement several XAI methods in this module to explain the semantic segmentation model. The explanation maps of all methods are extracted from the predictions of the segmentation model on the validation set, which will be used for the evaluation step.
We utilize several notable CAM-based XAI methods, such as GradCAM~\cite{selvaraju2017grad}, GradCAM++~\cite{chattopadhay2018grad}, XGradCAM~\cite{fu2020axiom}, HiResCAM~\cite{draelos2020hirescam}, ScoreCAM~\cite{wang2020score}, AblationCAM~\cite{ramaswamy2020ablation}, GradCAMElementWise~\cite{jacobgilpytorchcam}, EigenCAM~\cite{muhammad2020eigen}, EigenGradCAM~\cite{muhammad2020eigen} due to their applicabilities and plausibility in the semantic segmentation task. 
Besides, we also leverage a perturbation-based XAI method, namely RISE~\cite{petsiuk2018rise}, due to its model-agnostic mechanism.

The resulting saliency map \(L^c\) of a specific category \(c\) highlights the regions in the input image with the highest influence on the model's segmentation decision. The saliency map is typically represented as a heatmap, where warmer colors indicate regions of higher importance, and cooler colors represent regions of lower importance.

\subsection{Module 3 – XAI Evaluation}\label{ss:imp-xai-eval}
This component evaluates the \ac{XAI} methods with plausibility and faithfulness metrics on their explanations of the models with the validation set $\mathcal{D}_{\text{val}}$. Plausibility measures how well the explanations align with human intuition and understanding, while faithfulness measures how accurately the explanations reflect the underlying model’s decision-making process. By evaluating both plausibility and faithfulness, we can ensure that the chosen XAI method provides explanations that are both understandable to humans and accurately represent the model’s behavior. Eventually, the method achieving the highest scores in most metrics will be chosen as the core XAI method of the model enhancement step.
In the following, we introduce two relevant metrics, including the plausibility and faithfulness of XAI explanations. 

\subsubsection{Plausibility Evaluation Metrics} 
Plausibility, the alignment of explanations with human intuition, is assessed using measures like Energy-Based Pointing Game (EBPG)~\cite{wang2020score}, and \ac{IoU}~\cite{zhou2016learning}, and Bounding Box (Bbox)~\cite{schulz2020restricting}.
Based on human annotations, these measures validate the model by assessing the statistical superiority of explanations. 
\begin{itemize}
    \item \textbf{Energy-Based Pointing Game (EBPG):} evaluates the precision and denoising ability of XAI methods to identify the most influential region in an image for a given prediction~\cite{wang2020score}. It calculates how much the energy of the saliency map by pixel $L^c{(i, j)}$ falls inside the ground truth. A good explanation is considered to have a higher EBPG. EBPG formula is defined as follows:
    \begin{equation}
        \text{EBPG} = \frac{\sum L^c_{(i, j) \in \text{GT}}}{L^c_{(i, j) \in \text{GT}} + L^c_{(i, j) \notin \text{GT}}}
    \end{equation}

    \item \textbf{\ac{IoU}}~\cite{zhou2016learning} assesses the localization capability and the significance of the attributions captured in an explanation map. It measures the overlap between the saliency map and the ground truth annotation. IoU is defined as:
    \begin{equation}
    \text{IoU} = \frac{\text{Area}(L^c \cap \text{GT})}{\text{Area}(L^c \cup \text{GT})}
    \end{equation}

    \item \textbf{Bounding Box (Bbox)}~\cite{schulz2020restricting} is a variant of the \ac{IoU} metric that adapts to the size of the object of interest. It measures the overlap between the bounding box of the saliency map and the ground truth bounding box. Bbox is defined as:
    \begin{equation}
    \text{Bbox} = \frac{\text{Area}(\text{BBox}(L^c) \cap \text{BBox}(\text{GT}))}{\text{Area}(\text{BBox}(L^c) \cup \text{BBox}(\text{GT}))}
    \end{equation}
\end{itemize} 

\subsubsection{Faithfulness Evaluation Metrics}
Faithfulness, the alignment of explanations with the model's predictive behavior, is evaluated using the Deletion and Insertion metrics~\cite{petsiuk2018rise,samek2016evaluating,hooker2019benchmark}. These measures quantify the degree to which the explanations align with the predictive behavior of the model. 

Given a black box model \(\Theta\), input image \(I\), saliency map \(L^c\), and number of pixels \(N\) removed per step.
\begin{itemize}
\item \textbf{Deletion} measures the accuracy of saliency areas by removing pixels from the input image in order of saliency, from large to small. More accurate saliency areas will have a steeper deletion curve, and a smaller deletion metric value indicates higher accuracy. Deletion $d_{\text{Del}}$ is defined as:
\begin{equation}
d_{\text{Del}} = \text{AreaUnderCurve}(h_i \text{ vs. } i/n, \forall i = 0,...n) 
\end{equation}
where
\(h_i \gets f(I)\) while \(I\) has non-zero pixels, and according to \(L^c\), setting the next \(N\) pixels in \(I\) to 0 each iteration until  \(n\) iterations.
 
\item \textbf{Insertion} assesses the comprehensiveness of the saliency area by removing all pixels from the input image and then recovering them in order of saliency, from large to small. A more comprehensive saliency area will require fewer pixels to recover the object and have a faster-rising insertion curve. A higher insertion metric value indicates a more comprehensive saliency area. The Insertion $d_\text{Ins}$ is defined as:
\begin{equation}
d_\text{Ins} = \text{AreaUnderCurve}(h_i \text{ vs. } i/n, \forall i = 0,...n)
\end{equation}
where \(I' \gets \text{Blur}(I)\) and \(h_i \gets f(I)\) while $I \neq I'$, and according to $L^c$, setting the next $N$ pixels in $I'$ to the corresponding pixels in $I$ each iteration until $n$ iterations.
\end{itemize}

Finally, we choose the most suitable XAI method $\mathcal{X}$ based on the evaluation results with the validation set $\mathcal{D}_{\text{val}}$ and their capabilities for running on the mobile model $\theta$.

\subsection{Module 4 – XAI-guided Data Augmentation}\label{ss:imp-xai-data-aug}
Data augmentation strategies, such as altering data distribution or adjusting data and labels, have been employed to enhance model performance~\cite{bento2021improving}. 
In this module, we leverage the advisable XAI method $\mathcal{X}$, demonstrating the highest faithfulness and plausibility from the XAI evaluation step, to guide the annotation augmentation process.

\begin{figure*}
    \centering
    \includegraphics[width=\linewidth]{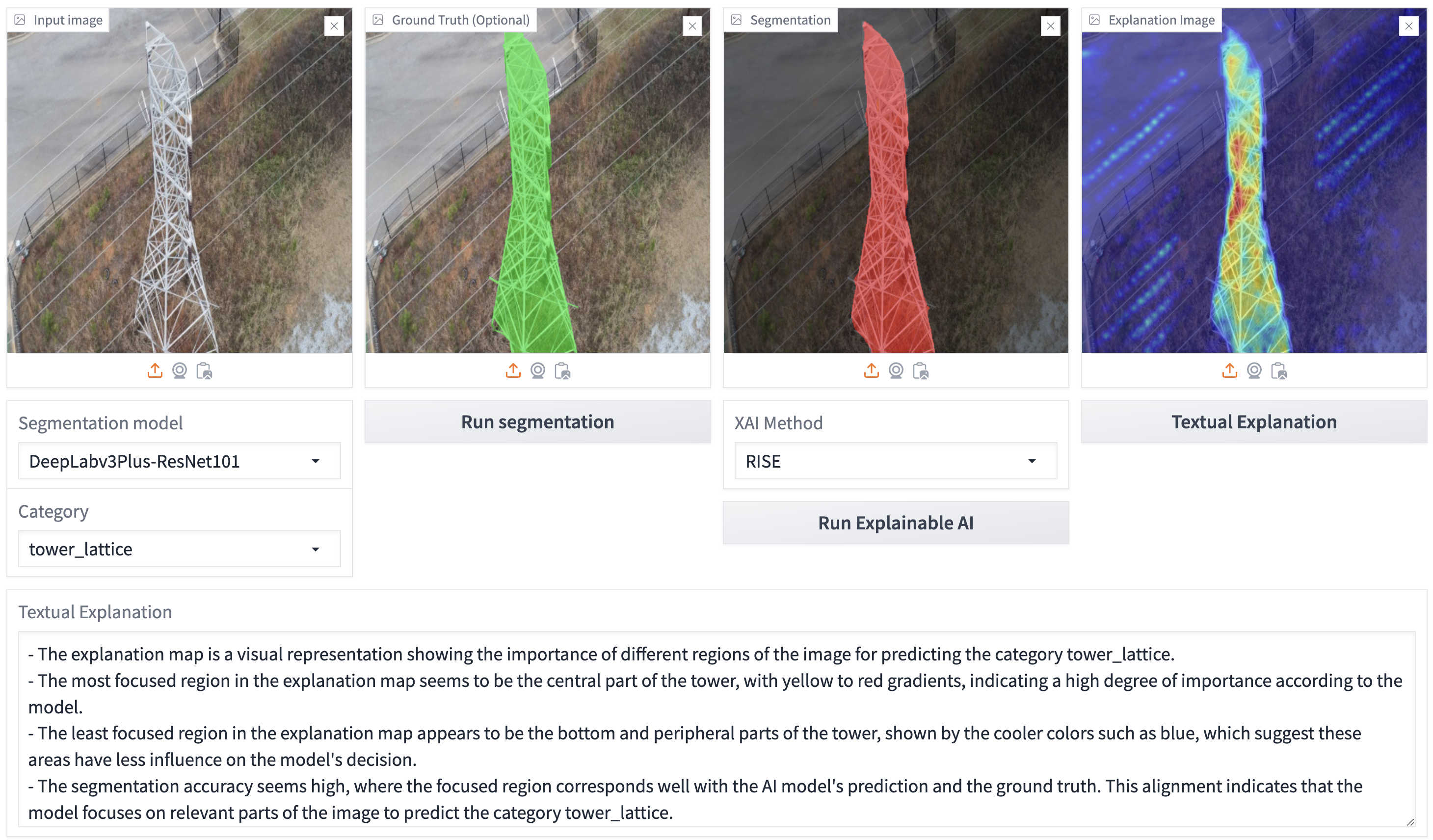}
    \caption{Web-based user interface for domain experts to monitor predictions, explanations, and textual explanations. The input image, segmentation output, saliency map, and textual explanation are displayed, allowing experts to assess the model's performance and provide feedback for annotation augmentation.}
    \label{fig:web-ui}
\end{figure*}

To facilitate the augmentation process, we develop a web-based user interface that allows domain experts to load any sample from the dataset and monitor the model's predictions, explanations in the form of saliency maps, and textual explanations. Figure~\ref{fig:web-ui} illustrates the user interface, which consists of several components. The input image is displayed on the left, while the segmentation output is shown in the center. The saliency map, generated by the advisable XAI method $\mathcal{X}$, is presented on the right, highlighting the regions that contribute most to the model's prediction. At the bottom, a textual explanation is provided, offering a human-readable interpretation of the model's decision-making process.
The web-based user interface plays a crucial role in this module, enabling domain experts to closely examine the model's behavior and provide informed recommendations for annotation augmentation.

Using this web UI, domain experts can thoroughly examine the model's predictions and explanations, identifying problems on the $\mathcal{A}_{\text{val}}$ and defining solutions for $\mathcal{A}_{\text{train}}$ need to be refined. Based on their expertise and the insights gained from the explanations, the experts provide recommendations for relabeling the annotations. These recommendations are then used to augment the training set, resulting in an enhanced dataset with improved annotations $\tilde{\mathcal{A}}_{\text{train}}$.

Subsequently, the base model $\Theta$ is retrained on the augmented training dataset $\tilde{\mathcal{A}}_{\text{train}}$ to achieve the enhanced model $\tilde{\Theta}$. To assess the impact of the XAI-guided data augmentation, we evaluate the enhanced model $\tilde{\Theta}$ on the original validation set $\mathcal{D}{\text{val}}$ by comparing the performance of the base model $\Theta$ before and after applying the data augmentation. Hence, we can demonstrate the potential of annotation augmentation, supported by XAI explanations, in enhancing semantic segmentation models.

\subsection{Module 5 – Edge Model Development}
In this subsection, we will explore the process of transferring the original model to edge devices. Given the limited computational resources of mobile appliances, adaptations are needed to ensure smooth compatibility. After optimization, the model undergoes various developments to be deployed on platforms like Android and iOS. Detailed descriptions of these processes are presented in the subsequent parts.

\subsubsection{Model Quantization, Pruning and Optimization for Mobile Devices}
After acquiring the enhanced model $\tilde{\Theta}$, we apply quantization, pruning, and optimization techniques to convert it into a mobile model $\theta$ that can efficiently run on smartphone devices.
Algorithm~\ref{alg:model-optimization} summarizes the entire model quantization, pruning, and optimization process for mobile deployment.

The first step is to disable batch normalization layers in the base model. We iterate over all the modules in the model and set the batch normalization layers to evaluation mode. This step is necessary to ensure that the model's statistics remain fixed during quantization.
Next, we apply dynamic quantization to the base model $\Theta$. Dynamic quantization is a technique that reduces the numerical precision of the model's weights and activations, thereby decreasing the model size and improving inference speed without significantly compromising accuracy. We target specific layers that are known to consume a substantial amount of computational resources, including the 2D convolutional layer (Conv2d), linear layer (Linear), rectified linear unit (ReLU), 2D batch normalization (BatchNorm2d), and 2D adaptive average pooling (AdaptiveAvgPool2d). We considerably reduce the model size by quantizing these layers' parameters to 8-bit integers using the dynamic quantization function from PyTorch, making it more suitable for deployment on devices with limited storage and processing capabilities.
We perform a forward pass on the model using a sample input tensor to simulate the inference process and ensure the quantized model functions correctly.

After quantization, we apply pruning to the model. Pruning is a technique that removes less important weights from the model, reducing its size and computational requirements. We iterate over the named modules of the base model and apply structured pruning to the convolutional layers (Conv2d). We set the pruning amount to 0.1, indicating that 10\% of channels in each convolutional layer will be pruned.
To remove the pruning re-parameterizations and obtain the final pruned model, we iterate over the named modules again and remove the pruning masks.

Next, we trace the pruned model using the just-in-time (JIT) compiled tracing function, which converts the PyTorch model into a TorchScript representation. TorchScript is a static graph representation that allows for optimizations and efficient execution on various platforms, including mobile devices. We pass a sample input tensor to the tracing function to capture the model's computational graph.

Finally, we apply mobile-specific optimizations to the traced model using the mobile optimization function. This function performs a series of optimizations tailored for mobile environments, such as operator fusion, constant folding, and dead code elimination. These optimizations help to reduce the model size further and improve its inference speed on mobile devices.

The resulting mobile model $\theta$ is now quantized, pruned, and optimized for deployment on smartphone devices. The quantization process reduces the model's memory footprint, the pruning process removes redundant weights, and the mobile-specific optimizations enhance the model's efficiency during inference.

\begin{algorithm}[h!]
\SetAlgoLined
\SetArgSty{textup}
\DontPrintSemicolon
\KwIn{base\_model $\Theta$: a base model }
\KwIn{input\_tensor: an input tensor}
\KwOut{mobile\_model $\theta$: an optimized, quantized, and pruned model}

{\small \tcc{Disable batch normalization}}
\For{module in base\_model.modules()}{
\If{isinstance(module, nn.BatchNorm2d)}{
module.eval()\;
}
}

{\small \tcc{Apply dynamic quantization}}
base\_model = torch.quantization.quantize\_dynamic(base\_model, \{nn.Conv2d, nn.Linear, nn.ReLU, nn.BatchNorm2d, nn.AdaptiveAvgPool2d\}, dtype=torch.qint8)\;

{\small \tcc{Forward pass to simulate inference}}
base\_model(input\_tensor)\;

{\small \tcc{Apply pruning}}
\For{name, module in base\_model.named\_modules()}{
\If{isinstance(module, nn.Conv2d)}{
prune.ln\_structured(module, name=`weight', amount=0.1, n=1, dim=0)\;
}
}

{\small \tcc{Remove pruning reparameterizations}}
\For{name, module in base\_model.named\_modules()}{
\If{isinstance(module, nn.Conv2d)}{
prune.remove(module, `weight')\;
}
}

{\small \tcc{Trace the model}}
traced\_model = torch.jit.trace(base\_model, input\_tensor)\;

{\small \tcc{Optimize the traced model for mobile}}
mobile\_model = optimize\_for\_mobile(traced\_model)\;

\Return mobile\_model\;
\caption{Model Quantization, Pruning, and Optimization for Mobile}
\label{alg:model-optimization}
\end{algorithm}

\subsubsection{Mobile Model Deployment}
After acquiring the optimized mobile model $\theta$, we proceed to the deployment process of the smartphone application, which incorporates the mobile model $\theta$ for both Android and iOS platforms. The deployment workflow encompasses packaging the model and its dependencies and integrating them with the mobile app. 
For Android devices, we leverage Maven, which is the process of compiling the app, bundling the mobile model, and generating an Android Package (APK) or Android App Bundle (AAB) for distribution, which is automated. 
While for iOS devices, we utilize CocoaPods, a dependency manager for Swift and Objective-C projects. Subsequently, the app is built and packaged using Xcode's build system.

Once the mobile app is installed on the end users' devices, field engineers can utilize it to capture real-world images and request semantic segmentation using the integrated mobile model \(\theta\). The app provides an interface that guides field engineers to capture images, which leverages the device's camera capabilities. During the inference process, the mobile model generates a segmentation mask overlaying on the uploaded picture \(\tilde{I}\) as the prediction on the edge \(y^{\theta}_{\tilde{I}}\) that identifies the visual quality objects being inspected in the image.

\subsection{Module 6 – Saliency and Textual Explanation for the Edge}\label{ss:imp-xai-edge}
In this section, we present the process of generating visual explanations as saliency maps and textual explanations for the segmentation results obtained by the mobile model \(\theta\) on edge devices. 

The visual explanations provide insights into the model's decision-making process, highlighting the important regions in the input image that contribute to the segmentation output. 
To generate the saliency map for the edge, the app first performs a forward pass of the input image through the mobile model \(\theta\) to obtain the segmentation output. 
The chosen XAI method \(\mathcal{X}\) receives the uploaded image \(\tilde{I}\), segmentation output \(y^{\theta}_{\tilde{I}}\), the detected category \(c\) and interaction with the model \(\theta\) to generate the explanation map.

The textual explanations, generated using LVLMs, offer a human-readable interpretation of the segmentation results, enhancing the interpretability and trustworthiness of the mobile app for field engineers.
In this research, we employ a recent member of the LVLMs family, GPT-4 Vision \cite{gpt4vision}, as the core vision language model due to its robust performance across diverse tasks \cite{r2-50} and a strong alignment with human evaluators \cite{r2-51}.
This LVLM processes a designed prompt, the segmented image, and the explanation map utilizing its pre-trained knowledge to generate human-readable explanations. 
The model's responses are based on its understanding of the visual content and its ability to associate relevant textual descriptions.
The generated explanations provide a concise and intuitive summary of the segmentation results, highlighting the main objects the model concentrates on, their attributes, and their relationships within the image.

Finally, the textual explanations are displayed to the field engineer alongside the segmented image and the saliency map. This combination of visual and textual explanations enhances the interpretability of the segmentation results, allowing field engineers to understand the model's behavior better and the rationale behind its predictions.

\begin{figure*}
\centering
\includegraphics[width=\linewidth]{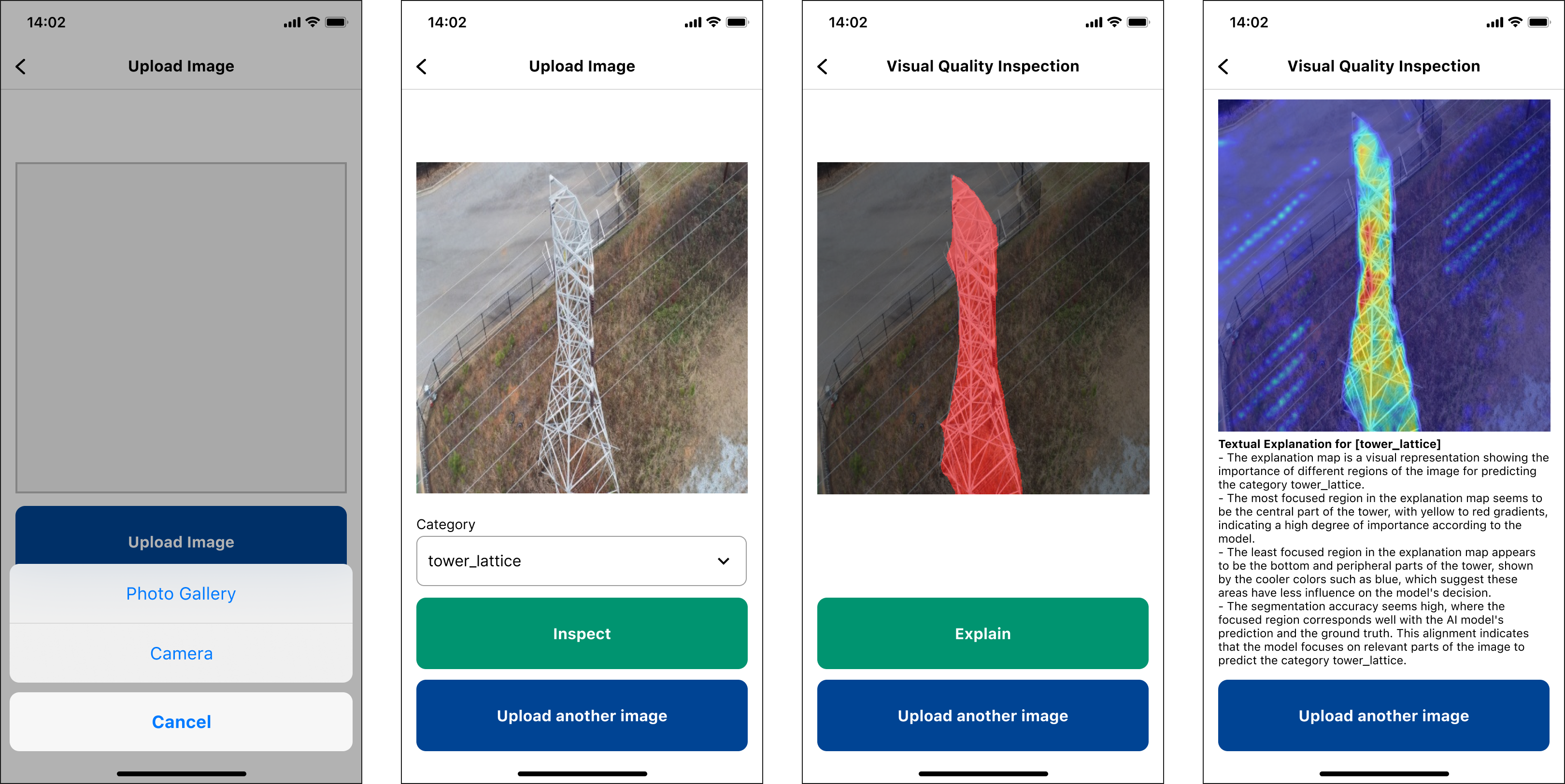}
\caption{The iOS user interface of the mobile application for end-users, designed for iPhone 11 Pro. Users can upload an image, initiate visual quality inspection for a particular category, view the segmentation results along with explanations, and have the option to upload another image.}
\label{fig:mobile-ui}
\end{figure*}

Figure~\ref{fig:mobile-ui} illustrates the iOS user interface of the mobile application designed for iPhone 11 Pro. The main screen allows users to upload an image by selecting from the photo gallery or capturing a new image using the device's camera. Upon uploading an image, the user can initiate the visual quality inspection process by choosing the object's category being inspected and tapping the ``Inspect" button.
The mobile model $\theta$ then performs semantic segmentation on the uploaded image $\tilde{I}$, generating a segmentation mask $y^{\theta}_{\tilde{I}}$ that identifies the visual quality objects being inspected. The segmentation mask is overlaid on the original image, providing a clear visual representation of the detected objects.

\section{Experiment 1: A Comprehensive Evaluation}\label{s:exp}
This section details the experimental setup, results, and analysis of our XAI-integrated Visual Quality Inspection framework, which is applied to an industrial hardware assets dataset for inspecting transmission towers and power lines using aerial imagery. We begin by training the base DeepLabv3Plus model with different backbones and assessing its performance on the validation set. Next, we conduct a series of comprehensive analyses to identify the optimal explaining method for implementation. Using this method, we apply data augmentation techniques to enhance the model's performance and generate textual explanations that provide human-understandable insights into the model's decision-making process. During the performance comparison stage, we consider the base model alongside the enhanced and mobile models.

\subsection{Dataset}
This experiment uses the TTPLA dataset for detecting and segmenting power-grid hardware components from aerial imagery~\cite{abdelfattah2020ttpla}. 
The dataset encompasses 1242 high-resolution aerial images featuring 8987 instances of transmission towers and power lines. 
These instances are classified into four distinct categories: \textit{cable}, \textit{tower\_wooden}, \textit{tower\_lattice}, and \textit{tower\_tucohy}, each representing a specific type of power-grid infrastructure component. Table \ref{tab:ttpla-detailed} summarizes the object characteristics by class, detailing counts, area percentages, and dimension ranges (width and height) for analysis.

\begin{table}[ht]
    \centering
    \resizebox{\linewidth}{!}{%
    {\rowcolors{2}{white}{gray!10}
    \begin{tabular}{lr>{\raggedleft\arraybackslash}p{2.2cm}>{\raggedleft\arraybackslash}p{2.1cm}>{\raggedleft\arraybackslash}p{2.1cm}}
    \toprule
    \multirow{2}{*}{\textit{Class}} & \multirow{2}{*}{\#Obj.} & Min/Max/Avg Area (\%) & Min/Max/Avg Width (px) & Min/Max/Avg Height (px) \\
    \midrule
    \textit{cable}           & 10082 & 0.00/9.88/0.26 & 4/3840/1181 & 4/2160/766 \\
    \textit{tower\_lattice}  & 404   & 0.04/58.28/7.22 & 34/3840/875 & 59/2160/1226 \\
    \textit{tower\_wooden}   & 333   & 0.01/43.61/2.46 & 25/3684/484 & 112/2160/1061 \\
    \textit{tower\_tucohy}   & 232   & 0.02/23.68/3.25 & 29/3385/738 & 137/2160/1322 \\
    \bottomrule
    \end{tabular}}}
    \caption{Summary of object characteristics by class in the TTPLA dataset, showing the number of objects, area percentages, and dimensions (width and height in pixels) with minimum, maximum, and average values.}
    \label{tab:ttpla-detailed}
\end{table}

The dataset is annotated in the \ac{COCO} format, facilitating detection and segmentation tasks, including semantic and instance segmentation. 
The diverse representation of objects against varying backgrounds, under different lighting conditions, and at multiple scales poses unique challenges to object detection and segmentation efforts. 
Furthermore, the TTPLA dataset supports the detection and semantic segmentation tasks and extends its utility to instance segmentation. This capability is crucial for identifying and differentiating between individual transmission towers and power lines, allowing for a deep analysis of the power grid infrastructure.

\begin{figure}[h!]
    \centering
    \subfloat[][]{\includegraphics[width=.24\linewidth]{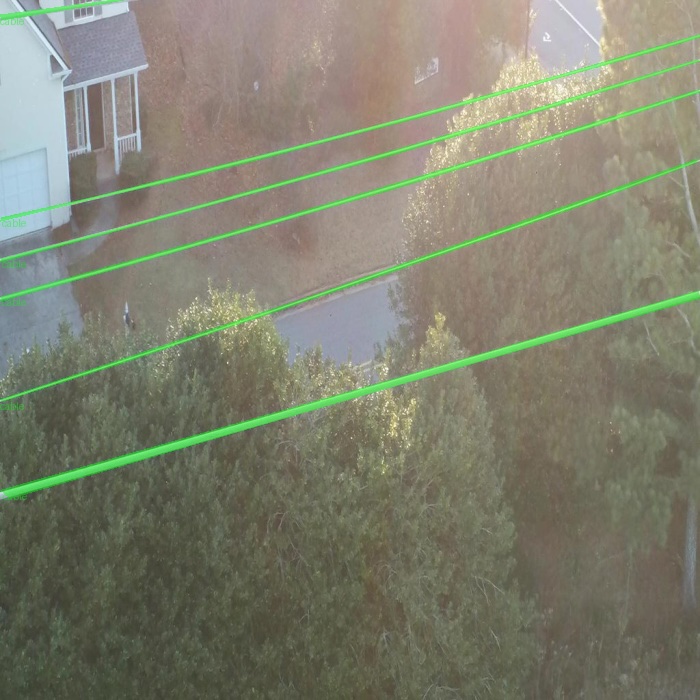}}\hfill
    \subfloat[][]{\includegraphics[width=.24\linewidth]{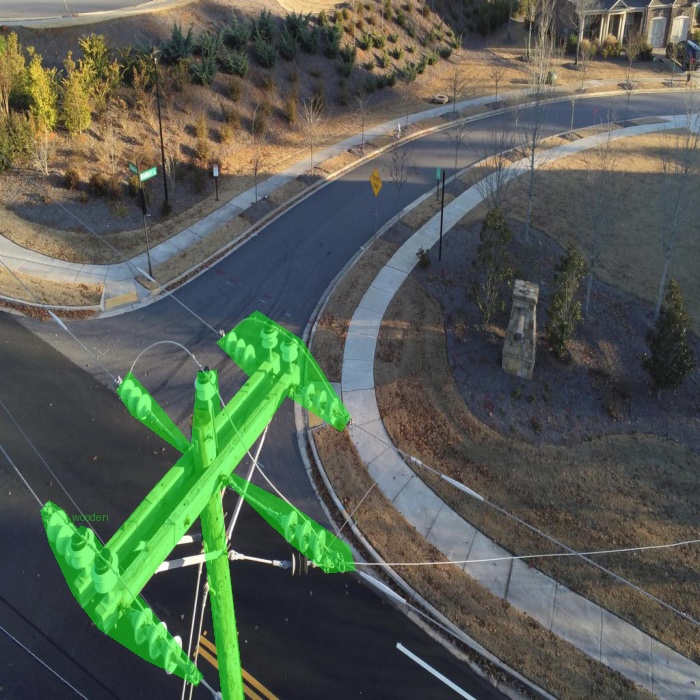}}\hfill
    \subfloat[][]{\includegraphics[width=.24\linewidth]{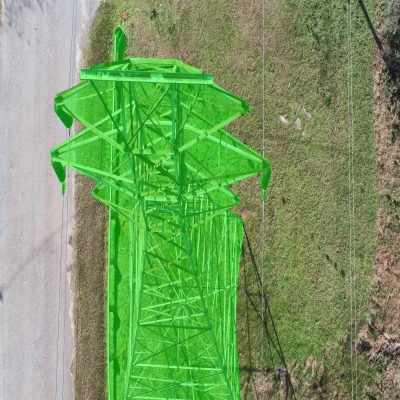}}\hfill
    \subfloat[]{\includegraphics[width=.24\linewidth]{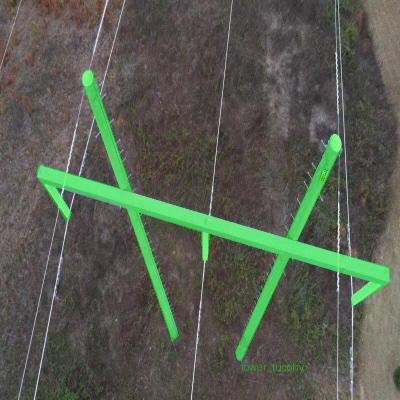}}
    \caption{Samples from the TTPLA dataset represent the main objects of categories in the green masks (a) \textit{cable}, (b) \textit{tower\_wooden}, (c) \textit{tower\_lattice}, (d) \textit{tower\_tucohy}.}
    \label{fig:datasamples}
\end{figure}

Figure~\ref{fig:datasamples} showcases a selection of aerial images from the TTPLA dataset, illustrating the diversity and complexity of the power-grid infrastructure components it encompasses, highlighting the challenges involved in the segmentation task due to the diverse backgrounds, lighting conditions, and object sizes.

\subsection{Base Model Performance}
\begin{figure}[htp]
    \centering
    \subfloat[Training loss]{%
        \includegraphics[width=0.49\linewidth]{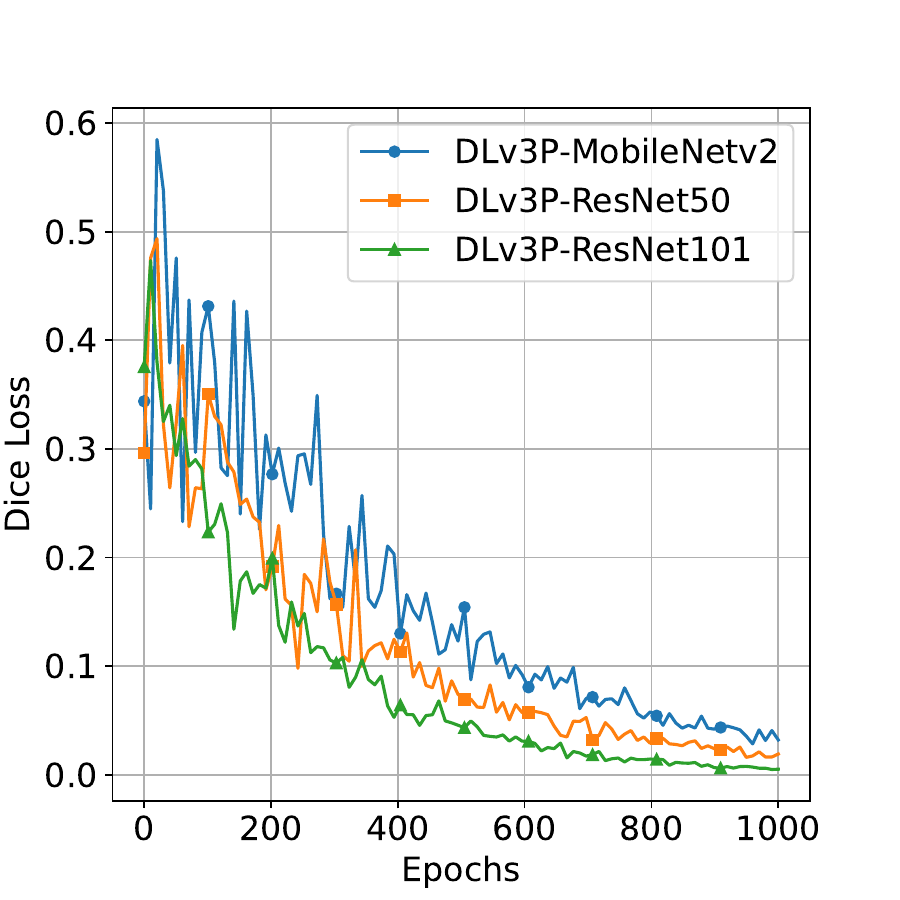}
        \label{fig:DLv3PLoss}
    }
    \subfloat[Training mIoU(\%)]{%
        \includegraphics[width=0.49\linewidth]{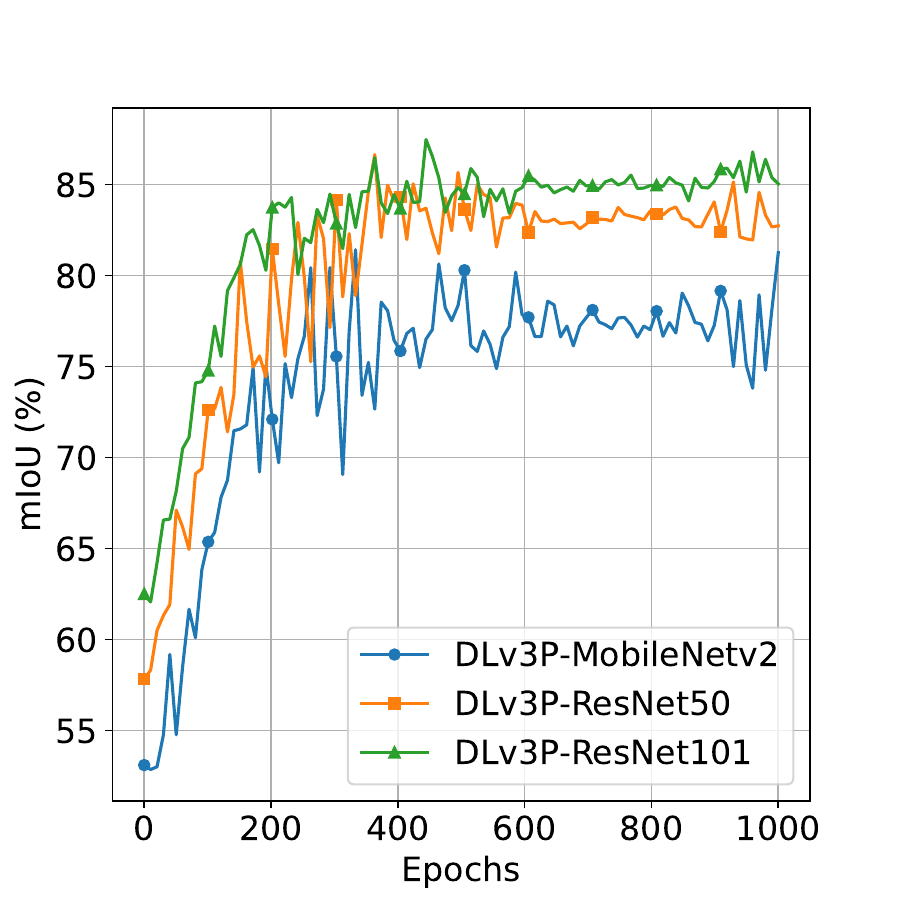}
        \label{fig:DLv3PmIoU}
    }
    \caption{The training loss in Dice loss and accuracy in mIoU(\%) of three base DeepLabv3Plus models over 1000 epochs.}
    \label{fig:training-loss-mIoU}
\end{figure}

In this section, we evaluate the performance of the base model $\Theta$ after the finetuning process. We leverage the DeepLabv3Plus model with three different backbones, namely MobileNetv2, ResNet50, and ResNet101. 
Each model is trained on the $\mathcal{D}_{\text{train}}$ for 1000 epochs, where the training loss and accuracy are presented in Figure~\ref{fig:training-loss-mIoU}.
The training loss, shown in Figure~\ref{fig:DLv3PLoss}, steadily decreases for all three models, indicating successful learning and convergence during the finetuning process. The DeepLabv3Plus with ResNet101 backbone achieves the lowest training loss, followed by ResNet50 and MobileNetv2.

The results reported in Table \ref{tab:result-detection-model} demonstrate the effectiveness of the DeepLabv3Plus architecture for the visual quality inspection on the validation set $\mathcal{D}_{\text{val}}$. 
The ResNet101 backbone, with its deeper network structure, captures more complex features and achieves the highest segmentation accuracy. 
The ResNet50 backbone balances performance and computational efficiency, while the MobileNetv2 backbone offers a lightweight option suitable for resource-constrained environments.

\begin{table*}[h]
\centering

\small
\renewcommand{\arraystretch}{1.3} 
\begin{tabular}{l|rr|rrrrr}
\hline
\textbf{Model} & \#Params & Size(MB) & \textit{cable} & \textit{tower\_wooden} & \textit{tower\_lattice} & \textit{tower\_tucohy} & mIoU(\%) \\
\hline
\hline
\textbf{DLv3P-MobileNetv2-B} & 4.37M & 16.71 & 53.94 & 80.11 & 88.19 & 86.49 & 77.18 \\
\textbf{DLv3P-MobileNetv2-E} & 4.37M & 16.71 & 54.37 & 86.49 & 80.98 & 88.78 & 77.82 \\
\textbf{DLv3P-MobileNetv2-M} & 3.51M & 13.39 & 48.74 & 84.54 & 78.55 & 86.12 & 75.48 \\
\hline
\textbf{DLv3P-ResNet50-B} & 26.67M & 101.76 & 56.66 & 92.31 & 93.18 & 90.63 & 83.20 \\
\textbf{DLv3P-ResNet50-E} & 26.67M & 101.76 & 57.42 & 92.97 & 91.34 & 93.87 & 83.90 \\
\textbf{DLv3P-ResNet50-M} & 21.36M & 81.48 & 53.67 & 90.39 & 88.80 & 91.26 & 81.53 \\
\hline
\textbf{DLv3P-ResNet101-B} & 45.66M & 174.21 & 57.22 & 95.23 & 96.02 & 91.42 & 84.97 \\
\textbf{DLv3P-ResNet101-E} & 45.66M & 174.21 & 58.36 & 96.54 & 92.89 & 97.61 & 86.35 \\
\textbf{DLv3P-ResNet101-M} & 36.57M & 139.52 & 54.43 & 94.07 & 90.22 & 95.09 & 83.95 \\
\hline
\end{tabular}
\caption{Accuracy comparison of DeepLabv3Plus (DLv3P) variants with different backbones (MobileNetv2, ResNet50, ResNet101) in terms of average IoU (\%) for each category and mIoU (\%). The models are evaluated at different stages: base (B), enhanced (E), and mobile (M) on the TTPLA validation set.}
\label{tab:result-detection-model}
\end{table*}

Overall, the finetuning process successfully adapts the base model $\Theta$ to the specific visual quality inspection task, achieving high IoU scores across different object categories, which can be a strong foundation for further optimization and deployment on edge devices.

\subsection{Explanation Evaluation}
To select the most advisable XAI method $\mathcal{X}$ for our framework, we perform both qualitative and quantitative evaluations of the explanations generated by several implemented XAI methods.

\subsubsection{Qualitative Evaluation}
\begin{figure*}[h!]
    \centering
    \includegraphics[width=\textwidth]{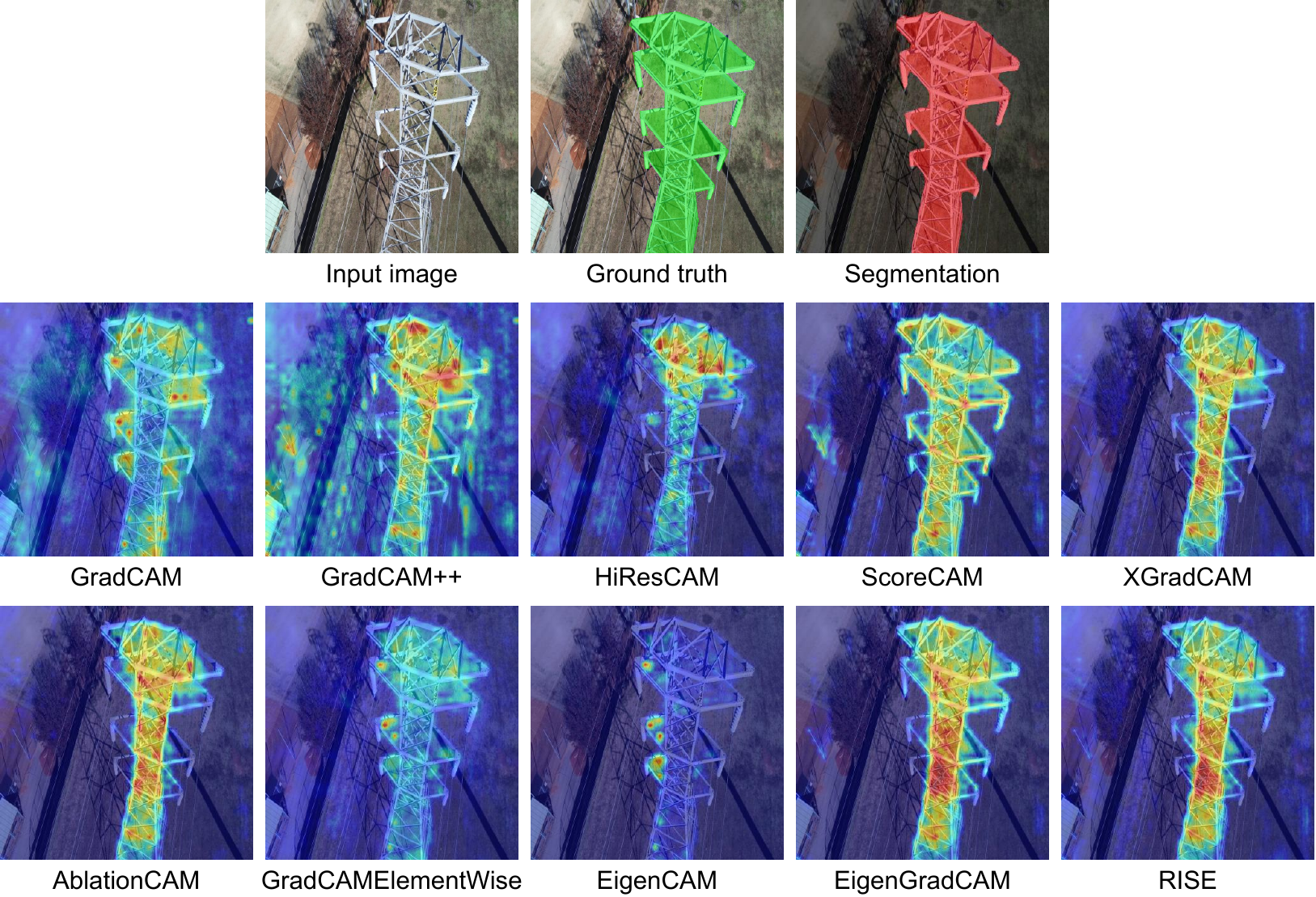}
    \caption{The qualitative evaluation of XAI methods in explaining the base DeepLabv3Plus-ResNet101 model on a validation sample. The category is the \textit{tower\_lattice}. The IoU value between the segmentation and the ground truth is 96.25\%.}
    \label{fig:xai-images}
\end{figure*}

The explanation maps of implemented XAI methods for the base model $\Theta$ on the TTPLA validation set $\mathcal{D}_\text{val}$ are demonstrated in Figure~\ref{fig:xai-images}. 
The figure presents a visual comparison of the explanation maps generated by various XAI methods, including RISE, EigenGradCAM, EigenCAM, GradCAM, AblationCAM, GradCAMElementWise, GradCAM++, HiResCAM, ScoreCAM, and XGradCAM.
The first row of the figure shows the input image, ground truth annotation, and the segmentation output produced by the model. 
The subsequent rows display the explanation maps generated by each XAI method. 

From a qualitative perspective, the explanation maps should provide valuable insights into the model's decision-making process. 
We observe variations in the highlighted regions by comparing the explanation maps across different XAI methods. 
Some methods, such as RISE and EigenGradCAM, produce more localized and fine-grained explanations, accurately capturing the relevant objects and their boundaries. Other methods, like GradCAM and GradCAM++, generate more coarse-grained explanations, highlighting larger regions of interest.
The qualitative evaluation also reveals the strengths and limitations of each XAI method. 
For instance, HiResCAM and ScoreCAM produce explanations with higher spatial resolution, enabling more precise localization of important features. On the other hand, methods like GradCAMElementWise and XGradCAM generate explanations with varying sensitivity levels to different image regions.

By visually comparing the explanation maps with the ground truth annotations and segmentation outputs, we can assess the plausibility of the explanations where plausible explanations should align well with human intuition and highlight regions that are semantically relevant to the target objects. 

\subsubsection{Quantitative Evaluation}
In this section, we present a quantitative evaluation of different XAI methods introduced in Module 2 (Section~\ref{ss:imp-base-model-xai}) on the validation set $\mathcal{D}_\text{val}$ using various plausibility and faithfulness metrics. 
The plausibility metrics include Energy-Based Pointing Game (EBPG), Intersection over Union (IoU), and Bounding Box (Bbox). 
In contrast, the faithfulness metrics include Deletion (Del) and Insertion (Ins), which are introduced in Module 3 (Section~\ref{ss:imp-xai-eval}). The evaluation results are summarized in Table~\ref{tab:quantitative-result}.

Among the evaluated methods, RISE stood out as the most advisable XAI method for our framework. RISE demonstrated exceptional performance in terms of faithfulness, achieving the best scores in both Deletion (0.123) and Insertion (0.691) metrics. These faithfulness results indicate that the explanations generated by RISE closely align with the model's predictive behavior, accurately capturing the most influential regions in the input images. It also demonstrated strong plausibility, with the second-best scores in EPBG (62.42\%) and IoU (56.13\%), and the best score in BBox (63.52\%). RISE's high scores in these plausibility metrics demonstrate that its explanations are highly interpretable and closely match human understanding of the important regions in the input images.

EigenGradCAM also shows strong performance in plausibility metrics, achieving the highest EBPG and IoU scores of 64.11\% and 60.93\%, respectively. It also obtains the second-highest Bbox score of 62.24\%. However, its faithfulness scores are not as impressive as RISE, suggesting that the salient regions identified by EigenGradCAM may not fully align with the model's predictive behavior. Other XAI methods, such as GradCAM++, XGradCAM, and AblationCAM, demonstrate competent performance in plausibility metrics but fall short in faithfulness compared to RISE.

Another advantage of RISE, particularly suitable for our visual quality framework, is its model-agnostic nature. As a model-agnostic XAI method, by treating the model as a ``black box'', RISE can be applied to any model without requiring access to its internal architecture or gradients. This property is crucial in our framework, where we may need to quantize or optimize the base model to a mobile model without being able to access or modify its architecture. RISE's flexibility ensures that we can still generate meaningful explanations for the model's predictions, regardless of any modifications made during the optimization process.

Based on our quantitative evaluation and considering the model-agnostic property of RISE, we use RISE as the advisable XAI method $\mathcal{X}$ for our visual quality framework. The selection of RISE as the advisable XAI method has significant implications for the subsequent modules in the framework, such as Module 4 (Section~\ref{ss:imp-xai-data-aug}) and Module 6 (Section~\ref{ss:imp-xai-edge}). By incorporating RISE into our framework, we can enhance the interpretability and performance of our models, enabling the domain expert to validate the model's predictions while allowing end-users to understand the model's rationale even with the mobile model $\theta$.

\begin{table}[h!]
\centering
\resizebox{\linewidth}{!}{%
{\rowcolors{2}{white}{gray!10}
\begin{tabular}{lrrrrr}
  \toprule
    Method & EPBG$\uparrow$ & BBox$\uparrow$ & IoU$\uparrow$ & Del$\downarrow$ & Ins$\uparrow$ \\
  \midrule
    GradCAM & 50.49 & 48.39 & 47.94 & 0.521 & 0.527  \\
    GradCAM++ & 58.13 & 52.24 & 53.22 & 0.517 & 0.547 \\
    HiResCAM & 60.81 & 41.69 & 52.19 & 0.501 & 0.559 \\
    ScoreCAM & 54.01 & 43.95 & 51.94 & \underline{0.434} & \underline{0.609} \\  
    XGradCAM & 57.94 & 47.81 & 53.09 & 0.594 & 0.551 \\
    AblationCAM & 61.03 & 51.39 & 54.73 & 0.498 & 0.589 \\
    GradCAMElementWise & 40.95 & 35.91 & 39.25 & 0.859 & 0.415 \\
    EigenCAM & 51.49 & 42.54 & 51.03 & 0.545 & 0.491 \\
    EigenGradCAM & \textbf{64.11} & \underline{62.24} & \textbf{60.93} & 0.520 & 0.534 \\
    RISE & \underline{62.42} & \textbf{63.52} & \underline{56.13} & \textbf{0.123} & \textbf{0.691} \\
  \bottomrule
\end{tabular}}
}
\caption{The quantitative result of XAI methods on the validation set $\mathcal{D}_{\text{val}}$ on different metrics such as EPBG(\%), BBox(\%), IoU(\%), Drop, Increase, and running time in seconds. For each metric, the arrow $\uparrow/\downarrow$ indicates higher/lower scores are better. The best is in \textbf{bold}, and the second best is in \underline{underline}.}\label{tab:quantitative-result}
\end{table}

\subsection{Model Improvement via XAI-guided Data Augmentation}
This section presents the experimental results of enhancing the DeepLabv3Plus-ResNet101 model's performance using annotation augmentation guided by the advisable XAI method $\mathcal{X}$ and the domain expert. 
The process begins with the XAI method $\mathcal{X}$ generating explanations for each image in the validation set $\mathcal{D}_\text{val}$. 
The domain expert, knowledgeable in semantic segmentation models and XAI algorithms, analyzes the saliency maps to guide data augmentation. 
 
\begin{figure}[h]
    \centering
    \includegraphics[width=\linewidth]{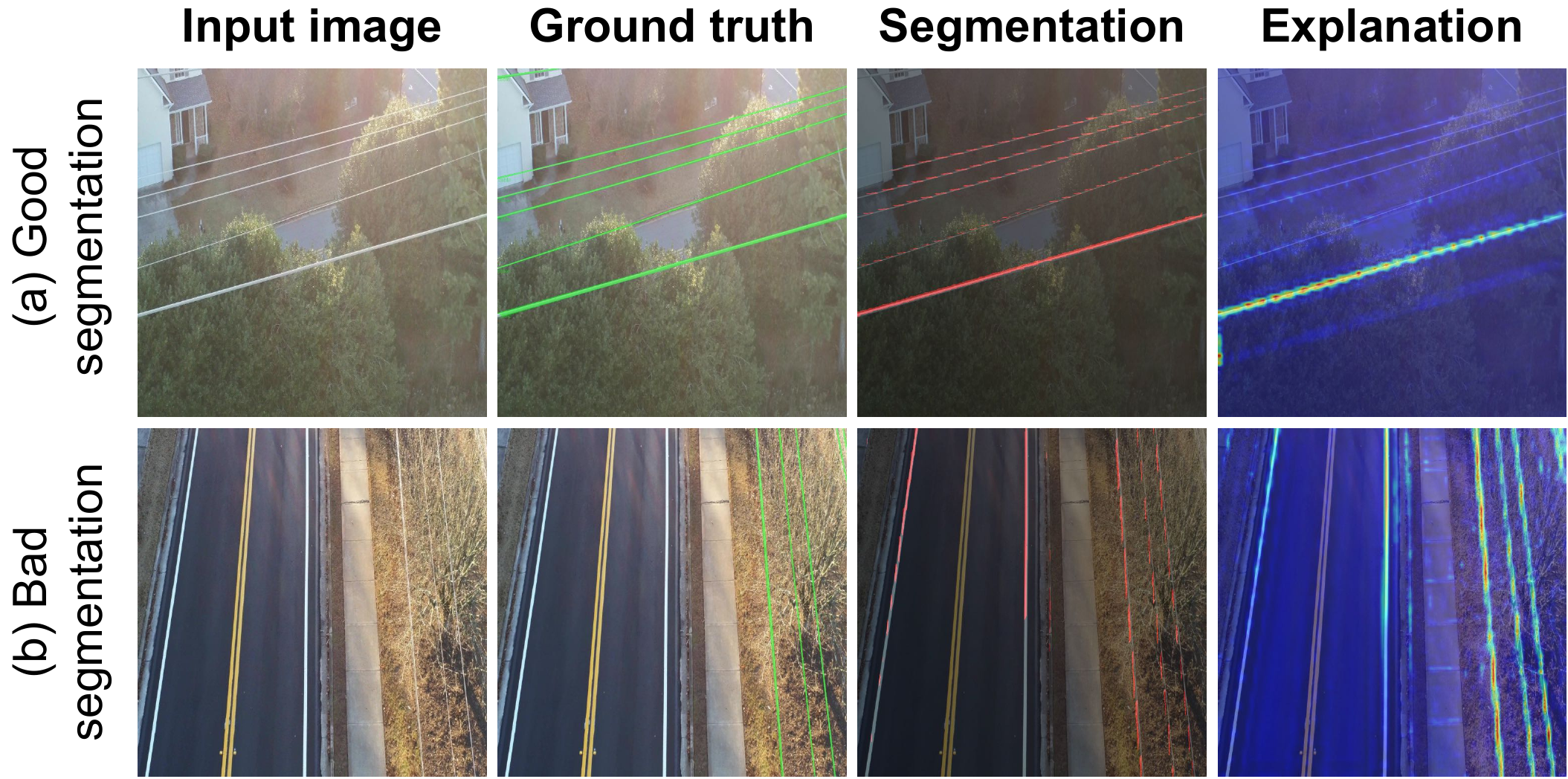}
    \caption{List of input images, ground truth, segmentation of the base DeepLabv3Plus-ResNet101-B model for the \textit{cable} inspection for two segmentation cases: (a) Good segmentation and (b) Bad segmentation.}
    \label{fig:hirescamxaiimp}
\end{figure}

As shown in Figure~\ref{fig:hirescamxaiimp}, the model effectively segments the cable from a clean or mixed-objects background, such as Figure \ref{fig:hirescamxaiimp}a. However, the model's performance decreases when the background contains objects resembling the target object, as shown in Figure \ref{fig:hirescamxaiimp}b. 
The explanations reveal that the model's attention is directed at the object and the surrounding background. 
However, the model lacks contextual attention to surrounding objects and backgrounds in complex cases. 
This behavior is due to the ability of models to leverage local and global contextual information from the original annotations~\cite{petsiuk2021black}.
A domain expert suggests annotation augmentation for each sample to enhance the model's performance.

\begin{figure}[h]
    \centering
        \subfloat[Annotation Enlargement]{%
        \includegraphics[width=0.75\linewidth]{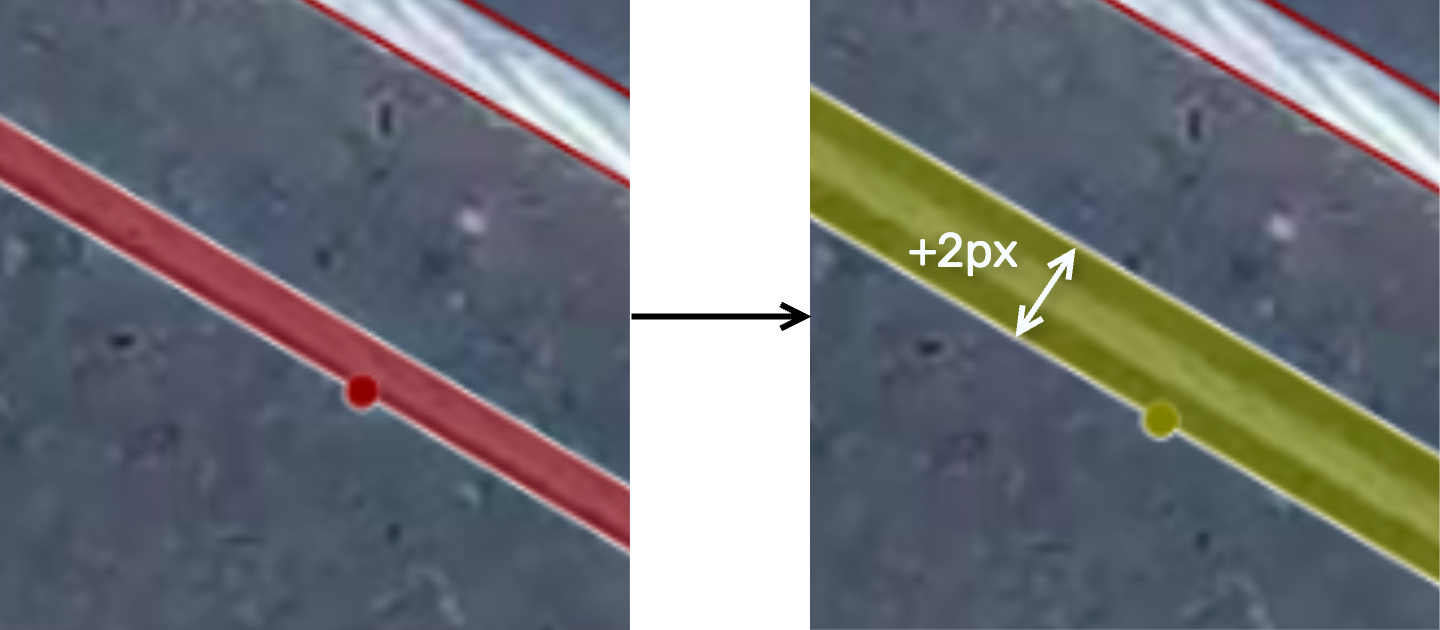}
        \label{fig:ann-enlargement}
    }
    \hfill
    \subfloat[Adding Annotations for Perplexed Objects]{%
        \includegraphics[width=0.75\linewidth]{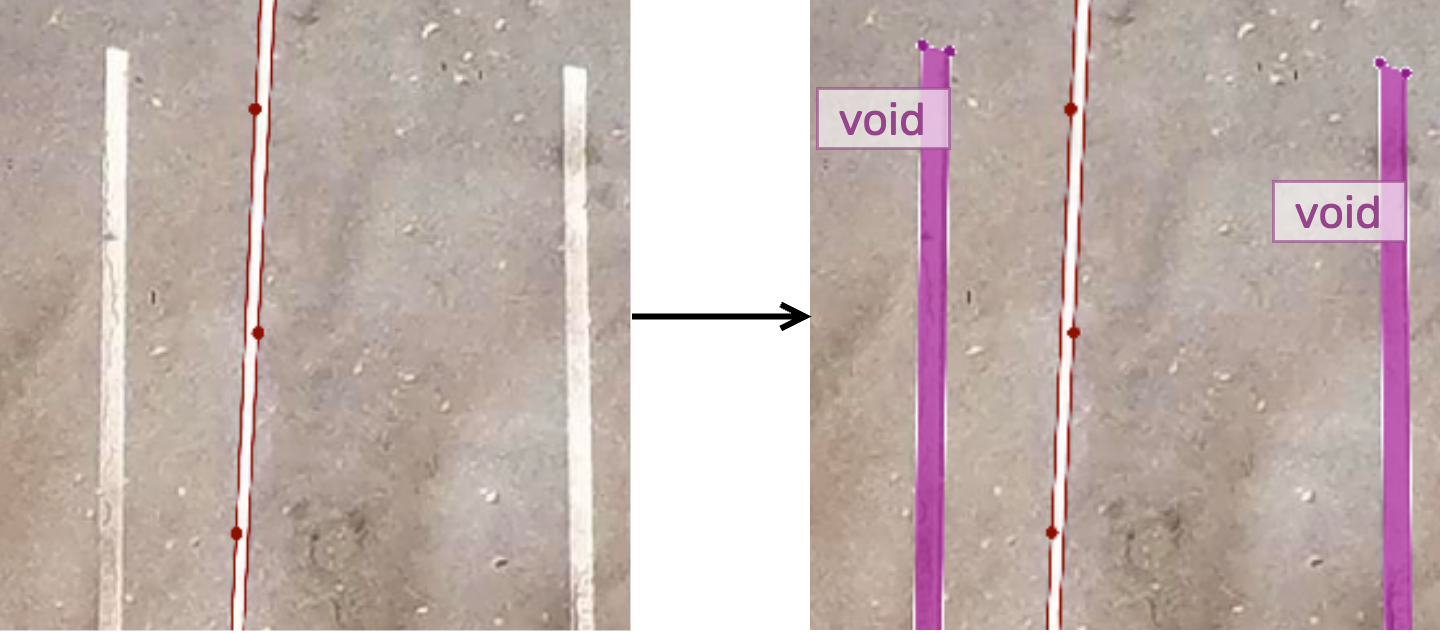}
        \label{fig:add-ann}
    }
    \caption{Annotation augmentation approaches: (a) Annotation enlargement where the annotation size for thin objects like cables is increased by 2 pixels on both sides. (b) Adding annotations for perplexed objects like the road surface marks to guide the model in differentiating between white cables and perplexed objects.}
    \label{fig:fix-anno}
\end{figure}

Two approaches are proposed, namely \textit{Annotation Enlargement} and \textit{Adding Annotations for Perplexed Objects}, as illustrated in Figure~\ref{fig:fix-anno}, which are described as follows:
\begin{itemize}
    \item \textbf{Annotation Enlargement} (Figure~\ref{fig:ann-enlargement}): Given that the model can leverage surrounding contextual information to improve its performance, we propose to enlarge the annotations of thin objects, especially thin cables, which the model often overlooks based on the saliency maps. We increase the object's size by 2 pixels on both sides.
    \item \textbf{Adding Annotations for Perplexed Objects} (Figure~\ref{fig:add-ann}): As the model often confuses cables with perplexing objects like road surface markings, we propose adding \textit{void} annotations to categorize these perplexing objects as unlabeled objects.
\end{itemize}

\begin{figure*}[h!]
    \centering
    \includegraphics[width=\textwidth]{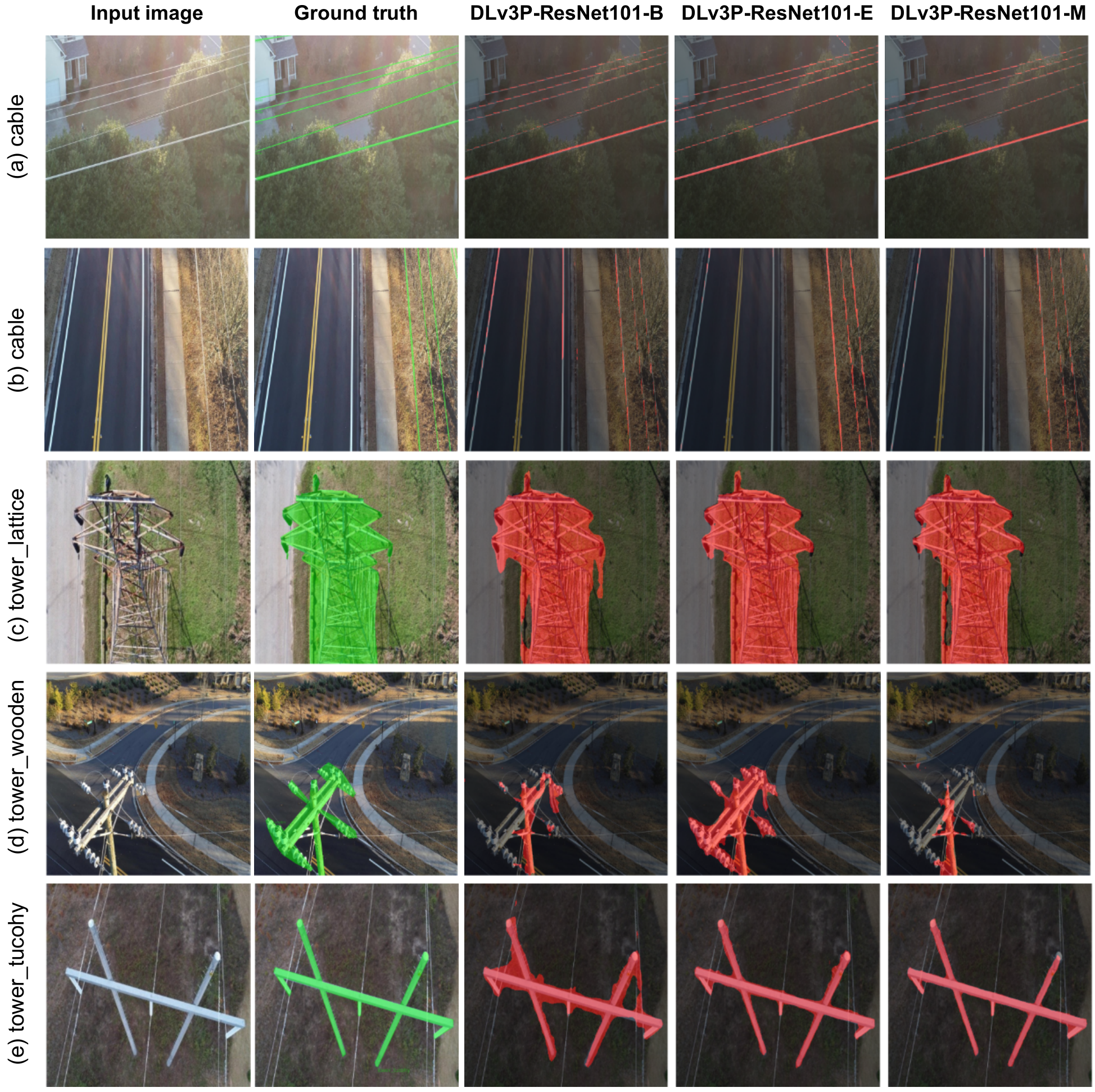}
    \caption{Qualitative results of DeepLabv3Plus (DLv3P)-ResNet101 on four categories of the TTPLA dataset. The models are evaluated at different stages: base (B), enhanced (E), and mobile (M).}
    \label{fig:segmentation}
\end{figure*}

The enhanced DeepLabv3Plus-ResNet101 (DLv3P-ResNet101-E) model demonstrates improved segmentation of thin objects from the background and perplexing objects (Figure~\ref{fig:segmentation}). 
The IoU of the enhanced model $\tilde{\Theta}$ is also higher than that of the base version $\Theta$, particularly with the $\mathtt{cable}$ category leading to a higher overall mIoU, as shown in Table~\ref{tab:result-detection-model}.

\subsection{Mobile Model Performance}
In this subsection, we evaluate the performance of the DeepLabv3Plus models after the quantization and deployment on mobile devices. 
The quantitative results are presented in Table~\ref{tab:result-detection-model} comparing the models' performance, while Figure~\ref{fig:segmentation} illustrating the qualitative results showcase segmentation outputs for the visual quality inspection task.

From Table~\ref{tab:result-detection-model}, we observe that, when deployed on mobile devices, the mobile model variants experience a slight performance drop due to the quantization and optimization process, which reduces the model's size and computational complexity to make it suitable for mobile inference.
The mobile DeepLabv3Plus-MobileNetv2 (DLv3P-MobileNetv2-M) model, specifically designed for mobile deployment, maintains a competitive mIoU of 75.48\% while having significantly fewer parameters (3.51M) and a smaller model size (13.39 MB) compared to the ResNet-based models.

Qualitatively, Figure~\ref{fig:segmentation} visually compares the segmentation results produced by the base, enhanced, and mobile versions of the DeepLabv3Plus model with the ResNet101 backbone. The first column shows the input images, while the second column demonstrates the corresponding ground truth annotations. The subsequent columns demonstrate the segmentation outputs of the base, enhanced, and mobile models, respectively.
The qualitative results show that the enhanced DeepLabv3Plus-ResNet101 (DLv3P-ResNet101-E) model produces highly accurate segmentation masks, closely resembling the ground truth annotations.
Especially, the enhanced model is less prone to perplexed objects, as shown in Figure~\ref{fig:segmentation}b. 
The segmentation quality of the mobile DeepLabv3Plus-ResNet101 (DLv3P-ResNet101-M) model is slightly lower compared to the enhanced version, with some minor discrepancies in the segmented regions. However, the mobile model still maintains good visual quality, successfully identifying and segmenting key objects of interest.

The results confirm the visual quality of the segmentation outputs produced by the mobile models, highlighting their effectiveness in real-world scenarios. 
Overall, our procedure of developing mobile models can provide a balance between performance and computational efficiency.

\subsection{Textual Explanation}
\newtcolorbox[auto counter]{definition}[1][]{
  enhanced,
  fonttitle=\scshape,
  #1
}

\begin{figure*}[h!]
\begin{definition}[label=template:A,
  title={Template \thetcbcounter: Prompt template for the GPT-4 Vision}
]
\textbf{System Message:} 
You are an Explainable AI expert for semantic segmentation models. \\
The first image is the original image. 
The second image is the ground truth of category \{\textcolor{ForestGreen}{category}\} in the original image. 
The third image is the prediction of an AI model for category \{\textcolor{ForestGreen}{category}\} in the original image.
The fourth image is the explanation map of the category \{\textcolor{ForestGreen}{category}\} in the original image.
Think step by step to understand how the explanation map and prediction align with the ground truth.
First, capture the image context in the original image.
Secondly, identify which parts belong to category \{\textcolor{ForestGreen}{category}\} in the ground truth.
Then, you look at the explanation map to see the saliency map for the segmentation mask of the category \{\textcolor{ForestGreen}{category}\}.
Your task is to check if the focused region in the explanation map supports the prediction for category \{\textcolor{ForestGreen}{category}\}.
Your final answer must be concise, simple, and separated by bullet points.
First, briefly describe the explanation map.
Secondly, describe the most focused region of category \{\textcolor{ForestGreen}{category}\} in the explanation.
Thirdly, describe the least focused region of category \{\textcolor{ForestGreen}{category}\} in the explanation.
Fourthly, assess the localization quality if the focused region aligns with the prediction and ground truth for category \{\textcolor{ForestGreen}{category}\}.
\\
\\
\textbf{User Message:} \\
image\_url: ``data:image/jpeg;base64,\{\textcolor{ForestGreen}{input\_image}\}"\\
image\_url: ``data:image/jpeg;base64,\{\textcolor{ForestGreen}{ground\_truth\_image}\}"\\
image\_url: ``data:image/jpeg;base64,\{\textcolor{ForestGreen}{segmentation\_image}\}"\\
image\_url: ``data:image/jpeg;base64,\{\textcolor{ForestGreen}{explanation\_image}\}"
\end{definition}
\end{figure*}

\begin{figure*}[h!]
\begin{definition}[label=template:B,
  title={Template \thetcbcounter: Textual Explanations of the TTPLA dataset}
]
\begin{minipage}{\linewidth}
    \centering
    \includegraphics[width=.7\linewidth]{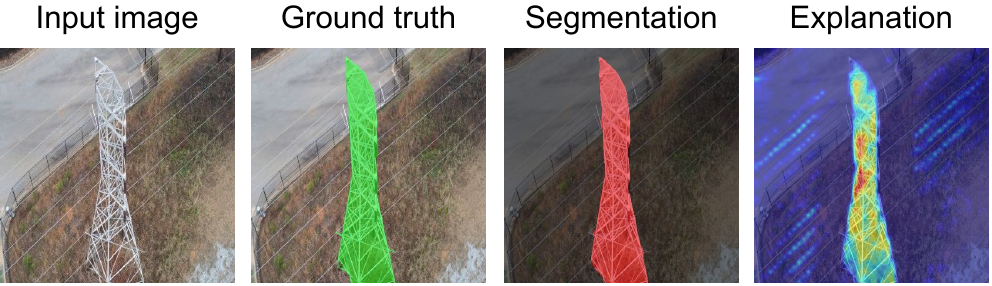}
    \caption{Good segmentation - Category: \textit{tower_lattice}}
    \label{fig:text-exp-tower-wooden}
\end{minipage}
\hfill
\begin{minipage}{\linewidth}
  
    \textbf{Model Output:} \\
    - The explanation map is a visual representation showing the importance of different regions of the image for predicting the category tower\_lattice. 
    
    - The most focused region in the explanation map seems to be the central part of the tower, with yellow to red gradients, indicating a high degree of importance according to the model.
    
    - The least focused region in the explanation map appears to be the bottom and peripheral parts of the tower, shown by the cooler colors such as blue, which suggest these areas have less influence on the model's decision.
    
    - The segmentation accuracy seems high, where the focused region corresponds well with the AI model's prediction and the ground truth. This alignment indicates that the model focuses on relevant parts of the image to predict the category tower\_lattice.
\end{minipage}
\\
\vspace{2pt}
\\
\textcolor[RGB]{0,0,0}{\rule{\linewidth}{0.2pt}}
\\
\vspace{2pt}
\\
\begin{minipage}{\linewidth}
    \centering
    \includegraphics[width=.7\linewidth]{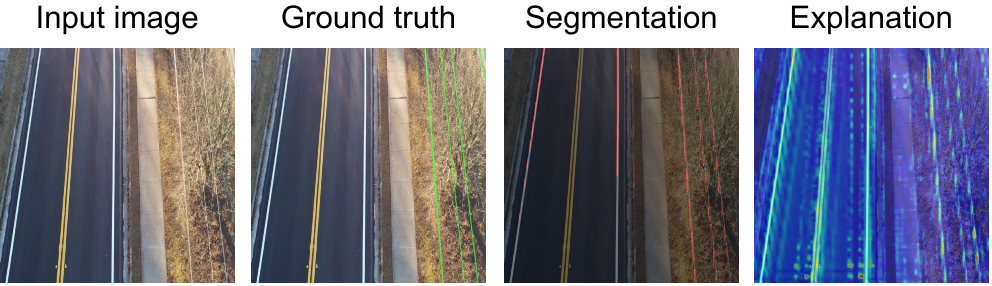}
    \caption{Bad segmentation - Category: \textit{cable}}
    \label{fig:text-exp-cable}
\end{minipage}
\hfill
\begin{minipage}{\linewidth}
  
    \textbf{Model Output:} \\
    - The explanation map is a visual representation that indicates which areas of the image the AI model deemed relevant when determining where the category ``cable" is present. 
    
    - The most focused region of the cable in the explanation map is evident along the linear features within the road, which correlate with the yellow lane markings and the road's edges.
    
    - The least focused region in the explanation appears to be the areas on the pavement free from markings and the natural vegetation areas on the right side of the image.
    
    - The focused region in the explanation map does not perfectly align with the ground truth for the cable category. The prediction appears to have false positives along the lane markings. It does not accurately distinguish between the road markings and actual cables, whereas the ground truth demarcates straight lines purported to be cables adjacent to the road on both sides. Therefore, there is a discrepancy between the focused regions in the explanation map and the true location of cables, as shown in the ground truth image.
\end{minipage}
\end{definition}
\end{figure*}
In this subsection, we present the results of generating textual explanations to provide a human-readable interpretation of the segmentation results on the TTPLA dataset. The textual explanations aim to enhance the interpretability and trustworthiness of the mobile app for field engineers by offering a concise and intuitive summary of the model's rationale. 

To generate the textual explanations, we design a prompt template that includes the original image, the ground truth image, the segmentation image, and the saliency map. 
The LVLM processes this information and leverages its pre-trained knowledge to generate human-readable explanations.
Template \ref{template:A} presents the prompt template used for generating these explanations.
The system message informs LVLM that it should act as an XAI expert in describing saliency maps retrieved from XAI methods for a semantic segmentation model. 
It outlines the input structure, including the original image, the ground truth image, the segmentation image, and the explanation map image. The system message also instructs the model to think step-by-step, identifying parts of each image's specific category. 
Furthermore, it guides the model in describing the concentrated regions of the explanation map. It provides instructions on how to format the final answer, emphasizing correctness and simplicity for end-user understanding.
The user message, on the other hand, provides the specific inputs for the model to process. 
In our framework, the user message includes the URLs of the original image, ground truth image, segmentation image, and explanation image.
These images are passed to the model as base64-encoded strings.

Template \ref{template:B} shows examples of the textual explanations generated by the LVLM for the good segmentation of \textit{tower_wooden} (Figure \ref{fig:text-exp-tower-wooden}) and bad segmentation of \textit{cable} (Figure \ref{fig:text-exp-cable}) categories, respectively.
The generated textual explanation provides a clear overview of the saliency map, indicating areas of interest where the prediction model concentrates on identifying the presence of a wooden tower structure or cables. The explanation highlights the most concentrated region and points out the least concentrated region.
Moreover, the textual explanation can also assess the model's segmentation performance, whether the model has a good or bad segmentation, by combining the information from the saliency image, prediction image, and ground truth image.
The textual explanation shown directly on the mobile devices and the saliency map aims to support end-users in understanding the model's rationale in a human-centered manner.

\section{Experiment 2: A Data-Centric Approach}\label{s:datacentric}
Based on the findings from Experiment 1, we further evaluate our framework in Experiment 2 using a data-centric approach. The experimental procedures remain consistent with those used in the previous experiment. However, this time we directly apply $\mathcal{X}$ as the explaining technique without conducting comparative analyses of different XAI methods. Additionally, we utilize a different public industrial dataset, focusing on the inspection of substation equipment.

\subsection{Dataset}
The Substation Equipment dataset comprises 1,660 images of electric substations collected using handheld cameras, Automated guided vehicle (AGV)-mounted cameras, and fixed-location cameras \cite{gomes7884270}. It features 15 categories of substation equipment, with a total of 50,705 annotated objects. These categories include \textit{open_blade_disconnect_switch}, \textit{breaker}, \textit{closed_blade_disconnect_switch}, \textit{open_tandem_disconnect_switch}, \textit{fuse_disconnect_switch}, \textit{porcelain_pin_insulator}, \textit{closed_tandem_disconnect_switch}, \textit{muffle}, \textit{potential_transformer}, \textit{lightning_arrester}, \textit{recloser}, \textit{power_transformer}, \textit{current_transformer}, \textit{glass_disc_insulator}, and \textit{tripolar_disconnect_switch}. Table \ref{tab:substation-detailed} summarizes the object characteristics by class, detailing counts, area percentages, and dimension ranges (width and height) for analysis.

\begin{table}[ht]
    \centering
    \resizebox{\linewidth}{!}{%
    {\rowcolors{2}{white}{gray!10}\begin{tabular}{>{\raggedright\arraybackslash}p{2.6cm}r>{\raggedleft\arraybackslash}p{2.2cm}>{\raggedleft\arraybackslash}p{2cm}>{\raggedleft\arraybackslash}p{2cm}}
    \toprule
    \multirow{2}{*}{\textit{Class}} & \multirow{2}{*}{\#Obj.} & Min/Max/Avg Area (\%) & Min/Max/Avg Width (px) & Min/Max/Avg Height (px) \\
    \midrule
    \textit{porcelain\_pin\_insulator}      & \multirow{2}{*}{26450} & \multirow{2}{*}{0.00/16.16/0.08} & \multirow{2}{*}{2/2244/57} & \multirow{2}{*}{4/2291/64} \\
    \textit{closed\_blade\_disconnect\_switch} & \multirow{2}{*}{5225}  & \multirow{2}{*}{0.00/16.22/0.35} & \multirow{2}{*}{3/1107/97} & \multirow{2}{*}{9/2003/216} \\
    \textit{glass\_disc\_insulator}         & \multirow{2}{*}{3180}  & \multirow{2}{*}{0.00/0.78/0.05}  & \multirow{2}{*}{3/283/66}  & \multirow{2}{*}{6/399/62} \\
    \textit{tripolar\_disconnect\_switch}   & \multirow{2}{*}{2348}  & \multirow{2}{*}{0.00/4.9/0.23}   & \multirow{2}{*}{3/839/98}  & \multirow{2}{*}{5/754/124} \\
    \textit{recloser}                       & 2330  & 0.00/25.44/1.52 & 4/1486/251 & 3/1355/230 \\
    \textit{current\_transformer}           & \multirow{2}{*}{2128}  & \multirow{2}{*}{0.00/2.68/0.25}  & \multirow{2}{*}{4/1232/101} & \multirow{2}{*}{6/761/125} \\
    \textit{lightning\_arrester}            & 1974  & 0.00/1.78/0.06  & 4/397/40   & 4/963/84 \\
    \textit{open\_tandem\_disconnect\_switch} & \multirow{2}{*}{1596} & \multirow{2}{*}{0.00/5.12/0.09}  & \multirow{2}{*}{4/659/100}  & \multirow{2}{*}{4/1114/108} \\
    \textit{muffle}                         & 1354  & 0.00/9.26/0.09  & 5/280/41   & 4/956/165 \\
    \textit{breaker}                        & 980   & 0.00/15.42/1.08 & 7/1487/156 & 6/1468/298 \\
    \bottomrule
    \end{tabular}}}
    \caption{Summary of object characteristics by class in the Substation dataset, showing the number of objects, area percentages, and dimensions (width and height in pixels) with minimum, maximum, and average values.}
    \label{tab:substation-detailed}
\end{table}

\begin{figure}[h!]
    \centering
    \subfloat[][]{\includegraphics[width=.24\linewidth]{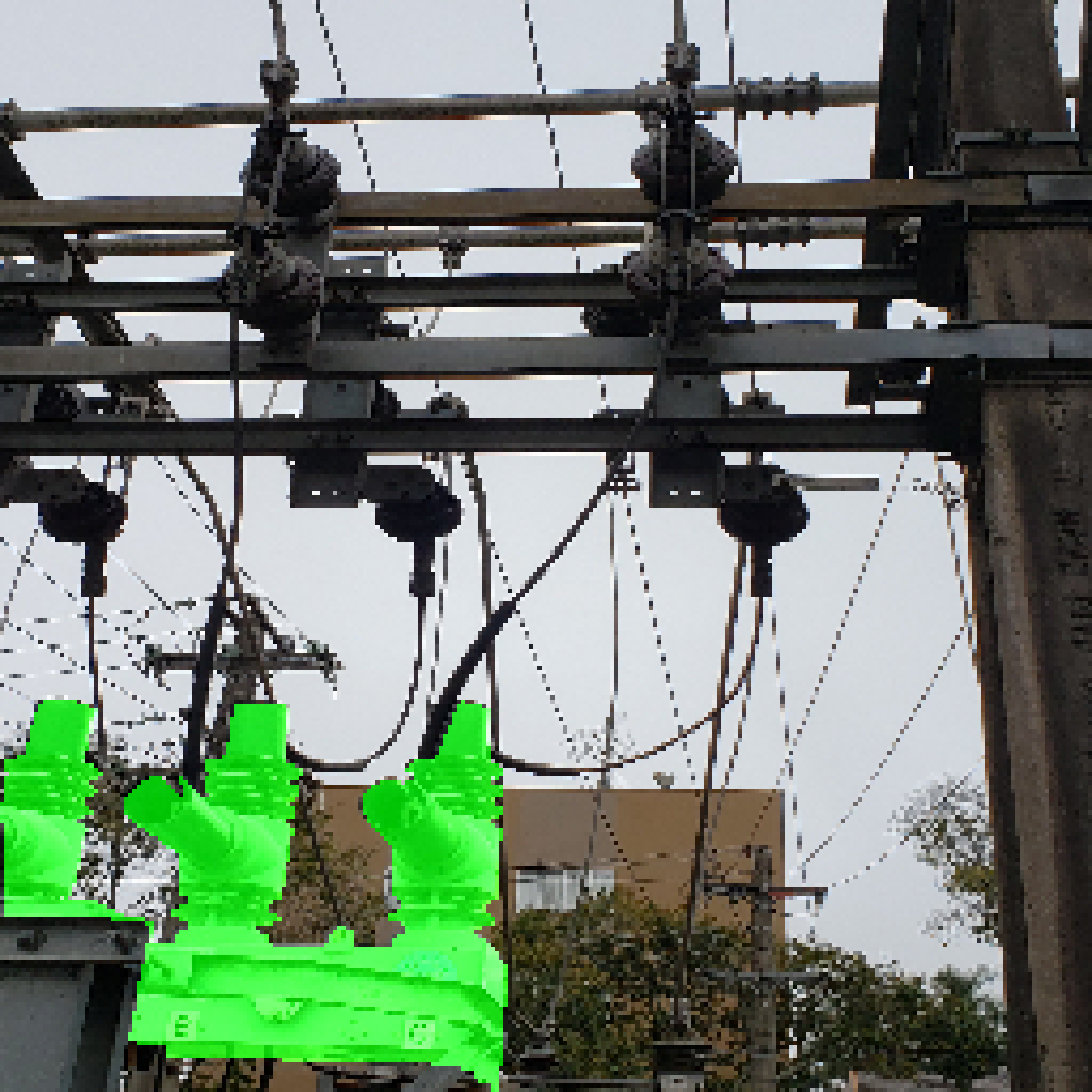}}\hfill
    \subfloat[][]{\includegraphics[width=.24\linewidth]{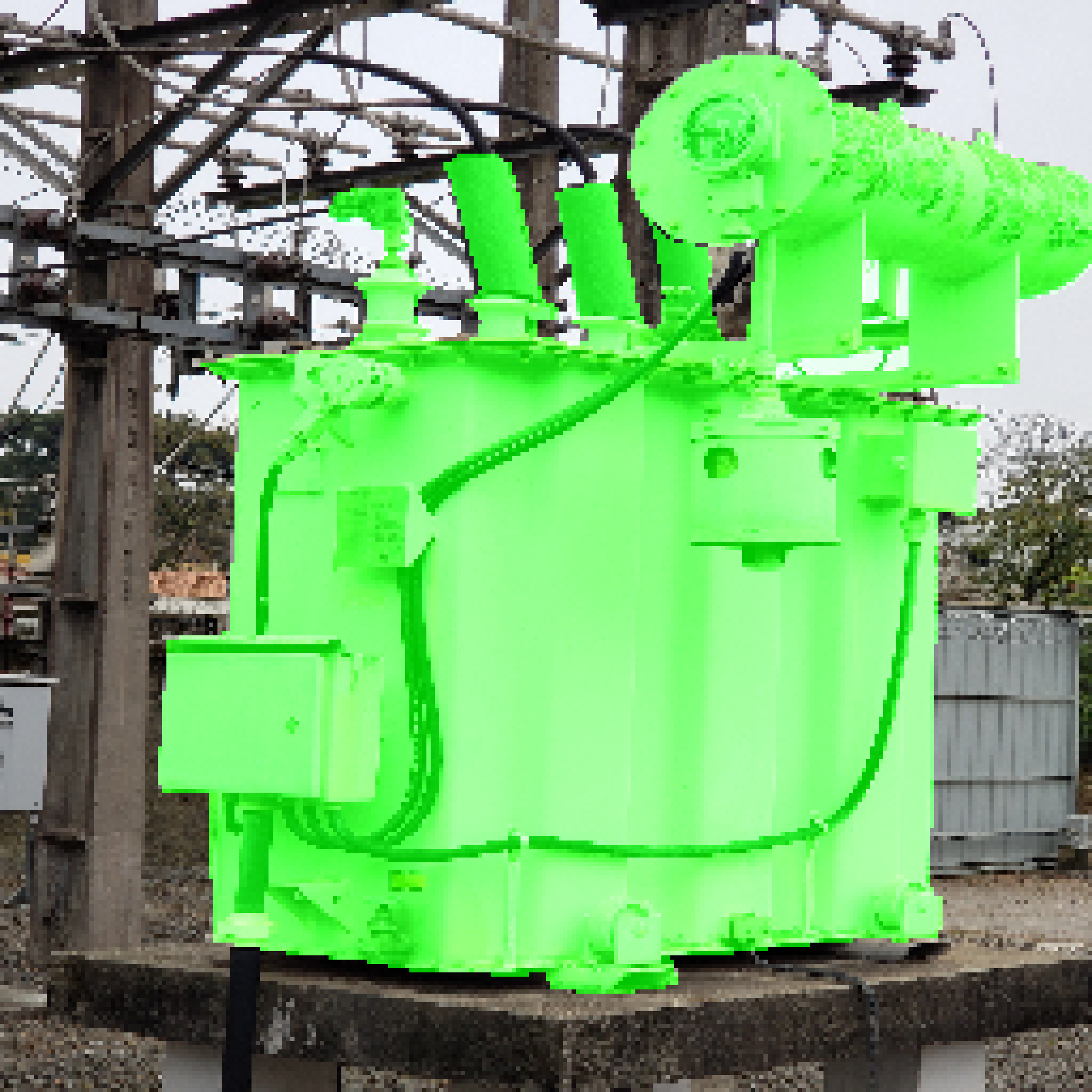}}\hfill
    \subfloat[][]{\includegraphics[width=.24\linewidth]{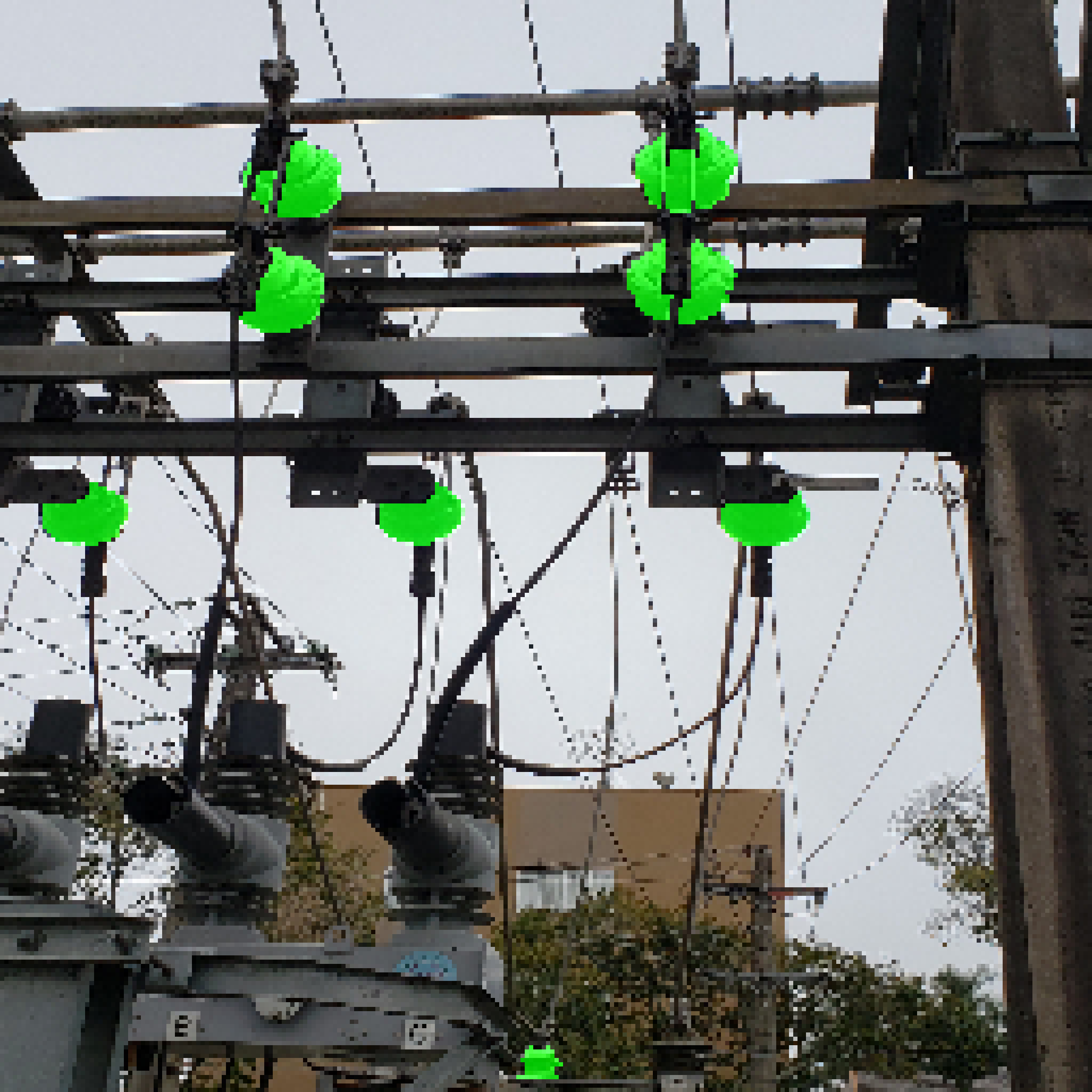}}\hfill
    \subfloat[]{\includegraphics[width=.24\linewidth]{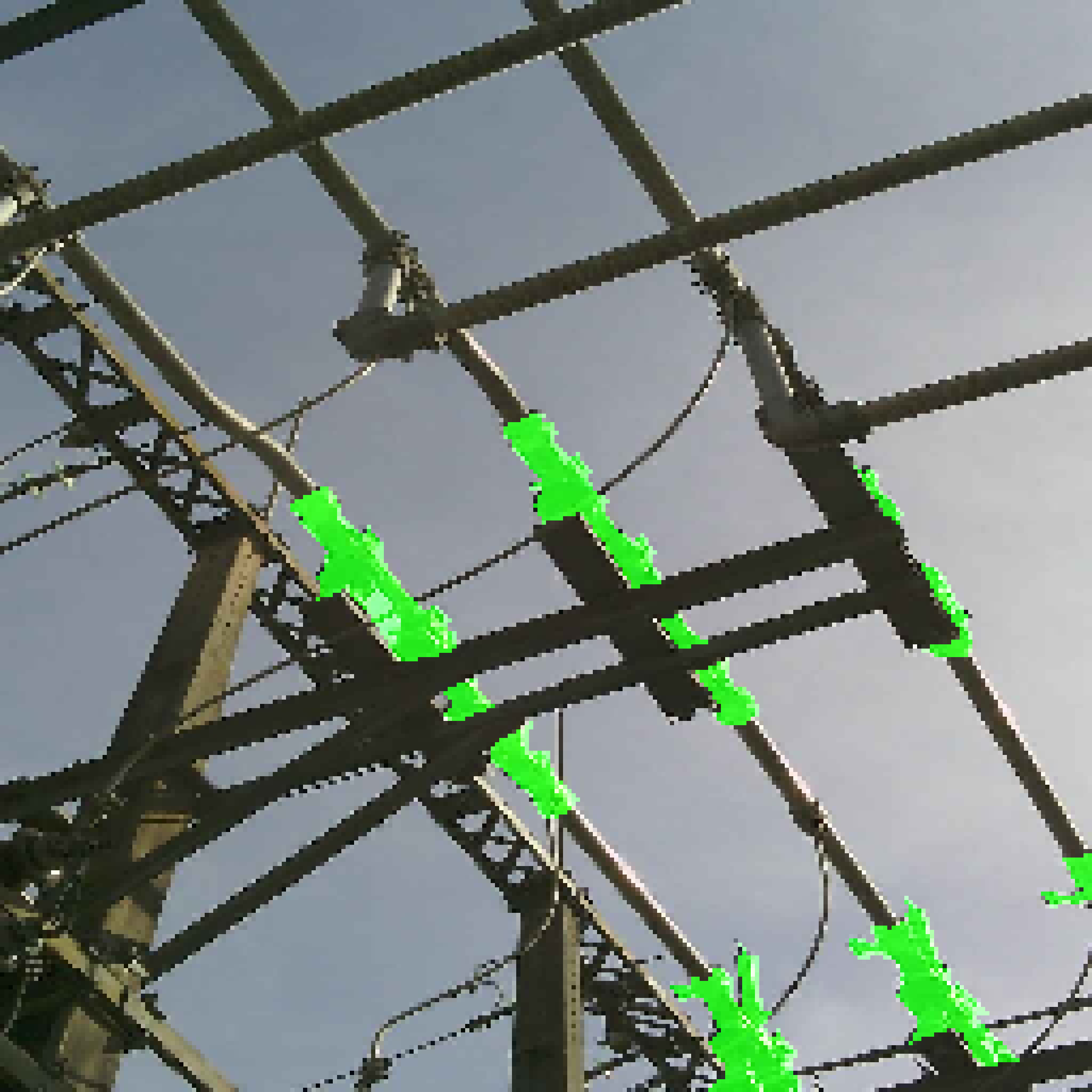}}
    \caption{Samples from the Substation Equipment dataset represent the main objects of categories in the green masks (a) \textit{recloser}, (b) \textit{power_transformer}, (c) \textit{porcelain_pin_insulator}, (d) \textit{tripolar_disconnect_switch}.}
    \label{fig:SE-datasamples}
\end{figure}

The dataset provides annotations in two formats: VOC-style polygonal JSON files and segmentation masks in Portable Network Graphic images. 
In our experiment, we use the VOC-style polygonal JSON files for semantic annotations. Figure \ref{fig:SE-datasamples} presents samples from several categories, such as recloser, power transformer, porcelain pin insulator, and tripolar disconnect switch.

The primary application of this dataset is to develop automated inspection systems for substations. Training semantic segmentation models on this dataset allows researchers to improve the accuracy and reliability of these systems, resulting in more efficient and effective monitoring of substation equipment.

\subsection{Base Model Performance}
In this experiment, we demonstrate the efficiency of our framework when applied to the DeepLabv3Plus model with a ResNet101 backbone. The model is trained on the training set $\mathcal{D}_{\text{train}}$ for 1000 epochs, and the accuracy for each category is presented in Table~\ref{tab:substation-results}.

The base DeepLabv3Plus-ResNet101 model achieves promising results on the validation set, with an overall mIoU of 73.45\%. 
However, the model's performance varies across different categories. Some categories, such as \textit{breaker}, \textit{open_blade_disconnect_switch}, and \textit{fuse_disconnect_switch} achieve high IoU scores above 90\%. 
In contrast, others, like \textit{porcelain_pin_insulator} and \textit{closed_blade_disconnect_switch}, have lower scores of around 30\% and 49\%, respectively. These performance variations underscored the challenges posed by the diverse object categories in the employed dataset.

\begin{figure}[h!]
    \centering
    \includegraphics[width=\linewidth]{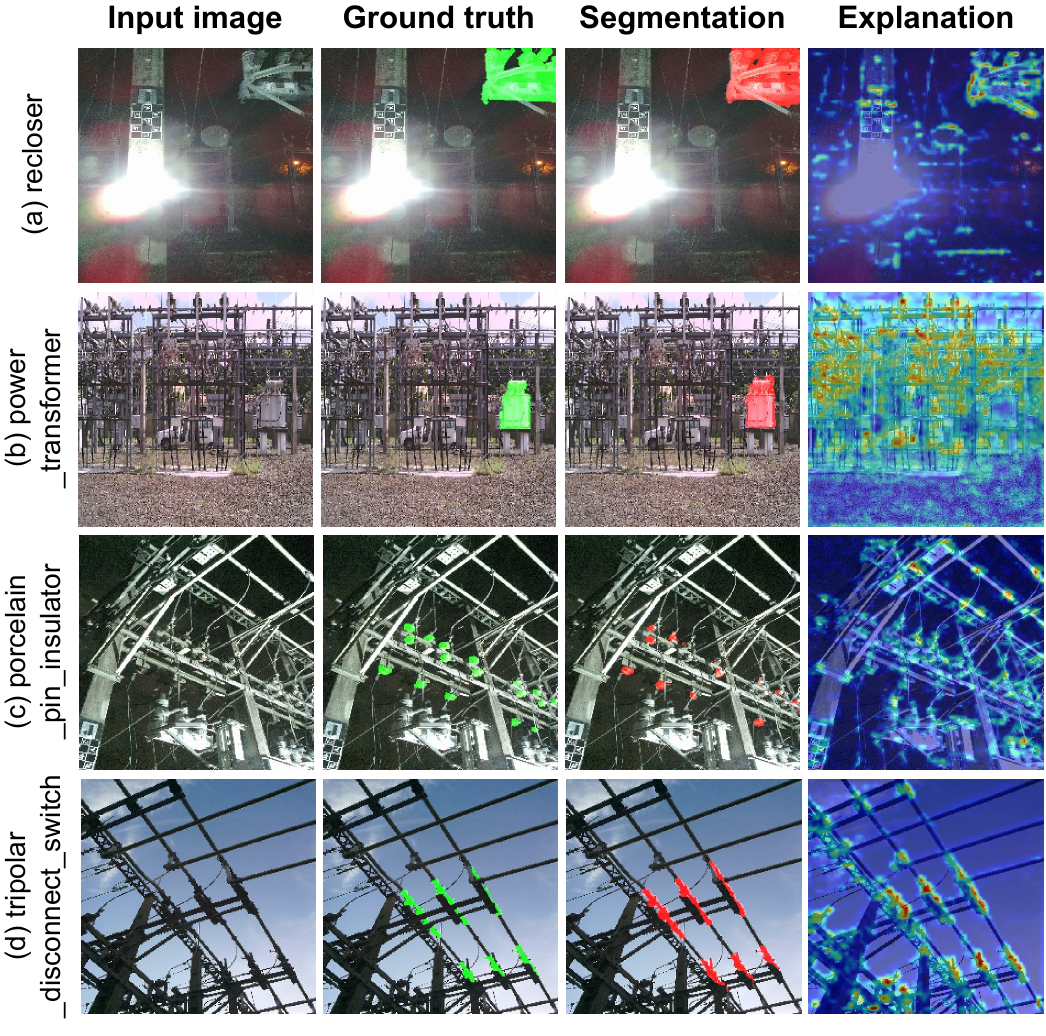}
    \caption{List of input images, ground truth, segmentation of the base DeepLabv3Plus-ResNet101-B model, and the corresponding explanations for four categories of the Substation equipment dataset.}
    \label{fig:substation-img-results}
\end{figure}

\begin{table}[h!]
    \centering
    \small
    {\rowcolors{2}{white}{gray!10}\begin{tabular}{lccc}
    \toprule
        \textbf{Model} & \textbf{B} & \textbf{E} & \textbf{M} \\
    \midrule
    \textit{breaker} & 92.93 & 94.51 & 92.51 \\
    \textit{closed\_blade\_disconnect\_switch} & 49.14 & 50.26 & 48.02 \\
    \textit{closed\_tandem\_disconnect\_switch} & 83.11 & 84.34 & 82.04 \\
    \textit{current\_transformer} & 62.23 & 63.25 & 61.07 \\
    \textit{fuse\_disconnect\_switch} & 90.87 & 92.77 & 90.03 \\
    \textit{glass\_disc\_insulator} & 53.55 & 55.42 & 53.08 \\
    \textit{lightning\_arrester} & 67.87 & 69.28 & 67.41 \\
    \textit{muffle} & 88.27 & 90.06 & 87.92 \\
    \textit{open\_blade\_disconnect\_switch} & 93.34 & 95.18 & 93.06 \\
    \textit{open\_tandem\_disconnect\_switch} & 83.85 & 85.54 & 83.51 \\
    \textit{porcelain\_pin\_insulator} & 30.11 & 32.03 & 30.05 \\
    \textit{potential\_transformer} & 78.54 & 80.72 & 78.01 \\
    \textit{power\_transformer} & 85.73 & 87.79 & 85.50 \\
    \textit{recloser} & 68.11 & 70.40 & 67.50 \\
    \textit{tripolar\_disconnect\_switch} & 83.22 & 85.24 & 82.79 \\
    \midrule
    mIoU & 73.45 & 75.79 & 72.58 \\
    \bottomrule
    \end{tabular}}
    \caption{Accuracy comparison of three DeepLabv3Plus models with the ResNet101 backbone at different stages: base (B), enhanced (E), and mobile (M) in terms of average IoU (\%) for each category and mIoU (\%) on the Substation equipment validation set.}
    \label{tab:substation-results}
\end{table}

\subsection{Model Improvement via XAI-guided Data Augmentation}
After evaluating the segmentation performance of the base model on the validation set $\mathcal{D}_\text{val}$, we analyze the corresponding explanation images generated by RISE, the recommended XAI method $\mathcal{X}$ identified in Experiment 1. 
RISE analysis reveals systematic failure patterns in object detection and segmentation across specific equipment categories, particularly \textit{recloser}, \textit{glass_disc_insulator}, \textit{porcelain_pin_insulator}, \textit{current_transformer}, \textit{lightning_arrester} and \textit{closed_blade_disconnect_switches}, resulting in IoU scores below 70\%.
As shown in Figure \ref{fig:text-exp-recloser}, the explanation maps indicate insufficient model activations on target objects, especially under challenging imaging conditions such as (1) small spatial dimensions relative to the image size (as depicted in Table \ref{tab:substation-detailed}), (2) low contrast due to similar visual features with surrounding infrastructure, or (3) truncated or out-of-frame object instances at image boundaries. 

\begin{figure}[h!]
    \centering
    \includegraphics[width=\linewidth]{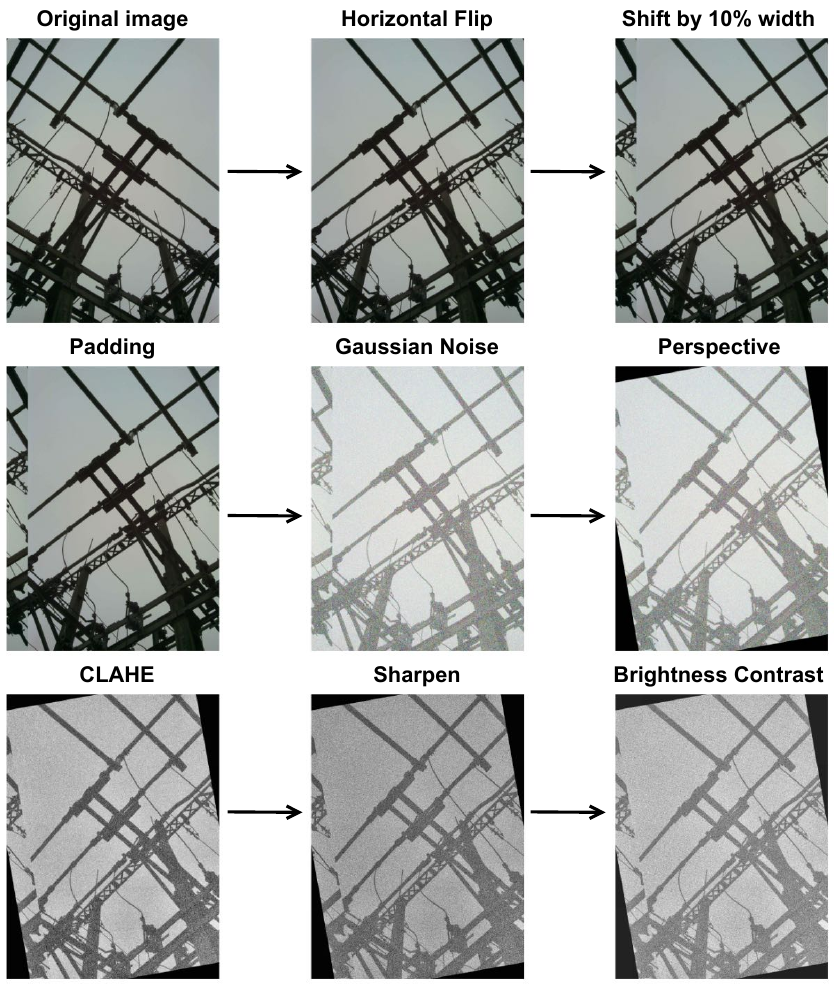}
    \caption{Augmentation techniques on the training set of the Substation equipment dataset.}
    \label{fig:augmentation}
\end{figure}

To address this issue, we employ a sequence of image augmentation techniques \cite{buslaev2020albumentations} to enhance the training data, as shown in Figure \ref{fig:augmentation}. The augmentation pipeline includes horizontal flipping, horizontal shifting, padding, Gaussian noise addition, perspective transformation, Contrast Limited Adaptive Histogram Equalization (CLAHE), sharpening, and brightness and contrast adjustments. These techniques aim to improve the model's robustness and generalization ability by simulating real-world variations in the training data.

After applying the image augmentation techniques to the training set $\mathcal{D}_\text{train}$, we retrain the model and implement the mobile optimization algorithm (Algorithm \ref{alg:model-optimization}) to obtain the enhanced model. Table \ref{tab:substation-results} shows that the enhanced model $\Tilde{\Theta}$ improves across all categories, with an overall mIoU of 75.79\%, surpassing the base model's performance.
Table \ref{tab:substation-results} presents the mobile model's performance. Despite undergoing quantization and pruning, the mobile model $\theta$ maintains a competitive overall mIoU of 72.58\%, experiencing only a slight decrease compared to the enhanced model. Its performance across individual categories remained consistent, with the largest drop being less than 3\% for the \textit{porcelain_pin_insulator} category.

\subsection{Textual Explanation}
\begin{figure*}[h!]
\begin{definition}[label=template:C,
  title={Template \thetcbcounter: Textual Explanations of the Substation Equipment dataset}
]
\begin{minipage}{\linewidth}
    \centering
    \includegraphics[width=.7\linewidth]{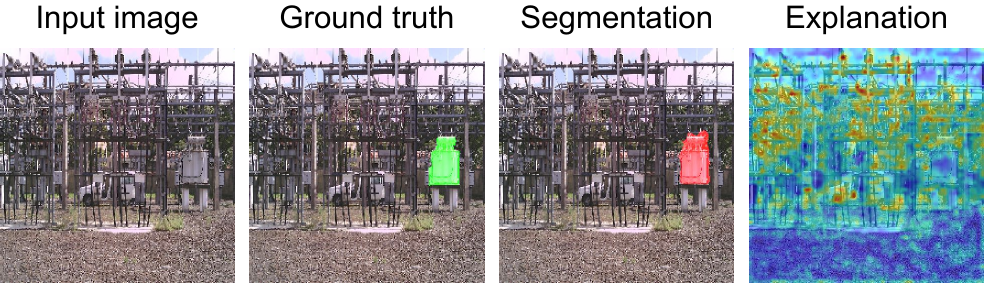}
    \caption{Good segmentation - Category: \textit{power_transformer}}
    \label{fig:text-exp-power-trans}
\end{minipage}
\hfill
\begin{minipage}{\linewidth}
    \textbf{Model Output:} \\
    - The explanation map highlights several regions of importance, with some focus around the power transformer but also attention on other structures.
    
    - The central structures and parts of the power transformer have significant focus in the explanation map.
    
    - The ground and less critical background structures have the least focus.
    
    - The focused regions in the explanation map partially support the prediction for the power transformer. The prediction is accurate, but the explanation map indicates that the model's focus is distributed, suggesting it uses broader contextual cues.
\end{minipage}
\\
\vspace{2pt}
\\
\textcolor[RGB]{0,0,0}{\rule{\linewidth}{0.2pt}}
\\
\vspace{2pt}
\\
\begin{minipage}{1\linewidth}
    \centering
    \includegraphics[width=.7\linewidth]{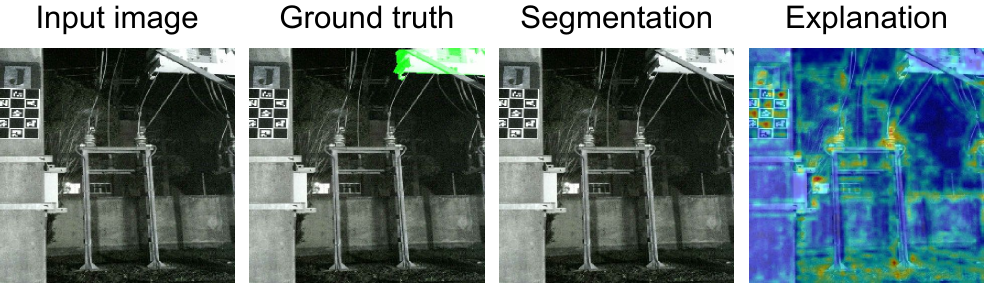}
    \caption{Bad segmentation - Category: \textit{recloser}}
    \label{fig:text-exp-recloser}
\end{minipage}
\hfill
\begin{minipage}{\linewidth}
    \textbf{Model Output:} \\
    - The explanation map highlights regions of importance, with significant focus on the left-side components and connectors, rather than the recloser.
    
    - The labeled components and connectors on the left are the most focused regions in the explanation map.
    
    - The background and some parts of the ground and structure are the least focused regions.
    
    - The focused regions in the explanation map do not support the prediction for the recloser. The model failed to segment the recloser, indicating a significant issue with the model's ability to recognize and segment this category.
\end{minipage}
\end{definition}
\end{figure*}

To generate textual explanations, we follow the prompting structure presented in Template \ref{template:A}. Specifically, 
Template \ref{template:C} provides the textual explanations generated by the LVLM for two segmentation cases: successful segmentation of an object in the \textit{power_transformer} category (Figure \ref{fig:text-exp-power-trans}) and unsuccessful segmentation of an object in the \textit{recloser} category (Figure \ref{fig:text-exp-recloser}). The results show that the LVLM effectively provided human-understandable textual explanations for both cases. 

In the successful segmentation case, the LVLM highlights the central structures and components of the \textit{power_transformer} as the primary focus in the explanation map, while the ground and less critical background structures received minimal attention. It also notes that these focused regions partially supported the prediction, indicating that the model utilized broader contextual cues for accurate segmentation.

In the failure case, the LVLM highlights the labeled components and connectors on the left as the primary focus in the explanation map rather than the \textit{recloser} itself. It observes that these focused regions did not support the prediction for the \textit{recloser}, indicating a significant issue with the model's ability to recognize and segment this category.

\section{Discussion}\label{s:discussion}
In this section, we discuss the key findings of our results, highlighting both the advantages and limitations of the proposed framework. We also offer a variety of suggestions for future development.

\subsection{XAI-guided Data Augmentation for Model Enhancement and Deployment on the Edge}
Our experimental results highlight the effectiveness of the proposed XAI-integrated Visual Quality Inspection framework for industrial assets in both cloud and edge computing contexts. Quantitative and qualitative analyses demonstrate that integrating XAI techniques significantly enhances the performance and interpretability of semantic segmentation models deployed on mobile devices.

A key aspect of our framework is the application of XAI methods, such as RISE, to guide data augmentation and improve model performance. By leveraging insights from the saliency maps generated by RISE, domain experts can refine training data annotations, particularly in areas where the model struggled with segmentation. This process significantly improved segmentation accuracy, especially for challenging categories. The enhanced DeepLabv3Plus-ResNet101 model's higher IoU scores compared to the base model highlight the value of integrating XAI into the model development process for edge deployment.

Another crucial feature of our framework is optimizing models for edge devices to enable real-time field inspections. Our results indicate that the quantized and pruned edge models maintain competitive performance while significantly reducing model size and computational requirements. This optimization is essential for efficient on-device inference, making the visual quality inspection system more practical and accessible for field engineers using edge devices with limited resources.

\subsection{Human-centered Textual Explanation}
Integrating textual explanations generated by \ac{LVLM} greatly enhances the interpretability and applicability of edge-deployed applications. These clear and intuitive summaries of the model's decision-making process help field engineers better understand and trust the segmentation results on their mobile devices. Specifically, the explanations accurately highlight the most and least focused regions in the saliency maps, allowing users to effectively assess the quality of the model's predictions by comparing the explanation maps with ground truth annotations. Delivering interpretable insights directly on mobile devices is essential for field engineers to make informed decisions and confidently rely on the system's outputs.

Overall, our proposed framework demonstrates the feasibility and benefits of generating human-readable explanations alongside saliency maps at the edge, resulting in more transparent and trustworthy visual quality inspection systems.

\subsection{Limitations and Future Developments}
While our \ac{XAI}-integrated Visual Quality Inspection framework shows promising results, there exist some limitations that should be considered.

Firstly, the \ac{XAI}-guided annotation augmentation process relies on the expertise and manual effort of domain experts. Automating or semi-automating this process could enhance scalability and efficiency, particularly for large-scale edge deployments. Future work could explore techniques to automate the annotation refinement process based on \ac{XAI} insights.

Secondly, generating textual explanations on edge devices may introduce latency and computational overhead. Although the current framework shows this approach is feasible, it also highlights the need for further research to address these drawbacks. This could involve exploring more lightweight language models or developing techniques specifically designed for edge computing scenarios.

Lastly, the framework's applicability to other visual inspection domains in edge computing scenarios remains to be explored. We recommend future work investigate the generalizability of the proposed approach across different visual quality inspection tasks and evaluate its performance and interpretability on various datasets and edge computing environments.

\section{Conclusion}\label{sec:conclusion}
%
This paper introduces an \ac{XAI}-integrated Visual Quality Inspection framework for industrial assets, meeting the increasing demand for explainable and trustworthy AI systems in edge computing environments. Our framework combines cutting-edge semantic segmentation models, \ac{XAI} techniques, quantization algorithms, \ac{LVLM}, and edge deployment to enhance both the performance and interpretability of visual quality inspection systems for end-users.
Our experimental results demonstrated the remarkable effectiveness of \ac{XAI} methods in data improvement and augmentation, which led to enhanced model accuracy and performance, especially for challenging and complex categories.

A notable feature of our framework is the enhancement module, which optimizes the model for edge devices through quantization and pruning, making it more lightweight and efficient while maintaining competitive performance. Additionally, integrating textual explanations generated by \ac{LVLM} significantly enhances the interpretability of the edge-deployed application. Last but not least, providing interpretable insights directly on edge devices empowers field engineers to make informed decisions and increases their confidence in the system's outputs.

On the other hand, the framework has certain limitations, including the human-in-the-loop effort required for \ac{XAI}-guided annotation augmentation and the potential latency and computational overhead introduced by generating explanations on edge devices. Future work could explore methods to automate the annotation refinement process and optimize the explanation generation stage for edge computing scenarios.

In conclusion, despite all the limitations discussed, our work has successfully paved the way for more transparent, trustworthy, and effective industrial visual inspection systems deployed at the edge. By addressing key challenges and demonstrating the feasibility of our approach, we lay a solid foundation for future advancements. This can significantly enhance the reliability and performance of edge-based visual inspection technologies, ultimately benefiting both end-users and developers by providing more accurate, interpretable, and efficient solutions.

\section*{Acknowledgment} 
This work was funded by the NBIF Talent Recruitment Fund (TRF2003-001) and the UNB-FCS Startup Fund (22-23 START UP/ H CAO). The computing resource used for this study was partially supported by CFI Project Number 39473 - Smart Campus Integration and Testing (SCIT Lab).
 
\biboptions{sort&compress}
\bibliographystyle{elsarticle-num}
\bibliography{ref}

\begin{thebibliography}{100}
\expandafter\ifx\csname url\endcsname\relax
  \def\url#1{\texttt{#1}}\fi
\expandafter\ifx\csname urlprefix\endcsname\relax\def\urlprefix{URL }\fi
\expandafter\ifx\csname href\endcsname\relax
  \def\href#1#2{#2} \def\path#1{#1}\fi

\bibitem{garouani2022towards}
M.~Garouani, A.~Ahmad, M.~Bouneffa, M.~Hamlich, G.~Bourguin, A.~Lewandowski, Towards big industrial data mining through explainable automated machine learning, The International Journal of Advanced Manufacturing Technology 120~(1) (2022) 1169--1188.

\bibitem{nguyen2023towards}
T.~T.~H. Nguyen, V.~B. Truong, V.~T.~K. Nguyen, Q.~H. Cao, Q.~K. Nguyen, Towards trust of explainable ai in thyroid nodule diagnosis, in: International Workshop on Health Intelligence, Springer, 2023, pp. 11--26.

\bibitem{wu2021locally}
T.-Y. Wu, Y.-T. Wang, Locally interpretable one-class anomaly detection for credit card fraud detection, in: 2021 International Conference on Technologies and Applications of Artificial Intelligence (TAAI), IEEE, 2021, pp. 25--30.

\bibitem{xu2020explainable}
Y.~Xu, X.~Yang, L.~Gong, H.-C. Lin, T.-Y. Wu, Y.~Li, N.~Vasconcelos, Explainable object-induced action decision for autonomous vehicles, in: Proceedings of the IEEE/CVF Conference on Computer Vision and Pattern Recognition, 2020, pp. 9523--9532.

\bibitem{lr7}
V.~Bento, M.~Kohler, P.~Diaz, L.~Mendoza, M.~A. Pacheco, \href{https://doi.org/10.1007/s44163-021-00008-y}{Improving deep learning performance by using explainable artificial intelligence (xai) approaches}, Discover Artificial Intelligence 1~(1) (2021) 9.
\newblock \href {https://doi.org/10.1007/s44163-021-00008-y} {\path{doi:10.1007/s44163-021-00008-y}}.
\newline\urlprefix\url{https://doi.org/10.1007/s44163-021-00008-y}

\bibitem{lr9}
S.~Teso, K.~Kersting, \href{https://doi.org/10.1145/3306618.3314293}{Explanatory interactive machine learning}, in: Proceedings of the 2019 AAAI/ACM Conference on AI, Ethics, and Society, AIES '19, Association for Computing Machinery, New York, NY, USA, 2019, p. 239–245.
\newblock \href {https://doi.org/10.1145/3306618.3314293} {\path{doi:10.1145/3306618.3314293}}.
\newline\urlprefix\url{https://doi.org/10.1145/3306618.3314293}

\bibitem{lr11}
S.~A. Bargal, A.~Zunino, V.~Petsiuk, J.~Zhang, K.~Saenko, V.~Murino, S.~Sclaroff, Guided zoom: Questioning network evidence for fine-grained classification (2020).
\newblock \href {http://arxiv.org/abs/1812.02626} {\path{arXiv:1812.02626}}.

\bibitem{weber2022beyond}
L.~Weber, S.~Lapuschkin, A.~Binder, W.~Samek, Beyond explaining: Opportunities and challenges of xai-based model improvement, Information Fusion (2022).

\bibitem{clement2023coping}
T.~Clement, H.~T.~T. Nguyen, N.~Kemmerzell, M.~Abdelaal, D.~Stjelja, Coping with data distribution shifts: Xai-based adaptive learning with shap clustering for energy consumption prediction, in: Australasian Joint Conference on Artificial Intelligence, Springer, 2023, pp. 147--159.

\bibitem{lr22}
S.-K. Yeom, P.~Seegerer, S.~Lapuschkin, A.~Binder, S.~Wiedemann, K.-R. Müller, W.~Samek, \href{https://www.sciencedirect.com/science/article/pii/S0031320321000868}{Pruning by explaining: A novel criterion for deep neural network pruning}, Pattern Recognition 115 (2021) 107899.
\newblock \href {https://doi.org/https://doi.org/10.1016/j.patcog.2021.107899} {\path{doi:https://doi.org/10.1016/j.patcog.2021.107899}}.
\newline\urlprefix\url{https://www.sciencedirect.com/science/article/pii/S0031320321000868}

\bibitem{lr23}
M.~Sabih, F.~Hannig, J.~Teich, Utilizing explainable ai for quantization and pruning of deep neural networks (2020).
\newblock \href {http://arxiv.org/abs/2008.09072} {\path{arXiv:2008.09072}}.

\bibitem{lr24}
D.~Becking, M.~Dreyer, W.~Samek, K.~M{\"u}ller, S.~Lapuschkin, \href{https://doi.org/10.1007/978-3-031-04083-2_14}{ECQ: Explainability-Driven Quantization for Low-Bit and Sparse DNNs}, Springer International Publishing, Cham, 2022, pp. 271--296.
\newblock \href {https://doi.org/10.1007/978-3-031-04083-2_14} {\path{doi:10.1007/978-3-031-04083-2_14}}.
\newline\urlprefix\url{https://doi.org/10.1007/978-3-031-04083-2_14}

\bibitem{10.1145/3580305.3599578}
J.~Gama, S.~Nowaczyk, S.~Pashami, R.~P. Ribeiro, G.~J. Nalepa, B.~Veloso, \href{https://doi.org/10.1145/3580305.3599578}{Xai for predictive maintenance}, in: Proceedings of the 29th ACM SIGKDD Conference on Knowledge Discovery and Data Mining, KDD '23, Association for Computing Machinery, New York, NY, USA, 2023, p. 5798–5799.
\newblock \href {https://doi.org/10.1145/3580305.3599578} {\path{doi:10.1145/3580305.3599578}}.
\newline\urlprefix\url{https://doi.org/10.1145/3580305.3599578}

\bibitem{molnar2022}
C.~Molnar, \href{https://christophm.github.io/interpretable-ml-book}{Interpretable Machine Learning}, 2nd Edition, 2022.
\newline\urlprefix\url{https://christophm.github.io/interpretable-ml-book}

\bibitem{arrieta2020explainable}
A.~B. Arrieta, N.~D{\'\i}az-Rodr{\'\i}guez, J.~Del~Ser, A.~Bennetot, S.~Tabik, A.~Barbado, S.~Garc{\'\i}a, S.~Gil-L{\'o}pez, D.~Molina, R.~Benjamins, et~al., Explainable artificial intelligence (xai): Concepts, taxonomies, opportunities and challenges toward responsible ai, Information fusion 58 (2020) 82--115.

\bibitem{zha2023data}
D.~Zha, Z.~P. Bhat, K.-H. Lai, F.~Yang, X.~Hu, Data-centric ai: Perspectives and challenges, in: Proceedings of the 2023 SIAM International Conference on Data Mining (SDM), SIAM, 2023, pp. 945--948.

\bibitem{bodria2023benchmarking}
F.~Bodria, F.~Giannotti, R.~Guidotti, F.~Naretto, D.~Pedreschi, S.~Rinzivillo, Benchmarking and survey of explanation methods for black box models, Data Mining and Knowledge Discovery 37~(5) (2023) 1719--1778.

\bibitem{rlb4}
V.~Bento, M.~Kohler, P.~Diaz, L.~Mendoza, M.~A. Pacheco, \href{https://doi.org/10.1007/s44163-021-00008-y}{Improving deep learning performance by using explainable artificial intelligence (xai) approaches}, Discover Artificial Intelligence 1~(1) (2021) 9.
\newblock \href {https://doi.org/10.1007/s44163-021-00008-y} {\path{doi:10.1007/s44163-021-00008-y}}.
\newline\urlprefix\url{https://doi.org/10.1007/s44163-021-00008-y}

\bibitem{lorente2021explaining}
M.~P.~S. Lorente, E.~M. Lopez, L.~A. Florez, A.~L. Espino, J.~A.~I. Mart{\'\i}nez, A.~S. de~Miguel, Explaining deep learning-based driver models, Applied Sciences 11~(8) (2021) 3321.

\bibitem{rlb5}
Z.~Li, \href{https://www.sciencedirect.com/science/article/pii/S2214367X22001466}{Leveraging explainable artificial intelligence and big trip data to understand factors influencing willingness to ridesharing}, Travel Behaviour and Society 31 (2023) 284--294.
\newblock \href {https://doi.org/https://doi.org/10.1016/j.tbs.2022.12.006} {\path{doi:https://doi.org/10.1016/j.tbs.2022.12.006}}.
\newline\urlprefix\url{https://www.sciencedirect.com/science/article/pii/S2214367X22001466}

\bibitem{dikmen2022effects}
M.~Dikmen, C.~Burns, The effects of domain knowledge on trust in explainable ai and task performance: A case of peer-to-peer lending, International Journal of Human-Computer Studies 162 (2022) 102792.

\bibitem{rlb6}
X.~Dastile, T.~Celik, Making deep learning-based predictions for credit scoring explainable, IEEE Access 9 (2021) 50426--50440.
\newblock \href {https://doi.org/10.1109/ACCESS.2021.3068854} {\path{doi:10.1109/ACCESS.2021.3068854}}.

\bibitem{vzlahtivc2023agile}
B.~{\v{Z}}lahti{\v{c}}, J.~Zavr{\v{s}}nik, H.~Bla{\v{z}}un~Vo{\v{s}}ner, P.~Kokol, D.~{\v{S}}uran, T.~Zavr{\v{s}}nik, Agile machine learning model development using data canyons in medicine: A step towards explainable artificial intelligence and flexible expert-based model improvement, Applied Sciences 13~(14) (2023) 8329.

\bibitem{rlb1}
P.~Guleria, P.~Naga~Srinivasu, S.~Ahmed, N.~Almusallam, F.~K. Alarfaj, \href{https://www.mdpi.com/2079-9292/11/24/4086}{Xai framework for cardiovascular disease prediction using classification techniques}, Electronics 11~(24) (2022).
\newblock \href {https://doi.org/10.3390/electronics11244086} {\path{doi:10.3390/electronics11244086}}.
\newline\urlprefix\url{https://www.mdpi.com/2079-9292/11/24/4086}

\bibitem{melo2022use}
E.~Melo, I.~Silva, D.~G. Costa, C.~M. Viegas, T.~M. Barros, On the use of explainable artificial intelligence to evaluate school dropout, Education Sciences 12~(12) (2022) 845.

\bibitem{nur2022explainable}
N.~Nur, A.~Benedict, O.~Eltayeby, W.~Dou, M.~Dorodchi, X.~Niu, M.~Maher, C.~Chambers, Explainable ai for data driven learning analytics: A holistic approach to engage advisors in knowledge discovery, in: EDULEARN22 Proceedings, IATED, 2022, pp. 10300--10306.

\bibitem{tsiakmaki2021case}
M.~Tsiakmaki, O.~Ragos, A case study of interpretable counterfactual explanations for the task of predicting student academic performance, in: 2021 25th International Conference on Circuits, Systems, Communications and Computers (CSCC), IEEE, 2021, pp. 120--125.

\bibitem{kardovskyi2021artificial}
Y.~Kardovskyi, S.~Moon, Artificial intelligence quality inspection of steel bars installation by integrating mask r-cnn and stereo vision, Automation in Construction 130 (2021) 103850.

\bibitem{diaz2021guided}
M.~T.~G. Diaz, D.~Ghosh, A.~Arantes, H.~Khorosgani, M.~Alam, G.~Sin, C.~Gupta, Guided visual inspection enabled by ai-based detection models, in: 2021 IEEE International Conference on Prognostics and Health Management (ICPHM), IEEE, 2021, pp. 1--8.

\bibitem{lr1}
C.~Eiras-Franco, B.~Guijarro-Berdiñas, A.~Alonso-Betanzos, A.~Bahamonde, \href{https://www.sciencedirect.com/science/article/pii/S0167923619301708}{A scalable decision-tree-based method to explain interactions in dyadic data}, Decision Support Systems 127 (2019) 113141.
\newblock \href {https://doi.org/https://doi.org/10.1016/j.dss.2019.113141} {\path{doi:https://doi.org/10.1016/j.dss.2019.113141}}.
\newline\urlprefix\url{https://www.sciencedirect.com/science/article/pii/S0167923619301708}

\bibitem{lr2}
M.~A. Islam, D.~T. Anderson, A.~J. Pinar, T.~C. Havens, G.~Scott, J.~M. Keller, Enabling explainable fusion in deep learning with fuzzy integral neural networks, IEEE Transactions on Fuzzy Systems 28~(7) (2020) 1291--1300.
\newblock \href {https://doi.org/10.1109/TFUZZ.2019.2917124} {\path{doi:10.1109/TFUZZ.2019.2917124}}.

\bibitem{lr3}
Ádamo L.~{de Santana}, C.~R. Francês, C.~A. Rocha, S.~V. Carvalho, N.~L. Vijaykumar, L.~P. Rego, J.~C. Costa, \href{https://www.sciencedirect.com/science/article/pii/S0169023X06001959}{Strategies for improving the modeling and interpretability of bayesian networks}, Data \& Knowledge Engineering 63~(1) (2007) 91--107, data Warehouse and Knowledge Discovery (DAWAK ’05).
\newblock \href {https://doi.org/https://doi.org/10.1016/j.datak.2006.10.005} {\path{doi:https://doi.org/10.1016/j.datak.2006.10.005}}.
\newline\urlprefix\url{https://www.sciencedirect.com/science/article/pii/S0169023X06001959}

\bibitem{lr4}
Y.~Rong, T.~Leemann, T.~Nguyen, L.~Fiedler, P.~Qian, V.~Unhelkar, T.~Seidel, G.~Kasneci, E.~Kasneci, Towards human-centered explainable ai: A survey of user studies for model explanations, IEEE Transactions on Pattern Analysis \& Machine Intelligence 46~(04) (2024) 2104--2122.
\newblock \href {https://doi.org/10.1109/TPAMI.2023.3331846} {\path{doi:10.1109/TPAMI.2023.3331846}}.

\bibitem{lr5}
T.~Clement, N.~Kemmerzell, M.~Abdelaal, M.~Amberg, \href{https://www.mdpi.com/2504-4990/5/1/6}{Xair: A systematic metareview of explainable ai (xai) aligned to the software development process}, Machine Learning and Knowledge Extraction 5~(1) (2023) 78--108.
\newline\urlprefix\url{https://www.mdpi.com/2504-4990/5/1/6}

\bibitem{lr6}
L.~Weber, S.~Lapuschkin, A.~Binder, W.~Samek, \href{https://www.sciencedirect.com/science/article/pii/S1566253522002238}{Beyond explaining: Opportunities and challenges of xai-based model improvement}, Information Fusion 92 (2023) 154--176.
\newblock \href {https://doi.org/https://doi.org/10.1016/j.inffus.2022.11.013} {\path{doi:https://doi.org/10.1016/j.inffus.2022.11.013}}.
\newline\urlprefix\url{https://www.sciencedirect.com/science/article/pii/S1566253522002238}

\bibitem{lr13}
J.~Sun, S.~Lapuschkin, W.~Samek, Y.~Zhao, N.~Cheung, A.~Binder, \href{https://doi.ieeecomputersociety.org/10.1109/ICPR48806.2021.9412941}{Explanation-guided training for cross-domain few-shot classification}, in: 2020 25th International Conference on Pattern Recognition (ICPR), IEEE Computer Society, Los Alamitos, CA, USA, 2021, pp. 7609--7616.
\newblock \href {https://doi.org/10.1109/ICPR48806.2021.9412941} {\path{doi:10.1109/ICPR48806.2021.9412941}}.
\newline\urlprefix\url{https://doi.ieeecomputersociety.org/10.1109/ICPR48806.2021.9412941}

\bibitem{lr14}
A.~Zunino, S.~Bargal, R.~Volpi, M.~Sameki, J.~Zhang, S.~Sclaroff, V.~Murino, K.~Saenko, \href{https://doi.ieeecomputersociety.org/10.1109/CVPRW53098.2021.00361}{Explainable deep classification models for domain generalization}, in: 2021 IEEE/CVF Conference on Computer Vision and Pattern Recognition Workshops (CVPRW), IEEE Computer Society, Los Alamitos, CA, USA, 2021, pp. 3227--3236.
\newblock \href {https://doi.org/10.1109/CVPRW53098.2021.00361} {\path{doi:10.1109/CVPRW53098.2021.00361}}.
\newline\urlprefix\url{https://doi.ieeecomputersociety.org/10.1109/CVPRW53098.2021.00361}

\bibitem{lr16}
J.~Blunk, N.~Penzel, P.~Bodesheim, J.~Denzler, Beyond debiasing: Actively steering feature selection via loss regularization, in: U.~K{\"o}the, C.~Rother (Eds.), Pattern Recognition, Springer Nature Switzerland, Cham, 2024, pp. 394--408.

\bibitem{lr17}
F.~Liu, B.~Avci, \href{https://aclanthology.org/P19-1631}{Incorporating priors with feature attribution on text classification}, in: A.~Korhonen, D.~Traum, L.~M{\`a}rquez (Eds.), Proceedings of the 57th Annual Meeting of the Association for Computational Linguistics, Association for Computational Linguistics, Florence, Italy, 2019, pp. 6274--6283.
\newblock \href {https://doi.org/10.18653/v1/P19-1631} {\path{doi:10.18653/v1/P19-1631}}.
\newline\urlprefix\url{https://aclanthology.org/P19-1631}

\bibitem{lr18}
L.~Rieger, C.~Singh, W.~J. Murdoch, B.~Yu, Interpretations are useful: penalizing explanations to align neural networks with prior knowledge, in: Proceedings of the 37th International Conference on Machine Learning, ICML'20, JMLR.org, 2020.

\bibitem{lr19}
J.~Zhang, S.~A. Bargal, Z.~Lin, J.~Brandt, X.~Shen, S.~Sclaroff, \href{https://doi.org/10.1007/s11263-017-1059-x}{Top-down neural attention by excitation backprop}, International Journal of Computer Vision 126~(10) (2018) 1084--1102.
\newblock \href {https://doi.org/10.1007/s11263-017-1059-x} {\path{doi:10.1007/s11263-017-1059-x}}.
\newline\urlprefix\url{https://doi.org/10.1007/s11263-017-1059-x}

\bibitem{lr20}
G.~Erion, J.~D. Janizek, P.~Sturmfels, S.~M. Lundberg, S.-I. Lee, \href{https://doi.org/10.1038/s42256-021-00343-w}{Improving performance of deep learning models with axiomatic attribution priors and expected gradients}, Nature Machine Intelligence 3~(7) (2021) 620--631.
\newblock \href {https://doi.org/10.1038/s42256-021-00343-w} {\path{doi:10.1038/s42256-021-00343-w}}.
\newline\urlprefix\url{https://doi.org/10.1038/s42256-021-00343-w}

\bibitem{lr21}
V.~Nagisetty, L.~Graves, J.~Scott, V.~Ganesh, xai-gan: Enhancing generative adversarial networks via explainable ai systems (2022).
\newblock \href {http://arxiv.org/abs/2002.10438} {\path{arXiv:2002.10438}}.

\bibitem{simonyan2014very}
K.~Simonyan, A.~Zisserman, Very deep convolutional networks for large-scale image recognition, arXiv preprint arXiv:1409.1556 (2014).

\bibitem{redmon2016you}
J.~Redmon, S.~Divvala, R.~Girshick, A.~Farhadi, You only look once: Unified, real-time object detection, in: Proceedings of the IEEE conference on computer vision and pattern recognition, 2016, pp. 779--788.

\bibitem{he2016deep}
K.~He, X.~Zhang, S.~Ren, J.~Sun, Deep residual learning for image recognition, in: Proceedings of the IEEE conference on computer vision and pattern recognition, 2016, pp. 770--778.

\bibitem{howard2017mobilenets}
A.~G. Howard, M.~Zhu, B.~Chen, D.~Kalenichenko, W.~Wang, T.~Weyand, M.~Andreetto, H.~Adam, Mobilenets: Efficient convolutional neural networks for mobile vision applications, arXiv preprint arXiv:1704.04861 (2017).

\bibitem{long2015fully}
J.~Long, E.~Shelhamer, T.~Darrell, Fully convolutional networks for semantic segmentation, in: Proceedings of the IEEE conference on computer vision and pattern recognition, 2015, pp. 3431--3440.

\bibitem{howard2019searching}
A.~Howard, M.~Sandler, G.~Chu, L.-C. Chen, B.~Chen, M.~Tan, W.~Wang, Y.~Zhu, R.~Pang, V.~Vasudevan, et~al., Searching for mobilenetv3, in: Proceedings of the IEEE/CVF international conference on computer vision, 2019, pp. 1314--1324.

\bibitem{deeplabv32018}
L.-C. Chen, G.~Papandreou, F.~Schroff, H.~Adam, Rethinking atrous convolution for semantic image segmentation, arXiv:1706.05587 (2017).

\bibitem{chen2016deeplab}
L.-C. Chen, G.~Papandreou, I.~Kokkinos, K.~Murphy, A.~L. Yuille, Deeplab: Semantic image segmentation with deep convolutional nets, atrous convolution, and fully connected crfs. arxiv 2016, arXiv preprint arXiv:1606.00915 1 (2016).

\bibitem{yang2023semantic}
Z.~Yang, Semantic segmentation method based on improved deeplabv3+, in: International Conference on Cloud Computing, Performance Computing, and Deep Learning (CCPCDL 2023), Vol. 12712, SPIE, 2023, pp. 32--37.

\bibitem{vinogradova2020towards}
K.~Vinogradova, A.~Dibrov, G.~Myers, Towards interpretable semantic segmentation via gradient-weighted class activation mapping (student abstract), in: Proceedings of the AAAI conference on artificial intelligence, Vol.~34, 2020, pp. 13943--13944.

\bibitem{rebuffi2020there}
S.-A. Rebuffi, R.~Fong, X.~Ji, A.~Vedaldi, There and back again: Revisiting backpropagation saliency methods, in: Proceedings of the IEEE/CVF Conference on Computer Vision and Pattern Recognition, 2020, pp. 8839--8848.

\bibitem{bach2015pixel}
S.~Bach, A.~Binder, G.~Montavon, F.~Klauschen, K.-R. M{\"u}ller, W.~Samek, On pixel-wise explanations for non-linear classifier decisions by layer-wise relevance propagation, PloS one 10~(7) (2015) e0130140.

\bibitem{shrikumar2017learning}
A.~Shrikumar, P.~Greenside, A.~Kundaje, Learning important features through propagating activation differences, in: International conference on machine learning, PMLR, 2017, pp. 3145--3153.

\bibitem{zhou2016learning}
B.~Zhou, A.~Khosla, A.~Lapedriza, A.~Oliva, A.~Torralba, Learning deep features for discriminative localization, in: Proceedings of the IEEE conference on computer vision and pattern recognition, 2016, pp. 2921--2929.

\bibitem{chattopadhay2018grad}
A.~Chattopadhay, A.~Sarkar, P.~Howlader, V.~N. Balasubramanian, Grad-cam++: Generalized gradient-based visual explanations for deep convolutional networks, in: 2018 IEEE winter conference on applications of computer vision (WACV), IEEE, 2018, pp. 839--847.

\bibitem{selvaraju2017grad}
R.~R. Selvaraju, M.~Cogswell, A.~Das, R.~Vedantam, D.~Parikh, D.~Batra, Grad-cam: Visual explanations from deep networks via gradient-based localization, in: Proceedings of the IEEE international conference on computer vision, 2017, pp. 618--626.

\bibitem{wang2020score}
H.~Wang, Z.~Wang, M.~Du, F.~Yang, Z.~Zhang, S.~Ding, P.~Mardziel, X.~Hu, Score-cam: Score-weighted visual explanations for convolutional neural networks, in: Proceedings of the IEEE/CVF conference on computer vision and pattern recognition workshops, 2020, pp. 24--25.

\bibitem{ramaswamy2020ablation}
H.~G. Ramaswamy, et~al., Ablation-cam: Visual explanations for deep convolutional network via gradient-free localization, in: Proceedings of the IEEE/CVF Winter Conference on Applications of Computer Vision, 2020, pp. 983--991.

\bibitem{muhammad2020eigen}
M.~B. Muhammad, M.~Yeasin, Eigen-cam: Class activation map using principal components, in: 2020 International Joint Conference on Neural Networks (IJCNN), IEEE, 2020, pp. 1--7.

\bibitem{fu2020axiom}
R.~Fu, Q.~Hu, X.~Dong, Y.~Guo, Y.~Gao, B.~Li, Axiom-based grad-cam: Towards accurate visualization and explanation of cnns, arXiv preprint arXiv:2008.02312 (2020).

\bibitem{nguyen2022secam}
P.~X. Nguyen, H.~Q. Cao, K.~V. Nguyen, H.~Nguyen, T.~Yairi, Secam: Tightly accelerate the image explanation via region-based segmentation, IEICE TRANSACTIONS on Information and Systems 105~(8) (2022) 1401--1417.

\bibitem{hasany2023seg}
S.~N. Hasany, C.~Petitjean, F.~M{\'e}riaudeau, Seg-xres-cam: Explaining spatially local regions in image segmentation, in: Proceedings of the IEEE/CVF Conference on Computer Vision and Pattern Recognition, 2023, pp. 3732--3737.

\bibitem{nguyen2024efficient}
Q.~K. Nguyen, T.~T.~H. Nguyen, V.~T.~K. Nguyen, V.~B. Truong, T.~Phan, H.~Cao, Efficient and concise explanations for object detection with gaussian-class activation mapping explainer, arXiv preprint arXiv:2404.13417 (2024).

\bibitem{zeiler2014visualizing}
M.~Zeiler, Visualizing and understanding convolutional networks, in: European conference on computer vision/arXiv, Vol. 1311, 2014.

\bibitem{lr10}
M.~T. Ribeiro, S.~Singh, C.~Guestrin, "why should i trust you?": Explaining the predictions of any classifier (2016).
\newblock \href {http://arxiv.org/abs/1602.04938} {\path{arXiv:1602.04938}}.

\bibitem{petsiuk2018rise}
V.~Petsiuk, A.~Das, K.~Saenko, Rise: Randomized input sampling for explanation of black-box models, arXiv preprint arXiv:1806.07421 (2018).

\bibitem{petsiuk2021black}
V.~Petsiuk, R.~Jain, V.~Manjunatha, V.~I. Morariu, A.~Mehra, V.~Ordonez, K.~Saenko, Black-box explanation of object detectors via saliency maps, in: Proceedings of the IEEE/CVF Conference on Computer Vision and Pattern Recognition, 2021, pp. 11443--11452.

\bibitem{yang2021mfpp}
Q.~Yang, X.~Zhu, J.-K. Fwu, Y.~Ye, G.~You, Y.~Zhu, Mfpp: Morphological fragmental perturbation pyramid for black-box model explanations, in: 2020 25th International conference on pattern recognition (ICPR), IEEE, 2021, pp. 1376--1383.

\bibitem{truong2023towards}
V.~B. Truong, T.~T.~H. Nguyen, V.~T.~K. Nguyen, Q.~K. Nguyen, Q.~H. Cao, \href{https://proceedings.mlr.press/v222/truong24a.html}{Towards better explanations for object detection}, in: B.~Yanıkoğlu, W.~Buntine (Eds.), Proceedings of the 15th Asian Conference on Machine Learning, Vol. 222 of Proceedings of Machine Learning Research, PMLR, 2024, pp. 1385--1400.
\newline\urlprefix\url{https://proceedings.mlr.press/v222/truong24a.html}

\bibitem{sacha2023protoseg}
M.~Sacha, D.~Rymarczyk, {\L}.~Struski, J.~Tabor, B.~Zieli{\'n}ski, Protoseg: Interpretable semantic segmentation with prototypical parts, in: Proceedings of the IEEE/CVF Winter Conference on Applications of Computer Vision, 2023, pp. 1481--1492.

\bibitem{heide2021x}
N.~F. Heide, E.~M{\"u}ller, J.~Petereit, M.~Heizmann, X 3 seg: model-agnostic explanations for the semantic segmentation of 3d point clouds with prototypes and criticism, in: 2021 IEEE International Conference on Image Processing (ICIP), IEEE, 2021, pp. 3687--3691.

\bibitem{draelos2020hirescam}
R.~L. Draelos, L.~Carin, Hirescam: Faithful location representation in visual attention for explainable 3d medical image classification, arXiv preprint arXiv:2011.08891 (2020).

\bibitem{jacobgilpytorchcam}
J.~Gildenblat, contributors, Pytorch library for cam methods, \url{https://github.com/jacobgil/pytorch-grad-cam} (2021).

\bibitem{tang2017manufacturing}
H.~Tang, Manufacturing system and process development for vehicle assembly, SAE International, 2017.

\bibitem{sun2018research}
X.~Sun, J.~Gu, S.~Tang, J.~Li, Research progress of visual inspection technology of steel products—a review, Applied Sciences 8~(11) (2018) 2195.

\bibitem{md2022review}
A.~Q. Md, K.~Jha, S.~Haneef, A.~K. Sivaraman, K.~F. Tee, A review on data-driven quality prediction in the production process with machine learning for industry 4.0, Processes 10~(10) (2022) 1966.

\bibitem{yasuda2022aircraft}
Y.~D. Yasuda, F.~A. Cappabianco, L.~E.~G. Martins, J.~A. Gripp, Aircraft visual inspection: A systematic literature review, Computers in Industry 141 (2022) 103695.

\bibitem{Ilchuk2023COMPUTER}
M.~Ilchuk, A.~Stadnyk, Computer visual inspection of pear quality, Measuring Equipment and Metrology (2023).
\newblock \href {https://doi.org/10.23939/istcmtm2023.01.025} {\path{doi:10.23939/istcmtm2023.01.025}}.

\bibitem{rovzanec2024adaptive}
J.~M. Ro{\v{z}}anec, B.~{\v{S}}ircelj, B.~Fortuna, D.~Mladeni{\'c}, Adaptive explainable artificial intelligence for visual defect inspection., Procedia Computer Science 232 (2024) 3034--3043.

\bibitem{lupi2023framework}
F.~Lupi, M.~Biancalana, A.~Rossi, M.~Lanzetta, A framework for flexible and reconfigurable vision inspection systems, The International Journal of Advanced Manufacturing Technology 129~(1-2) (2023) 871--897.

\bibitem{gunraj2023soldernet}
H.~Gunraj, P.~Guerrier, S.~Fernandez, A.~Wong, Soldernet: Towards trustworthy visual inspection of solder joints in electronics manufacturing using explainable artificial intelligence, in: Proceedings of the AAAI Conference on Artificial Intelligence, Vol.~37, 2023, pp. 15668--15674.

\bibitem{hoffmann2023systematic}
R.~Hoffmann, C.~Reich, A systematic literature review on artificial intelligence and explainable artificial intelligence for visual quality assurance in manufacturing, Electronics 12~(22) (2023) 4572.

\bibitem{kok2023explainablegreen}
{\.I}.~K{\"o}k, Y.~Ergun, N.~U{\u{g}}ur, Explainable ai-powered edge computing solution for smart building energy management in green iot, in: Low-Cost Digital Solutions for Industrial Automation (LoDiSA 2023), Vol. 2023, IET, 2023, pp. 150--157.

\bibitem{garg2023trusted}
S.~Garg, K.~Kaur, G.~S. Aujla, G.~Kaddoum, P.~Garigipati, M.~Guizani, Trusted explainable ai for 6g-enabled edge cloud ecosystem, IEEE Wireless Communications 30~(3) (2023) 163--170.

\bibitem{dutta2023human}
J.~Dutta, D.~Puthal, Human-centered explainable ai at the edge for ehealth, in: 2023 IEEE International Conference on Edge Computing and Communications (EDGE), IEEE, 2023, pp. 227--232.

\bibitem{wang2019designing}
D.~Wang, Q.~Yang, A.~Abdul, B.~Y. Lim, Designing theory-driven user-centric explainable ai, in: Proceedings of the 2019 CHI conference on human factors in computing systems, 2019, pp. 1--15.

\bibitem{yu2023towards}
S.~Yu, Towards trustworthy and understandable ai: Unraveling explainability strategies on simplifying algorithms, appropriate information disclosure, and high-level collaboration, in: Proceedings of the 26th International Academic Mindtrek Conference, 2023, pp. 133--143.

\bibitem{bertrand2023selective}
A.~Bertrand, T.~Viard, R.~Belloum, J.~R. Eagan, W.~Maxwell, On selective, mutable and dialogic xai: A review of what users say about different types of interactive explanations, in: Proceedings of the 2023 CHI Conference on Human Factors in Computing Systems, 2023, pp. 1--21.

\bibitem{poli2021generation}
J.-P. Poli, W.~Ouerdane, R.~Pierrard, Generation of textual explanations in xai: The case of semantic annotation, in: 2021 IEEE International Conference on Fuzzy Systems (FUZZ-IEEE), IEEE, 2021, pp. 1--6.

\bibitem{park2018multimodal}
D.~H. Park, L.~A. Hendricks, Z.~Akata, A.~Rohrbach, B.~Schiele, T.~Darrell, M.~Rohrbach, Multimodal explanations: Justifying decisions and pointing to the evidence, in: Proceedings of the IEEE conference on computer vision and pattern recognition, 2018, pp. 8779--8788.

\bibitem{hendricks2018grounding}
L.~A. Hendricks, R.~Hu, T.~Darrell, Z.~Akata, Grounding visual explanations, in: Proceedings of the European conference on computer vision (ECCV), 2018, pp. 264--279.

\bibitem{xu2015show}
K.~Xu, J.~Ba, R.~Kiros, K.~Cho, A.~Courville, R.~Salakhudinov, R.~Zemel, Y.~Bengio, Show, attend and tell: Neural image caption generation with visual attention, in: International conference on machine learning, PMLR, 2015, pp. 2048--2057.

\bibitem{kim2018textual}
J.~Kim, A.~Rohrbach, T.~Darrell, J.~Canny, Z.~Akata, Textual explanations for self-driving vehicles, in: Proceedings of the European conference on computer vision (ECCV), 2018, pp. 563--578.

\bibitem{dai2024instructblip}
W.~Dai, J.~Li, D.~Li, A.~M.~H. Tiong, J.~Zhao, W.~Wang, B.~Li, P.~N. Fung, S.~Hoi, Instructblip: Towards general-purpose vision-language models with instruction tuning, Advances in Neural Information Processing Systems 36 (2024).

\bibitem{brown2020language}
T.~Brown, B.~Mann, N.~Ryder, M.~Subbiah, J.~D. Kaplan, P.~Dhariwal, A.~Neelakantan, P.~Shyam, G.~Sastry, A.~Askell, et~al., Language models are few-shot learners, Advances in neural information processing systems 33 (2020) 1877--1901.

\bibitem{chowdhery2023palm}
A.~Chowdhery, S.~Narang, J.~Devlin, M.~Bosma, G.~Mishra, A.~Roberts, P.~Barham, H.~W. Chung, C.~Sutton, S.~Gehrmann, et~al., Palm: Scaling language modeling with pathways, Journal of Machine Learning Research 24~(240) (2023) 1--113.

\bibitem{peng2023kosmos}
Z.~Peng, W.~Wang, L.~Dong, Y.~Hao, S.~Huang, S.~Ma, F.~Wei, Kosmos-2: Grounding multimodal large language models to the world, arXiv preprint arXiv:2306.14824 (2023).

\bibitem{awadalla2023openflamingo}
A.~Awadalla, I.~Gao, J.~Gardner, J.~Hessel, Y.~Hanafy, W.~Zhu, K.~Marathe, Y.~Bitton, S.~Gadre, S.~Sagawa, et~al., Openflamingo: An open-source framework for training large autoregressive vision-language models, arXiv preprint arXiv:2308.01390 (2023).

\bibitem{fuyu-8b}
R.~Bavishi, E.~Elsen, C.~Hawthorne, M.~Nye, A.~Odena, A.~Somani, S.~Ta\c{s}\i{}rlar, \href{https://www.adept.ai/blog/fuyu-8b}{Introducing our multimodal models} (2023).
\newline\urlprefix\url{https://www.adept.ai/blog/fuyu-8b}

\bibitem{chen2022pali}
X.~Chen, X.~Wang, S.~Changpinyo, A.~Piergiovanni, P.~Padlewski, D.~Salz, S.~Goodman, A.~Grycner, B.~Mustafa, L.~Beyer, et~al., Pali: A jointly-scaled multilingual language-image model, arXiv preprint arXiv:2209.06794 (2022).

\bibitem{gpt4vision}
OpenAI, Gpt-4v(ision) system card, \url{https://cdn.openai.com/papers/GPTV_System_Card.pdf}, accessed: 2024-02-14 (2023).

\bibitem{dong2024internlm}
X.~Dong, P.~Zhang, Y.~Zang, Y.~Cao, B.~Wang, L.~Ouyang, S.~Zhang, H.~Duan, W.~Zhang, Y.~Li, et~al., Internlm-xcomposer2-4khd: A pioneering large vision-language model handling resolutions from 336 pixels to 4k hd, arXiv preprint arXiv:2404.06512 (2024).

\bibitem{zhu2023minigpt}
D.~Zhu, J.~Chen, X.~Shen, X.~Li, M.~Elhoseiny, Minigpt-4: Enhancing vision-language understanding with advanced large language models, arXiv preprint arXiv:2304.10592 (2023).

\bibitem{nguyen2024langxai}
H.~Nguyen, T.~Clement, L.~Nguyen, N.~Kemmerzell, B.~Truong, K.~Nguyen, M.~Abdelaal, H.~Cao, \href{https://doi.org/10.24963/ijcai.2024/1025}{Langxai: Integrating large vision models for generating textual explanations to enhance explainability in visual perception tasks}, in: K.~Larson (Ed.), Proceedings of the Thirty-Third International Joint Conference on Artificial Intelligence, {IJCAI-24}, International Joint Conferences on Artificial Intelligence Organization, 2024, pp. 8754--8758.
\newblock \href {https://doi.org/10.24963/ijcai.2024/1025} {\path{doi:10.24963/ijcai.2024/1025}}.
\newline\urlprefix\url{https://doi.org/10.24963/ijcai.2024/1025}

\bibitem{chen2023x}
Y.~Chen, X-iqe: explainable image quality evaluation for text-to-image generation with visual large language models, arXiv preprint arXiv:2305.10843 (2023).

\bibitem{deng2009imagenet}
J.~Deng, W.~Dong, R.~Socher, L.-J. Li, K.~Li, L.~Fei-Fei, Imagenet: A large-scale hierarchical image database, in: 2009 IEEE conference on computer vision and pattern recognition, Ieee, 2009, pp. 248--255.

\bibitem{sudre2017generalised}
C.~H. Sudre, W.~Li, T.~Vercauteren, S.~Ourselin, M.~Jorge~Cardoso, Generalised dice overlap as a deep learning loss function for highly unbalanced segmentations, in: Deep Learning in Medical Image Analysis and Multimodal Learning for Clinical Decision Support: Third International Workshop, DLMIA 2017, and 7th International Workshop, ML-CDS 2017, Held in Conjunction with MICCAI 2017, Qu{\'e}bec City, QC, Canada, September 14, Proceedings 3, Springer, 2017, pp. 240--248.

\bibitem{murphy1996finley}
A.~H. Murphy, The finley affair: A signal event in the history of forecast verification, Weather and forecasting 11~(1) (1996) 3--20.

\bibitem{kingma2014adam}
D.~P. Kingma, Adam: A method for stochastic optimization, arXiv preprint arXiv:1412.6980 (2014).

\bibitem{schulz2020restricting}
K.~Schulz, L.~Sixt, F.~Tombari, T.~Landgraf, Restricting the flow: Information bottlenecks for attribution, arXiv preprint arXiv:2001.00396 (2020).

\bibitem{samek2016evaluating}
W.~Samek, A.~Binder, G.~Montavon, S.~Lapuschkin, K.-R. M{\"u}ller, Evaluating the visualization of what a deep neural network has learned, IEEE transactions on neural networks and learning systems 28~(11) (2016) 2660--2673.

\bibitem{hooker2019benchmark}
S.~Hooker, D.~Erhan, P.-J. Kindermans, B.~Kim, A benchmark for interpretability methods in deep neural networks, Advances in neural information processing systems 32 (2019).

\bibitem{bento2021improving}
V.~Bento, M.~Kohler, P.~Diaz, L.~Mendoza, M.~A. Pacheco, Improving deep learning performance by using explainable artificial intelligence (xai) approaches, Discover Artificial Intelligence 1 (2021) 1--11.

\bibitem{r2-50}
Z.~Yang, L.~Li, K.~Lin, J.~Wang, C.-C. Lin, Z.~Liu, L.~Wang, The dawn of lmms: Preliminary explorations with gpt-4v(ision) (2023).
\newblock \href {http://arxiv.org/abs/2309.17421} {\path{arXiv:2309.17421}}.

\bibitem{r2-51}
X.~Zhang, Y.~Lu, W.~Wang, A.~Yan, J.~Yan, L.~Qin, H.~Wang, X.~Yan, W.~Y. Wang, L.~R. Petzold, Gpt-4v(ision) as a generalist evaluator for vision-language tasks (2023).
\newblock \href {http://arxiv.org/abs/2311.01361} {\path{arXiv:2311.01361}}.

\bibitem{abdelfattah2020ttpla}
R.~Abdelfattah, X.~Wang, S.~Wang, Ttpla: An aerial-image dataset for detection and segmentation of transmission towers and power lines, in: Proceedings of the Asian Conference on Computer Vision, 2020.

\bibitem{gomes7884270}
A.~Gomes, \href{https://doi.org/10.5281/zenodo.7884270}{{A Semantically Annotated 15-Class Ground Truth Dataset for Substation Equipment}} (May 2023).
\newblock \href {https://doi.org/10.5281/zenodo.7884270} {\path{doi:10.5281/zenodo.7884270}}.
\newline\urlprefix\url{https://doi.org/10.5281/zenodo.7884270}

\bibitem{buslaev2020albumentations}
A.~Buslaev, V.~I. Iglovikov, E.~Khvedchenya, A.~Parinov, M.~Druzhinin, A.~A. Kalinin, Albumentations: fast and flexible image augmentations, Information 11~(2) (2020) 125.

\end{thebibliography}

\end{document}